\documentclass{article}

\usepackage[final]{neurips_2024}

\usepackage{graphicx} 
\usepackage[utf8]{inputenc} 
\usepackage[T1]{fontenc}    
\usepackage{hyperref}       
\usepackage{url}            
\usepackage{booktabs}       
\usepackage{amsfonts}       
\usepackage{nicefrac}       
\usepackage{microtype}      
\usepackage{xcolor}         

\usepackage{natbib}
\usepackage{tabularx}

\title{Enhancing Reasoning to Adapt Large Language Models for Domain-Specific Applications}

\author{
  Bo Wen\textsuperscript{1}, Xin Zhang\textsuperscript{1,2}\\
 \textsuperscript{1}  IBM T. J. Watson Research Center, Yorktown Heights, NY, USA \\
 \textsuperscript{2} MIT-IBM Watson AI Lab, Cambridge, MA\\
  \texttt{Emails: bwen@us.ibm.com, xzhang@us.ibm.com} \\
  }

\begin{document}

\maketitle

\begin{abstract}
This paper presents SOLOMON, a novel Neuro-inspired Large Language Model (LLM) Reasoning Network architecture that enhances the adaptability of foundation models for domain-specific applications. Through a case study in semiconductor layout design, we demonstrate how SOLOMON enables swift adaptation of general-purpose LLMs to specialized tasks by leveraging Prompt Engineering and In-Context Learning techniques. Our experiments reveal the challenges LLMs face in spatial reasoning and applying domain knowledge to practical problems. Results show that SOLOMON instances significantly outperform their baseline LLM counterparts and achieve performance comparable to state-of-the-art reasoning model, o1-preview. We discuss future research directions for developing more adaptive AI systems that can continually learn, adapt, and evolve in response to new information and changing requirements.
\end{abstract}

\section{Introduction}
The rapid advancements in large language models (LLMs) have revolutionized various aspects of artificial intelligence, enabling them to understand and generate human-like text with remarkable proficiency. However, adapting these general-purpose models to domain-specific tasks remains a significant challenge. In this paper, we introduce SOLOMON (System for Optimizing Language Outputs through Multi-agent Oversight Networks), a Neuro-inspired LLM Reasoning Network Architecture that leverages Prompt Engineering and In-Context Learning techniques, and demonstrate how SOLOMON can effectively adapt from its original design purpose in medical applications to a new domain: semiconductor layout design. Section \ref{sec:architecture} presents the SOLOMON architecture and highlights its design principles that contribute to enhanced adaptability.

To provide context for our experiment, we first examine how a designer might attempt to use ChatGPT (with GPT-4o mode) for a via connection design task in section \ref{sec:problem}. This exploration reveals a critical limitation: while LLMs can accurately recite textbook definitions of domain-specific concepts, they struggle to extract and apply expert knowledge to solve practical tasks. Human needs to translate high-level concepts into specific geometric requirements, which the LLM can then use to generate code for drawing shapes. This highlights the key challenge in adapting LLMs for domain-specific applications: their limited reasoning capabilities.

In section \ref{sec:evaluation}, we developed a set of 25 tasks ranging from basic geometric shapes to complex semiconductor structures, to evaluate our SOLOMON architecture against five different LLMs. These tasks assess spatial reasoning capabilities and adaptability across various complexity levels. Through these experiments, we demonstrate SOLOMON's superior performance compared to standalone LLMs, and reaching the level of state-of-the-art reasoning models like O1-preview. 

Our findings emphasize the crucial role of reasoning in enhancing LLMs' adaptability to diverse domain applications. This study contributes to ongoing research in adaptive foundation models, providing insights into how to improve reasoning capabilities with inspiration from neuroscience.

\section{Neuro-inspired LLM Reasoning Network Architecture}
\label{sec:architecture}
SOLOMON's architecture (Fig. \ref{fig:solomon_architecture}) is inspired by two neuro-inspired theories: Brain-like AGI \cite{Byrnes2022} and the Free Energy Principle (FEP) \cite{Parr2022}. The former inspired us to use a pool of thoughts from multiple LLMs to discover the best reasoning plan. From the latter, we applied the FEP's main claim, \textit{human attention focuses on minimizing the differences between goals and perceptions}, to select relevant information and avoid common pitfalls. The key components of SOLOMON are:

\begin{figure}[ht]
\centering
\includegraphics[width=\textwidth]{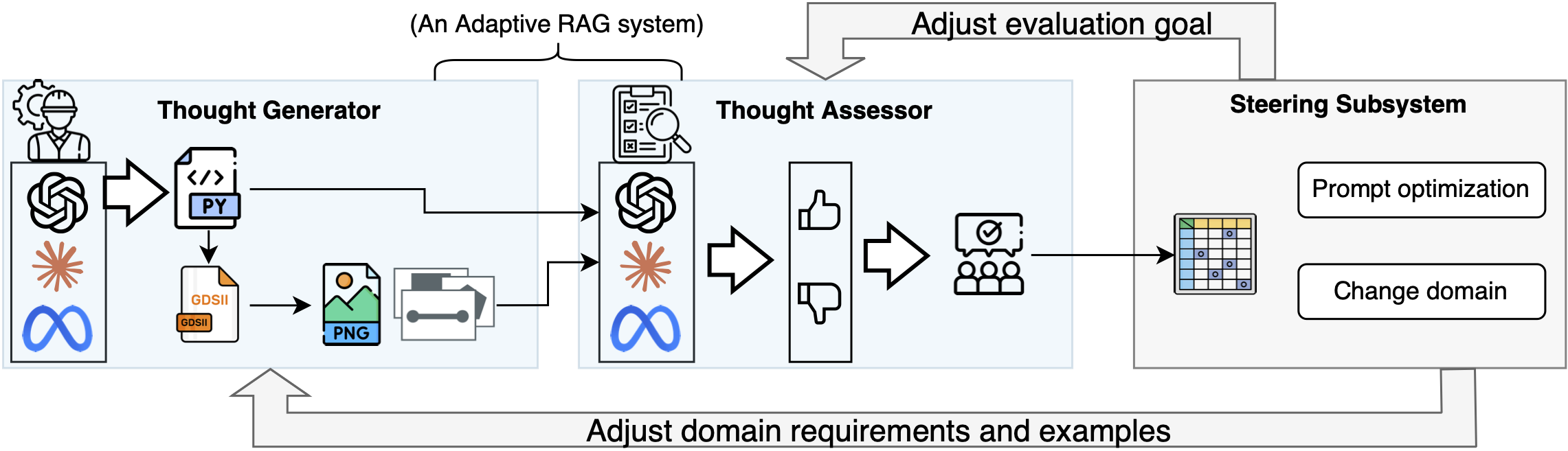}
\caption{SOLOMON Architecture Diagram}
\label{fig:solomon_architecture}
\end{figure}

\paragraph{Thought Generators:} A diverse pool of LLMs generating thoughts for the target task. This component forms an efficient parallel search engine through the Tree-of-Thoughts \cite{yao2023treethoughtsdeliberateproblem, zhang2024cumulativereasoninglargelanguage, Besta_2024, besta2024demystifyingchainstreesgraphs} and functions as an adaptive RAG system for the Thought Assessor. By pooling thoughts from multiple LLMs with distinct knowledge bases and reasoning abilities, it provides a more flexible and effective mechanism for sampling diverse ideas compared to common embedding-based RAG. This approach also mitigates biases inherent in single LLM knowledge bases. Noted that the individual LLMs in the Thought Generators can be further enhanced with proprietary knowledge through classic RAG techniques.

\paragraph{Thought Assessor:} An LLM-based system that analyzes the proposed ``Thoughts'' to generate a refined output. It conducts in-context learning on the Thought Generators' output and follows the Free Energy Principle for goal-oriented assessments on consensus and differences. This approach enhances the LLM-as-a-Judge method \cite{zheng2023judgingllmasajudgemtbenchchatbot, lin2023llmevalunifiedmultidimensionalautomatic}, enabling self-reflection \cite{ji2023mitigatinghallucinationlargelanguage} and guarding against hallucinations \cite{guerreiro2023lookingneedlehaystackcomprehensive}, thus improving AI safety and reliability.

\paragraph{Steering Subsystem:} A human-operated component that controls the attention of the Thought Generators and Thought Assessor. It uses Prompt Engineering to modify the goals of other components, enabling swift adaptation to different domain requirements through goal-directed exploration of the search space. This enhances the system's versatility across various applications by simply adjusting the attention focus.

This architecture offers significant advantages over traditional fine-tuning approaches, eliminating the need for recurrent fine-tuning associated with upgrading underlying LLMs or updating domain-specific knowledge. Basing on Prompt Engineering techniques, SOLOMON enables building more flexible AI systems capable of addressing diverse specialized contexts.

\section{Problem: Spatial Reasoning and Domain Knowledge Application}
\label{sec:problem}
Layout design in semiconductor processes requires not only generating correct basic geometric shapes but also spatial reasoning to create proper ``layouts'' that meet specific requirements. Via connections, which create 3D electrical pathways between different chip layers, exemplify this challenge. While seemingly simple, typically consisting of circular vias and rectangular metal connections, they demand precise positioning and sizing to ensure no short or open circuits when building the 2D layout into a 3D structure.

\begin{figure}[ht]
\centering
\includegraphics[width=0.8\textwidth]{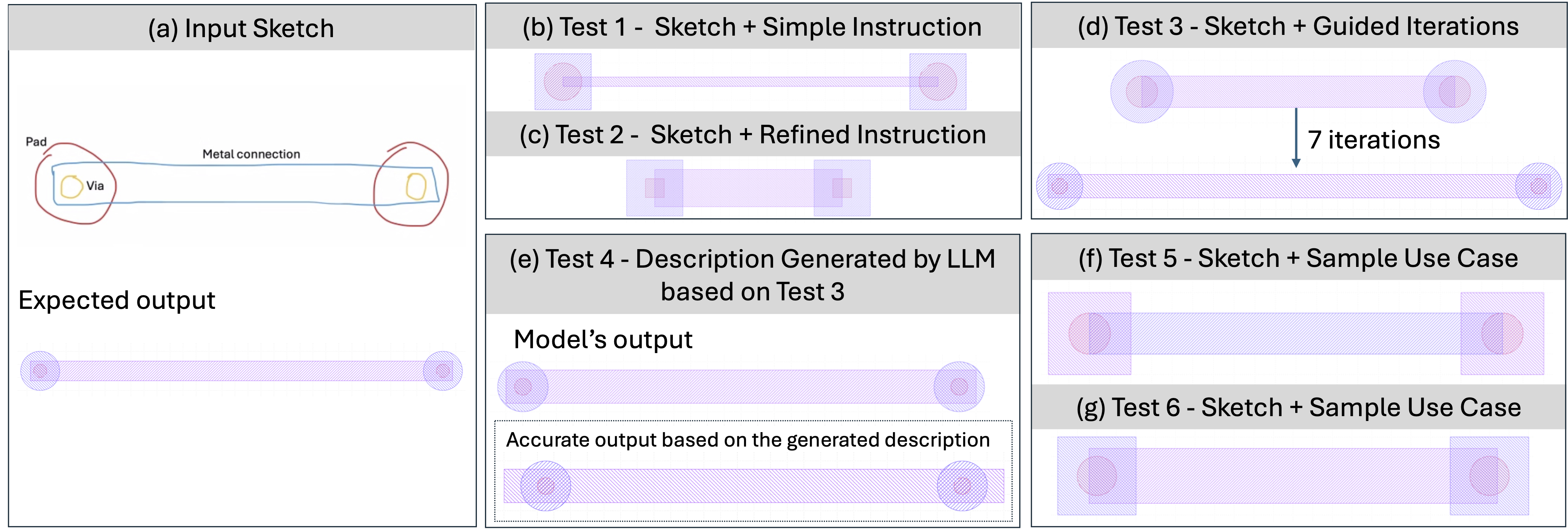}
\caption{Sketch input and ChatGPT-generated outputs for the via connection experiment. The sketch depicts a desired layout with two vias connected by a metal layer and circular contact pads on top. The outputs show the progression of ChatGPT's understanding and refinement of the layout based on iterative feedback and context provided by the user.}
\label{fig:via_experiment}
\end{figure}

We conducted a series of tests by providing a sketch(image) together with different text prompts to ChatGPT (GPT-4o). The sketch are color-coded to represent different layers (e.g., yellow for via, blue for metal layer, red for contact pad) to help ChatGPT understand the spatial relationships. Figure \ref{fig:via_experiment} illustrates the sketch inputs and corresponding ChatGPT-generated outputs for each test case.
In \textbf{Test 1}, we provided the sketch (Figure ~\ref{fig:via_experiment}(a)) with simple instructions. ChatGPT generated a runable code, but the metal layer width is narrower than the via. \textbf{Test 2} refined the instructions, but the output incorrectly used square vias \footnote{square vias are difficult to fabricate in semiconductor etching process, so they are not used in practice}, and failed to properly cover the vias with metal. In \textbf{Test 3}, we provided a more detailed description. After seven iterations of feedback, the LLM finally produced the correct layout (Figure ~\ref{fig:via_experiment}(d)). See Appendix \ref{appendix:via_connection} for detailed prompts and the iterative process.

\textbf{Test 4} reversed the process by asking the LLM to create a detailed prompt based on the correct layout from \textbf{Test 3} (Appendix \ref{appendix:via_connection}). The LLM described each component's size and location in detail but hallucinated an additional requirement: \textit{a 50-unit space between the vias and the edges of the metal connection}. This would result in the layout shown in the dotted rectangle in Figure~\ref{fig:via_experiment}(e), where the metal extends beyond the contact pad. Interestingly, when given this ``wrong'' prompt, GPT-4o ignored the added specification and produced a layout matching the original design, with the metal not extending beyond the pad. Code inspection revealed that the LLM used another requirement, \textit{Leave a margin of 10 units between the edge of the metal and the pads}, to calculate the metal edge position in both x and y directions, although this statement was intended only for the y-direction margin. Using this version of prompt in the baseline evaluations (Section \ref{sec:evaluation}), o1-preview and Llama-3.1-405B each produced the ``non-extending'' version in one out of 5 runs, indicating some ambiguity in the specification.

To further test our hypothesis, we conducted \textbf{Tests 5} and \textbf{6}, removing numerical values from the prompt and incorporating domain-specific context (e.g., 3D packaging and Through-Silicon Vias). This approach, however, degraded LLM performance, revealing a critical limitation: while LLMs possess textbook knowledge of semiconductor concepts, they struggle to translate this into practical design requirements. For instance, LLMs failed to apply common engineering knowledge, such as using wider metal layers to connect vias or leaving margin space between components in different layers to account for layer misalignment.

This finding highlights a critical insight: to enhance the adaptability of LLM-based AI systems, simply increasing the model size to memorize more information is insufficient; instead, we should prioritize developing LLMs' reasoning capacity to effectively utilize their knowledge in practical problem-solving scenarios.

\section{SOLOMON Performance and Comparison}
\label{sec:evaluation}
To evaluate SOLOMON's effectiveness in enhancing spatial reasoning for semiconductor layout design, we created a dataset of 25 layout design tasks. These tasks were categorized into four groups: Basic Shapes 1 and 2 included simple geometric shapes such as circles, polygons and text, which serve as the building blocks for more complex layouts. The Advanced Shapes category involved more intricate designs, such as serpentine and spirals, to test the models' ability to handle complex geometries. Finally, the Complex Structures category included tasks that required the composition of multiple shapes to form functional layouts, such as a Dense Layer Diode (DLD) chip, MicrofluidicChip, and the ViaConnection test case. These tasks were designed to benchmark the AI systems' capability in generating layouts that are representative of real-world semiconductor design needs.

We provided task requirements with a system prompt asking the LLMs to use Chain-of-Thought to analyze the task and write Python code to create a GDSII output. The evaluation process involved running the generated code to produce GDSII files, which were then converted to PNG images. Human evaluators categorized the output into five categories: correct, scaling error, partially correct, shape error, and runtime error. Five LLMs (GPT-4o, Claude-3.5-Sonnet, Llama-3.1-70B, Llama-3.1-405B, and o1-preview) were used for the baseline experiment, with each task run 5 times per model. (See Appendix \ref{appendix:task_prompts_and_performance} for details of prompts and example outputs.)

To evaluate SOLOMON, we utilized 20 thoughts generated by GPT-4o, Claude, and two Llama-3.1 models from the baseline experiment. We created four SOLOMON instances, each using one of these LLMs as a Thought Assessor, excluding o1-preview which served as our benchmark for state-of-the-art reasoning performance.

\begin{figure}[ht]
  \centering
  \includegraphics[width=\textwidth]{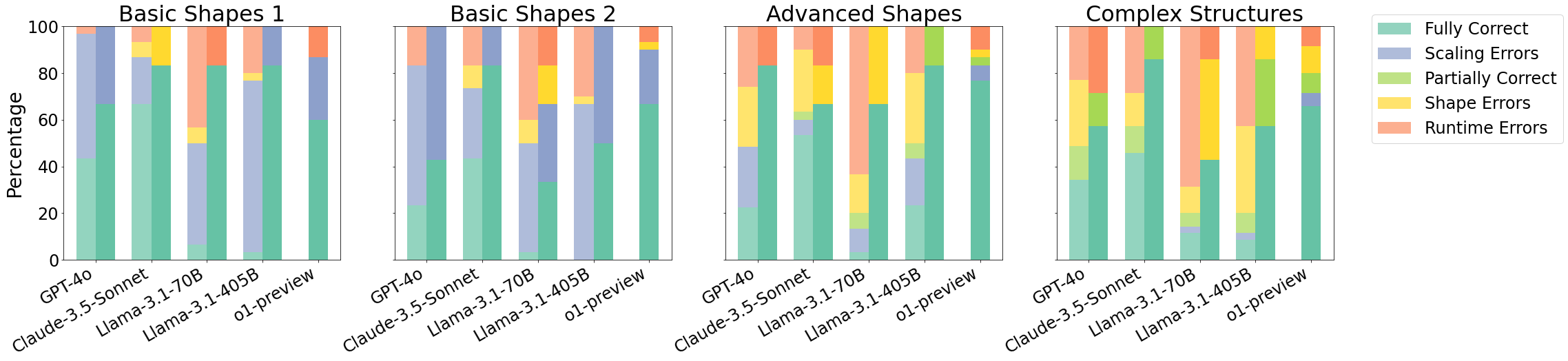}
  \caption{Performance comparison between SOLOMON instances, their baseline counterpart single LLMs, and o1-preview across different layout design task categories. Lighter colored bars on left represent baseline performance of individual LLMs, while darker bars on right show the performance of corresponding SOLOMON instances. O1-preview results serve as a benchmark for state-of-the-art reasoning performance.}
  \label{fig:performance_comparison}
\end{figure}

Figure \ref{fig:performance_comparison} presents a performance comparison between the SOLOMON instances, their baseline counterparts, and the o1-preview model.

The results demonstrate that the SOLOMON architecture significantly improves the performance of all four LLMs compared to their baseline. The most notable improvements are observed in the reduction of runtime errors, which can be attributed to the Thought Assessor seeing the error log of previous generated code and knowing what to avoid. This aligns with the design principle of the hierarchical, self-reflection mechanism, which aims to mitigate individual LLM's hallucination and blind-spot.

The second most notable problem in the baseline is scaling errors. We intentionally requested basic shapes to be drawn in millimeters to challenge the LLMs: they need to recall that the default unit in the gdspy library is micrometers. Sometimes LLMs simply failed to notice this and produced incorrect results. Additionally, each LLM seems to have bias when they hallucinate the default unit: Llama models perfers millimeters, Claude models sometimes recalls nanometers, and GPT-4o occasionally used meters. This issue was particularly problematic for Llama-3 models, when sometimes it correctly recalled the micrometer default but would insist that the user was wrong to request millimeters and proceed to draw without scaling, justifying it with comments like ``not mm, as the GDSII format is in micrometers.'' Such ``arrogant'' behavior and misalignment with simple instructions could be harmful for deploying LLMs as fully autonomous AI agents. A recent Nature paper \cite{ZhouNature2024} has also discussed similar observations.

The SOLOMON architecture improves performance across all models, including Llama-3. By incorporating diverse perspectives, it reduces stubbornness and increases accuracy. SOLOMON instances also show enhanced ability to handle shape errors and partial correctness issues, as the Thought Assessor can identify and correct errors related to arithmetic miscalculations or incorrect relative positioning of shapes. For more details, see Appendix \ref{appendix:baseline_errors}.

Comparing SOLOMON instances with o1-preview, we find that SOLOMON achieves comparable or superior results. All SOLOMON instances outperformed o1-preview in Basic Shape 1 categories, with the Claude-based SOLOMON surpassing o1-preview in 3 categories overall. 

Interestingly, Llama-3 based SOLOMON instances also received significant performance boost, eventhough they don't receive the image inputs, suggesting that the thought assessment mechanism indeed works for more than just image understanding. Additionally, insufficient information linking images to corresponding code and error logs sometimes resulted in misinterpretation for GPT-4o and Claude.

Analysis of SOLOMON errors reveals that performance depends heavily on the quality and consensus of initial thoughts. Tasks with ambiguous requirements often leads to significant disagreement among initial thoughts, leading to confusion of Thought Assessor and degraded performance, see Appendix Table \ref{table:dldchip}. These areas present opportunities for future improvements to the SOLOMON architecture.

\section{Conclusion and Future Work}
The introduction of the SOLOMON architecture significantly improved performance in semiconductor layout design tasks, particularly in reducing runtime errors and enhancing spatial reasoning capabilities. Our experiments demonstrated that SOLOMON instances outperformed their baseline LLM counterparts across various task categories, with some instances even surpassing the state-of-the-art o1-preview model in certain areas. This improvement validates the effectiveness of our neuro-inspired approach in enhancing LLMs' adaptability to domain-specific applications.

However, challenges remain in translating domain knowledge into practical design requirements. Our via connection experiment revealed that while LLMs can accurately recite textbook definitions of domain-specific concepts, they struggle to extract and apply expert knowledge to solve practical tasks. Investigating the potential of stacking multiple SOLOMON layers to form a hierarchical reasoning model capable of recalling and reasoning with domain knowledge for task-solving is one of our major future focus.

Other future research directions include:
(1) Developing more comprehensive benchmark datasets for evaluating AI systems in layout design tasks.
(2) Improving the linking between multimodal inputs (images and corresponding code+error) in the thoughts to enhance the Thought Assessor's interpretation abilities.
(3) Exploring SOLOMON's performance when initial thoughts are of lower quality, and developing goal-oriented iterative learning mechanisms to improve thought quality through feedback loops.
(4) Applying the SOLOMON architecture to a broader range of domain-specific tasks, such as power grid design and financial modeling.

In conclusion, while our results demonstrate the promise of LLMs as layout design copilots, further advancements in reasoning capabilities and domain knowledge application are necessary for their effective integration into semiconductor design processes and other specialized domains. The SOLOMON architecture represents a significant step towards creating more adaptive and capable AI systems for complex, domain-specific applications.

\paragraph{Open Source Code and Dataset:} Visit our GitHub repository for the complete benchmark dataset of 25 tasks and LLM-calling code under the Apache 2.0 license at \url{https://github.com/wenboown/generative-ai-for-semiconductor-physical-design}. See Appendix \ref{appendix:code_and_dataset} for more details on Experiments Compute Resources requirements.

\paragraph{Acknowledgment:} We thank Kuan Yu Hsieh for her valuable contribution in creating the dataset of 25 tasks with ground truth and her exploration work on the via connection test cases.

\bibliographystyle{plainnat}
\bibliography{maintext}

\newpage
\appendix

\section{Appendix}
\subsection{Open Source Code and Dataset}
\label{appendix:code_and_dataset}

To ensure reproducibility and facilitate further research, we release the complete benchmark dataset of 25 tasks and LLM-calling code under the Apache 2.0 license at \url{https://github.com/wenboown/generative-ai-for-semiconductor-physical-design}. This repository includes 5 runs of results (LLM answers, Python code, error logs, and PNGs) for each task in the baseline experiment for reproducibility. While the SOLOMON code is proprietary, its output results are included.

All LLM experiments were conducted by calling APIs: GPT-4o via OpenAI, Claude via Anthropic, and Llama 3s via IBM Watsonx. Total API costs are about \$50 including re-runs of failed tasks and iterative testings. The local code (for calling APIs, collecting responses and saving to disk, running the LLM-generated code, and analysis) was run on a virtual machine with RHEL 8.0, equipped with a 32-core CPU and 256GB of memory.

The complete dataset, including prompts, ground truths, and LLM outputs, is available in our repository. This contains all materials needed to reproduce our baseline experiments and conduct further research.

\subsection{Task Prompts and Baseline LLM Performance}
\label{appendix:task_prompts_and_performance}

The system prompt used for baseline experiment (thought generating) for all tasks was as follows:

\begin{verbatim}
You are an expert Python developer specialized in generating layout designs 
in GDS (GDSII) format. Your task is to assist the user in creating Python 
code that accurately draws layout designs while being mindful of the 
geometric relationships and layout accuracy.

Write down your thinking step by step before you start coding:
1. Always start by understanding the overall design requirements provided 
   by the user.
2. Break down the design into smaller components and define each geometric 
   shape with precise coordinates.
3. Ensure that all shapes and elements maintain their correct geometric 
   relationships, such as alignment, spacing, and proportional dimensions.
4. Validate each step of the design process to avoid errors and maintain 
   accuracy.

Use the 'gdspy' library to generate the GDS layout:
1. Parse the user's design specifications.
2. Define the library and cell for the GDS layout.
3. Create each geometric element (e.g., rectangles, polygons) with precise 
   coordinates.
4. Ensure elements are placed correctly and maintain their intended 
   relationships.
5. Save the design to a GDS file.

Provide all the code in a single ```python ``` block to the user without 
postamble. Do not include any other ``` block in your response to avoid 
parsing error in following steps.

Be meticulous in your approach, and always consider the geometric 
relationships and layout accuracy in every step of the design process.
\end{verbatim}

Here are the complete list of all 25 task prompts (questions):
\begin{table}[htbp]
\caption{Basic Shapes 1 Questions}
\label{table:questions-basic_shapes_1}
\begin{tabularx}{\textwidth}{@{}lX@{}}
\hline
\textbf{Shape} & \textbf{Question} \\ \hline
Circle & Write a Python code to generate GDSII for a circle on layer 0, radius = 10 mm, center at 0,0. \\ \hline
Donut & Generate a donut shape with 10 mm outer radius and 5 mm inner radius. Make the circle smoother by setting max distance between point 0.01mm. \\ \hline
Oval & Generate an oval with major axis of 20 mm, minor axis of 13 mm, on layer 0, center at 0,0. \\ \hline
Square & Generate a square with width 10 mm, put lower right corner of the square at 0,0. \\ \hline
Triangle & Generate a triangle with each edge 10 mm, center at 0,0. \\ \hline
Grid & Draw the GDSII for a grid: Grid on Layer 1, DATATYPE 4, 5 µm grid, and total width is 200 µm and height is 400 µm, placed at coordinates (100,800) nanometers. \\ \hline
\end{tabularx}
\end{table}

\begin{table}[htbp]
\caption{Basic Shapes 2 Questions}
\label{table:questions-basic_shapes_2}
\begin{tabularx}{\textwidth}{@{}lX@{}}
\hline
\textbf{Shape} & \textbf{Question} \\ \hline
Heptagon & Generate a Heptagon with each edge 10 mm, center at 0,0. \\ \hline
Octagon & Generate an Octagon with each edge 10 mm, center at 0,0. \\ \hline
Trapezoid & Generate a Trapezoid with upper edge 10 mm, lower edge 20 mm, height 8 mm, center at 0,0. \\ \hline
Hexagon & Generate a regular hexagon with each edge 10 mm, center at 0,0. \\ \hline
Pentagon & Generate a regular pentagon with each edge 10 mm, center at 0,0. \\ \hline
Text & Generate a GDS file with the text "Hello, GDS!" centered at (0,0), with a height of 5 mm, on layer 1. \\ \hline
\end{tabularx}
\end{table}

\begin{table}[htbp]
\caption{Advanced Shapes Questions}
\label{table:questions-advanced_shapes}
\begin{tabularx}{\textwidth}{@{}lX@{}}
\hline
\textbf{Shape} & \textbf{Question} \\ \hline
Arrow & Generate an Arrow pointing to the right with length 10 mm, make the body 1/3 width of the head, start at 0,0. \\ \hline
SquareArray & Generate a square array with 5*5 mm square, for 10 columns and 10 rows, each 20 mm apart, the lower left corner of the upper right square is at 0,0. \\ \hline
Serpentine & Generate a serpentine pattern with a path width of 1 µm, 15 turns, each segment being 50 µm long and tall, starting at (0,0), on layer 2, datatype 6. \\ \hline
RoundedSquare & Draw a 10*10 mm square, and do corner rounding for each corner with r=1 mm. \\ \hline
Spiral & Generate a Parametric spiral with r(t) = e\^{}(-0.1t), for 0 <= t <= 6pi, line width 1. \\ \hline
BasicLayout & 1. Draw a rectangular active region with dimensions 10 µm x 5 µm.
2. Place a polysilicon gate that crosses the active region vertically at its center, with a width of 1 µm.
3. Add two square contact holes, each 1 µm x 1 µm, positioned 1 µm away from the gate on either side along the active region. \\ \hline
\end{tabularx}
\end{table}

\begin{table}[htbp]
\caption{Complex Structures Questions}
\label{table:questions-complex_structures}
\begin{tabularx}{\textwidth}{@{}lX@{}}
\hline
\textbf{Shape} & \textbf{Question} \\ \hline
RectangleWithText & Generate a GDS with a 30*10 mm rectangle on layer 0 with a text "IBM Research" at the center of the rectangle. Put the text on layer 1. \\ \hline
MicrofluidicChip & Draw a design of a microfluidic chip. On layer 0, it is the bulk of the chip. It is a 30 * 20 mm rectangle. On layer 2 (via level), draw two circular vias, with 2 mm radius, and 20 mm apart horizontally. On layer 3 (channel level), draw a rectangular shaped channel (width = 1 mm) that connects the two vias at their center. \\ \hline
ViaConnection & Create a design with three layers: via layer (yellow), metal layer (blue), and pad layer (red). The via radius is 10 units, pad radius is 30 units, and metal connection width is 40 units with a total length of 600 units. Position the first via at (50, 150) and the second via at (550, 150). Ensure the metal connection fully covers the vias and leaves a margin of 10 units between the edge of the metal and the pads. Leave a space of 50 units between the vias and the edges of the metal connection. \\ \hline
FiducialCircle & Draw a 3.2 mm circle, with fiducial marks inside. The fiducial marks should be a "+" sign, with equal length and width. Each marker should be 200 um apart. There will be annotations next to each marker. Row: A -> Z, column: start from 1. \\ \hline
ComplexLayout & 1. Draw three rectangular active regions with dimensions 20 µm x 5 µm, positioned horizontally with 5 µm spacing between them.
2. Create a complex polysilicon gate pattern consisting of multiple vertical and horizontal lines, with widths of 0.5 µm, forming a grid-like structure.
3. Add several contact holes (each 1 µm x 1 µm) positioned at the intersections of the polysilicon gate pattern and the active regions. \\ \hline
DLDChip & Draw a deterministic lateral displacement chip - include channel that can hold the array has gap size = 225 nm, circular pillar size = 400 nm, width = 30 pillars, row shift fraction = 0.1, add an inlet and outlet 40 µm diameter before and after the channel, use a 20*50 µm bus to connect the inlet and outlet to the channel. \\ \hline
FinFET & Draw a FinFET with the following specifications:
- Fin width: 0.1 µm
- Fin height: 0.2 µm
- Fin length: 1.0 µm
- Gate length: 0.1 µm
- Source/drain length: 0.4 µm
- Source/drain extension beyond the fin: 0.2 µm
Use separate layers for the fin, gate, and source/drain regions. \\ \hline
\end{tabularx}
\end{table}

To aid human evaluators, we organized task prompts, ground truth images, and LLM output images from different runs in a tabular format. This presentation offers a clear view of various models' performance in generating GDSII layouts, enabling easy comparison between expected results and actual outputs from different LLMs and the SOLOMON system.

Tables \ref{table:arrow} through \ref{table:dldchip} at the Appendix's end showcase 5 representative experiment results. For complete results, see our GitHub repository.

\subsection{Errors in Baseline Experiment}
\label{appendix:baseline_errors}

\subsubsection{Scaling Errors}
\label{appendix:scaling_errors}

The default unit in the gdspy library is micrometers. We requested basic shapes to be drawn in millimeters to test whether LLMs could correctly handle this unit conversion. All LLMs struggled to various degrees:

(a) Some LLMs failed to pay attention to the requested unit (millimeters) and did not perform the necessary scaling.

(b) In some cases, LLMs paid attention to the requested unit but made incorrect assumptions about gdspy's default unit. We observed biased hallucinations: Llama models tended to assume millimeters, Claude models sometimes defaulted to nanometers, while GPT-4o occasionally interpreted the default unit as meters.

(c) As mentioned in the main text, Llama-3 models were especially vulnerable to this issue. They sometimes assumed the user had made a mistake by requesting millimeters, and proceeded to draw in micrometers instead, justifying this choice with comments like "not mm, as the GDSII format is in micrometers".

\subsubsection{Shape Errors}
\label{appendix:shape_errors}

Incorrect shapes often resulted from LLMs making basic arithmetic errors. For instance, in the "Hexagon" task, Llama-3.1-405B once used an internal angle of 120 degrees, producing a triangle instead of a hexagon. However, in other runs, it correctly calculated the angle based on the number of edges. Many of these errors can be mitigated through Chain-of-Thought (CoT) prompting, which encourages the model to do calculations step-by-step.

\subsubsection{Runtime Errors}
\label{appendix:runtime_errors}

This section provides a detailed breakdown of the errors encountered during the baseline experiment for each LLM. The most frequent error across all models was \textit{AttributeError: module 'gdspy' has no attribute 'LayoutViewer'}, occurring 26 times (59.09\%) with GPT-4o and 33 times (61.11\%) with Claude-3.5-Sonnet. This error was less common in other models, appearing only once each for Llama-3.1-70B and o1-preview, and not at all for Llama-3.1-405B.

The prevalence of this error indicates that GPT-4o and Claude-3.5-Sonnet attempted to provide GUI output, which was unavailable in the runtime environment. However, this issue stems from a lack of specification about the runtime environment in the prompt, rather than being entirely the LLMs' fault.

To ensure a fair comparison, we re-ran all generated code with `LayoutViewer` lines commented out. The analysis presented in Figure \ref{fig:performance_comparison} and the following breakdown reflect these adjusted results.

Other common errors included hallucinations of nonexistent `gdspy` functions or methods, resulting in various `AttributeErrors` (e.g., `'CrossSection'`, `'Circular'`, `'Ellipse'`) and `TypeErrors`. Some errors were due to spelling mistakes, such as misspelling \textit{gdspy.Text} as \textit{gdspy.text}.

The subsequent analysis presents a detailed error breakdown for each LLM, ranked by ascending number of errors.

\paragraph{o1-preview}
Total errors: 12

The main errors for o1-preview included:
\begin{itemize}
    \item TypeError: GdsLibrary.write\_gds() got an unexpected keyword argument 'unit' (16.67\%)
    \item SyntaxError: invalid syntax (16.67\%)
    \item Various TypeErrors and AttributeErrors related to unexpected keyword arguments or missing attributes (66.66\%)
\end{itemize}

\paragraph{GPT-4o}
Total errors: 18

The most common errors for GPT-4o were:
\begin{itemize}
    \item TypeError related to unexpected keyword arguments (27.78\%)
    \item SyntaxError: invalid syntax (16.67\%)
    \item TypeError: 'float' object cannot be interpreted as an integer (11.11\%)
\end{itemize}

Other errors included various AttributeErrors, IndexErrors, and ValueErrors, each occurring once or twice.

\paragraph{Claude-3-5-sonnet}
Total errors: 21

The most frequent error for Claude-3-5-sonnet was:
\begin{itemize}
    \item TypeError: Text.\_\_init\_\_() got an unexpected keyword argument 'anchor' (38.10\%)
\end{itemize}

Other errors included:
\begin{itemize}
    \item TypeError: Path.\_\_init\_\_() got an unexpected keyword argument 'layer' (9.52\%)
    \item Various TypeErrors, ValueErrors, and AttributeErrors, each occurring once (52.38\%)
\end{itemize}

\paragraph{Llama-3-405b}
Total errors: 36

The most frequent errors for Llama-3-405b were:
\begin{itemize}
    \item TypeError: 'int' object is not subscriptable (8.33\%)
    \item SyntaxError: invalid syntax (8.33\%)
    \item Various TypeErrors related to unexpected keyword arguments or multiple values for arguments (16.68\%)
\end{itemize}

This model also encountered several ValueErrors and AttributeErrors.

\paragraph{Llama-3-70b}
Total errors: 68

The most prevalent error for Llama-3-70b was:
\begin{itemize}
    \item AttributeError: module 'gdspy' has no attribute 'Library'. Did you mean: 'library'? (36.76\%)
\end{itemize}

Other common errors included:
\begin{itemize}
    \item TypeError: GdsLibrary.write\_gds() got an unexpected keyword argument 'unit' (7.35\%)
    \item SyntaxError related to assignment (5.88\%)
    \item Various TypeErrors and AttributeErrors related to unexpected keyword arguments or missing attributes (25.00\%)
\end{itemize}

These error patterns suggest that all models struggled with correctly using the gdspy library, often attempting to use non-existent attributes or passing incorrect arguments to functions. Syntax errors were also common across models, indicating issues with code structure and Python syntax.

\subsubsection{Inefficient Code}
\label{appendix:inefficient_code}

In the DLDChip task, which involves creating a dense array of identical shapes, the Llama-3.1-405B model generated code that created a large number of objects and performed numerous boolean operations. This led to high memory usage and extended execution time, requiring the code to be terminated after approximately 15 minutes of runtime.

\subsubsection{Ambiguous Instructions}
\label{appendix:ambiguous_instructions}

In some cases, we observed that the LLM results mainly fell into two categories. After inspecting the prompts, we found that the instructions could be interpreted in two ways. In these cases, we counted both types of results as correct. However, when implementing a copilot, the agent should ask for clarification if the instructions are ambiguous.

\subsection{Via Connection Test Cases}
\label{appendix:via_connection}

\begin{figure}[ht]
\centering
\includegraphics[width=0.5\linewidth]{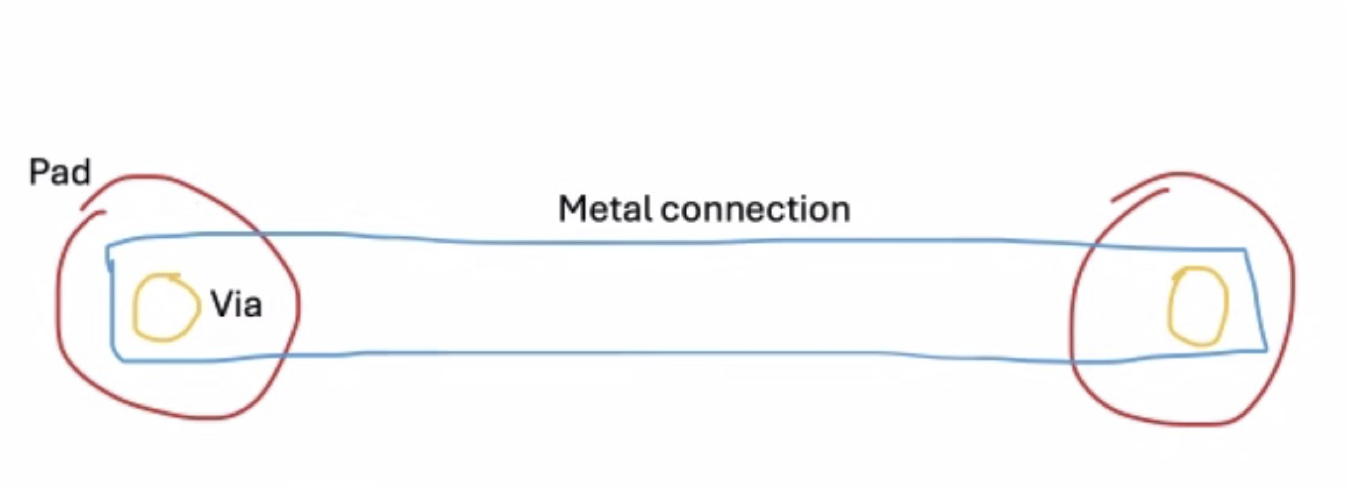}
\caption{Sketch used for via connection test cases}
\label{fig:sketch}
\end{figure}

Figure \ref{fig:sketch} shows the sketch used in all via connection test cases discussed in Section \ref{sec:problem}.

\subsubsection{Prompts for Via Connection Tests}
\label{appendix:via_prompts}

\paragraph{Test 1:} ``I have a sketch idea that i want to draw in GDSII, generate the python code for this design. each color represents an individual layer. We want to use a metal to connect two vias and put a pad on top of each via''

\paragraph{Test 2:} ``I have a sketch idea that i want to draw in GDSII, generate the python code for this design. each color represents an individual layer. We want to have two vias near each end on a piece of metal. And a pad on top of the metal.''

\paragraph{Test 3:} ``I have a sketch idea that i want to draw in GDSII, generate the python code for this design. each color represents an individual layer. we want use to connect two vias using a piece of metal and put a circular padding on top of each via''

\begin{figure}[ht]
\centering
\includegraphics[width=0.5\linewidth]{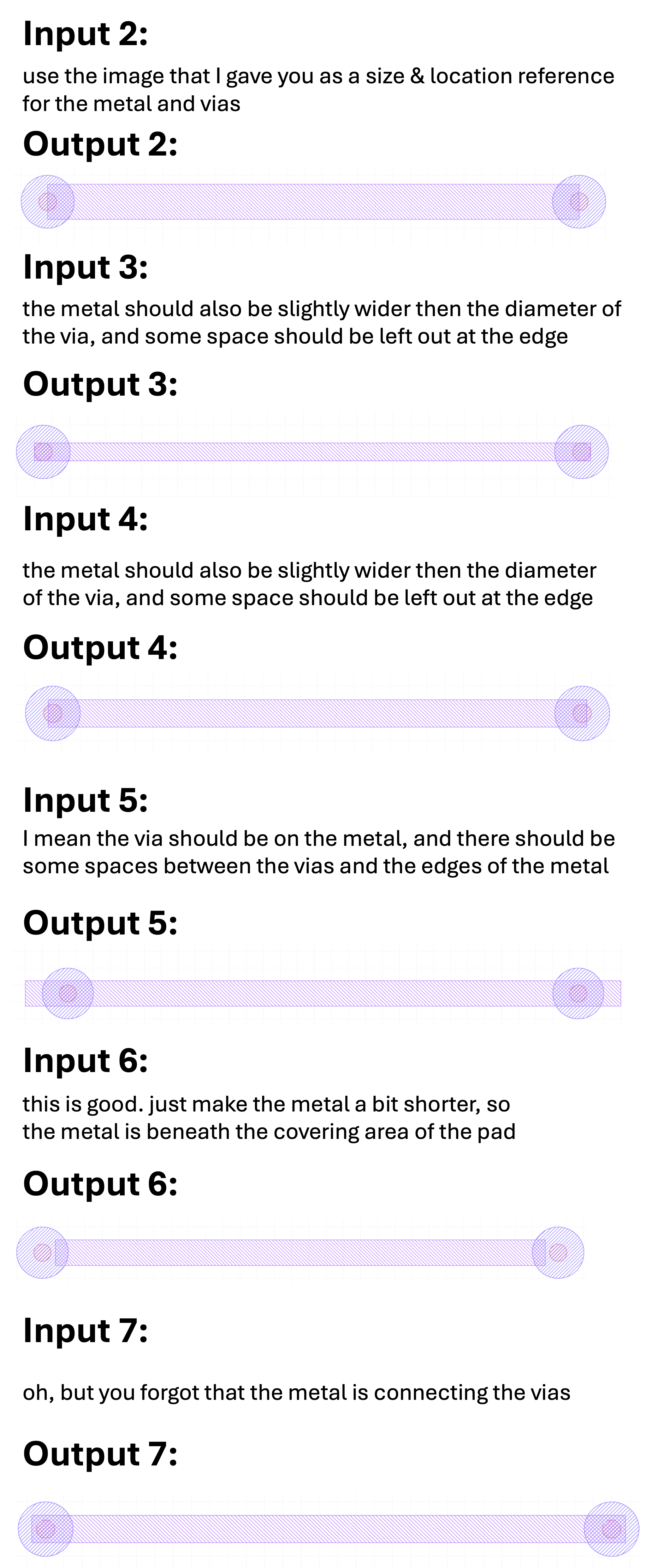}
\caption{Test 3 - Iterations to guide the model to generate desired output}
\label{fig:test3_chat}
\end{figure}

Figure \ref{fig:test3_chat} illustrates the iterative prompting process for Test 3, as discussed in Section \ref{sec:problem}.

\paragraph{Test 4 (Generated by LLM based on the final output in Test 3):}

\begin{verbatim}
Layers and Colors:
The design consists of three layers: via layer (yellow), 
metal layer (blue), and pad layer (red).
Dimensions:
Via: The radius of each via is 10 units.
Pad: The radius of each pad is 30 units.
Metal Connection: The width of the metal connection is 40 units, 
and the total length is 600 units.
Positions:
The first via is positioned at coordinates (50, 150).
The second via is positioned at coordinates (550, 150).
Connections and Coverage:
The metal connection should fully cover the vias, extending slightly 
beyond their edges.
Ensure the metal connection is slightly wider than the via diameter 
to provide full coverage.
Spacing and Margins:
Leave a margin of 10 units between the edge of the metal and the pads.
Ensure there is a space of 50 units between the vias and the edges 
of the metal connection.
Additional Requirements:
The metal connection should be shorter than the total length to fit 
beneath the covering area of the pads, leaving some space at the edges.
By providing detailed information like this, you can ensure that 
the design is accurately reproduced. 
If you have any specific design rules or preferences, make sure to 
include those as well.
\end{verbatim}

\paragraph{Test 5:} ``I have sketched a design for 3d packaging, where we have a metal connecting two TSVs , please generate the python code to draw a GDSII for this design.''

\paragraph{Test 6:} ``I have sketched a design for 3d packaging, where we have a metal connecting two TSVs , please generate the python code to draw a GDSII based on the sketch. each color represents an individual layer. The metal connection should fully cover the vias, extending slightly beyond their edges. Ensure the metal connection is slightly wider than the via diameter to provide full coverage.''

\clearpage
\begin{table}[p]
  \caption{Oval Task Question: Generate an oval with major axis of 20 mm, minor axis of 13 mm, on layer 0, center at 0,0.}
  \label{table:oval}
  \centering
  \begin{tabularx}{0.9\textwidth}{@{}XXXXXX@{}}
    \toprule
    \begin{tabular}{@{}c@{}}Ground Truth \\ \includegraphics[width=0.13\textwidth]{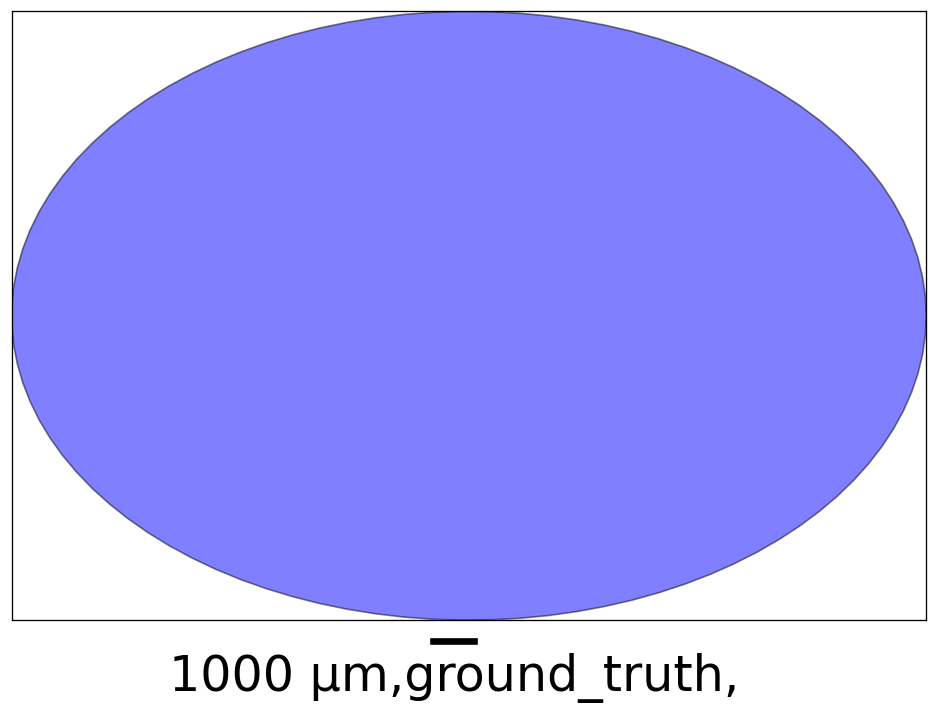}\end{tabular} & GPT-4o & Claude-3.5 & Llama-3-70B & Llama-3-405B & o1-preview \\
    \midrule
    SOLOMON & \includegraphics[width=0.13\textwidth]{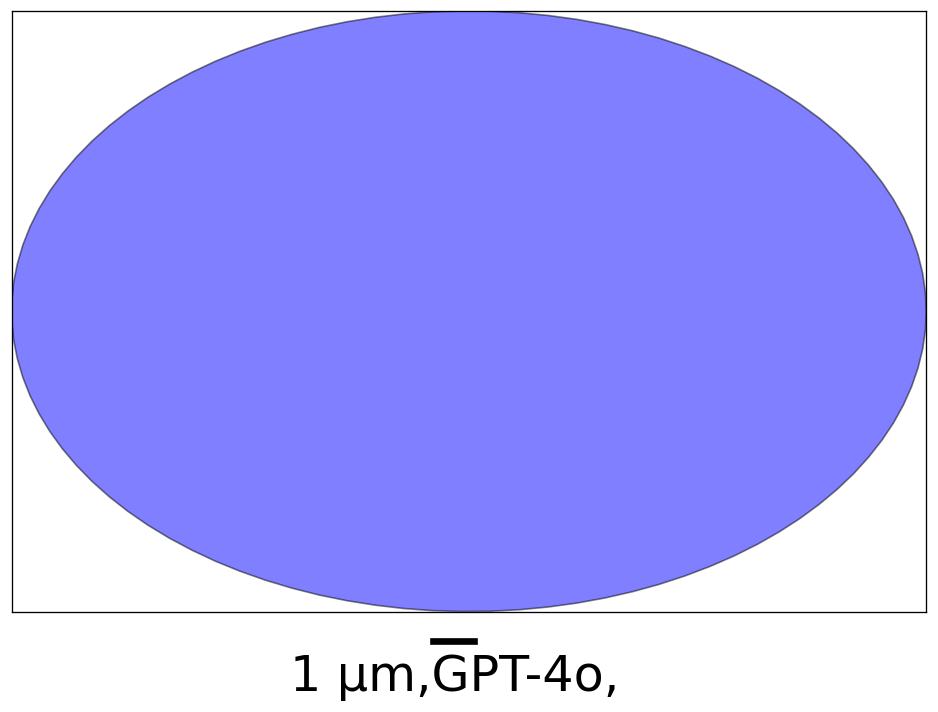} & \includegraphics[width=0.13\textwidth]{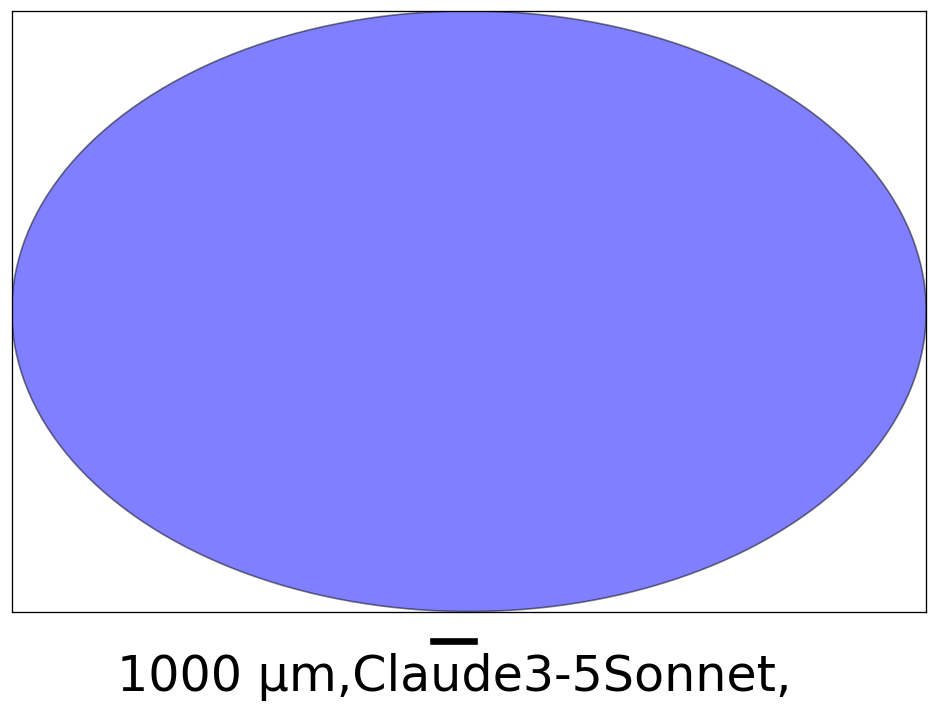} & \includegraphics[width=0.13\textwidth]{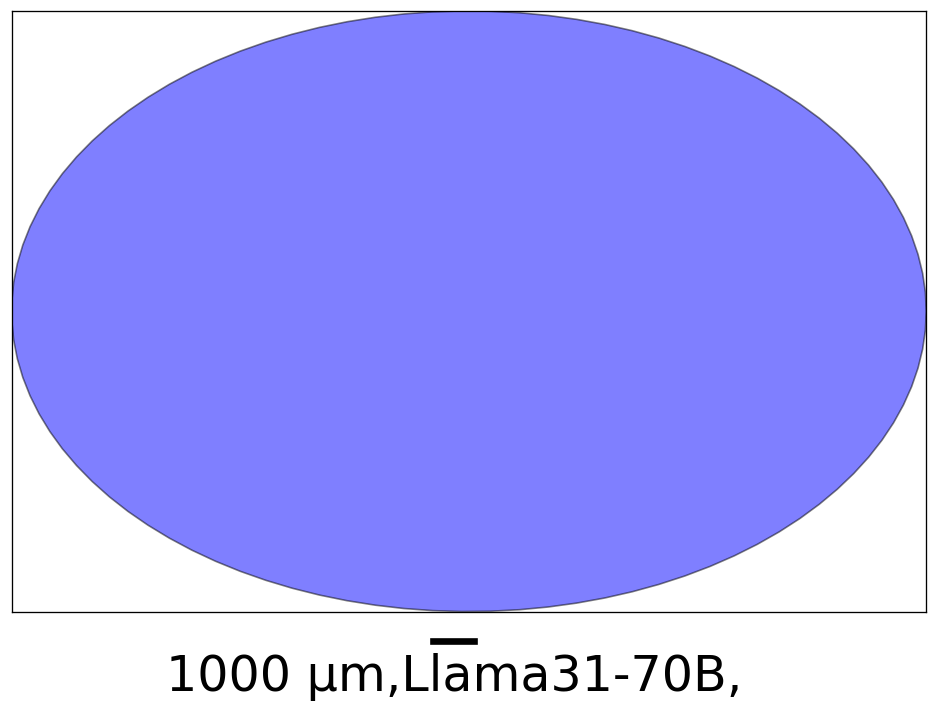} & \includegraphics[width=0.13\textwidth]{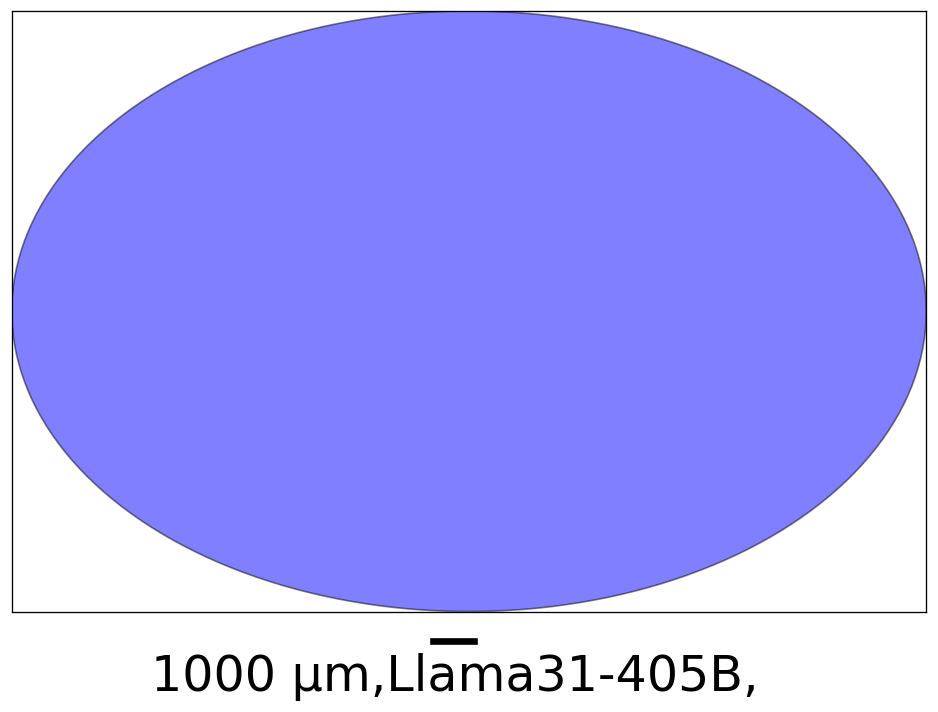} &  \\
    \begin{tabular}{@{}c@{}}Single LLM \\ Baseline \\ Run 1\end{tabular} & \includegraphics[width=0.13\textwidth]{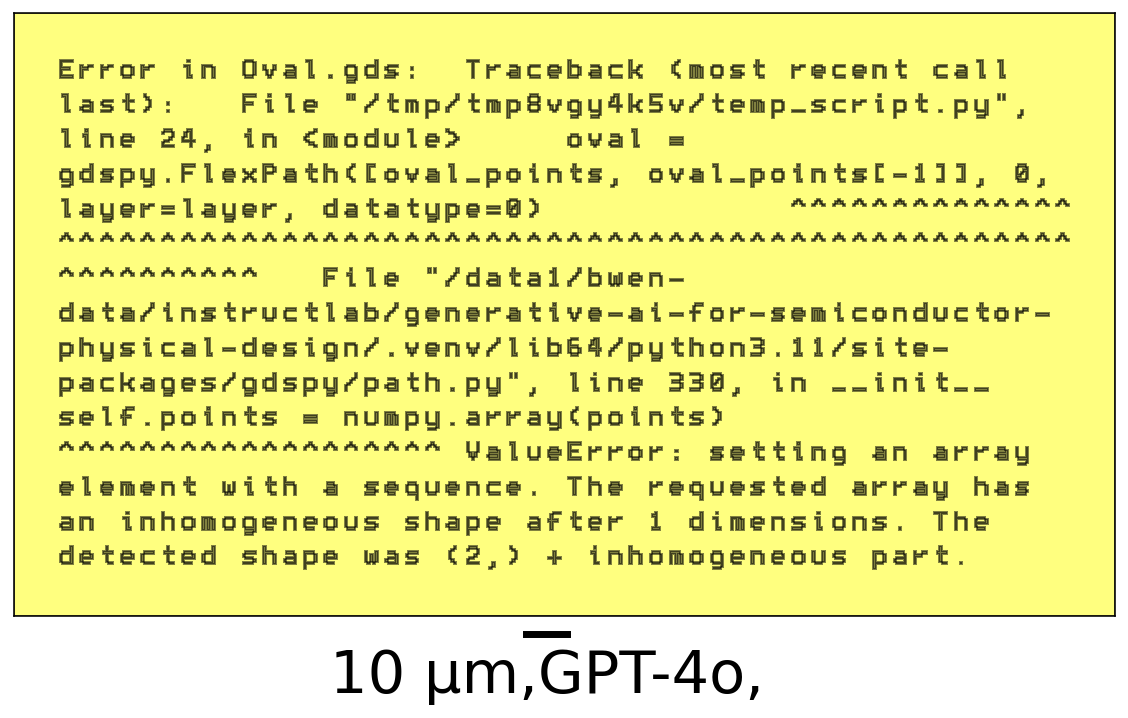} & \includegraphics[width=0.13\textwidth]{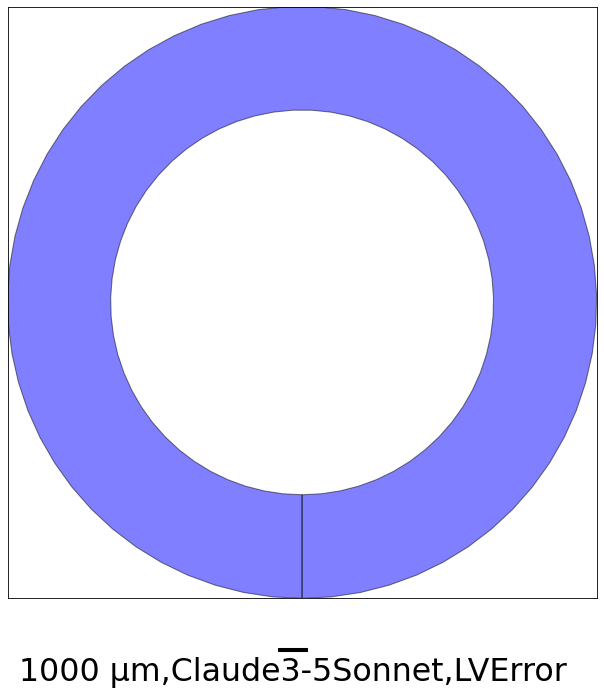} & \includegraphics[width=0.13\textwidth]{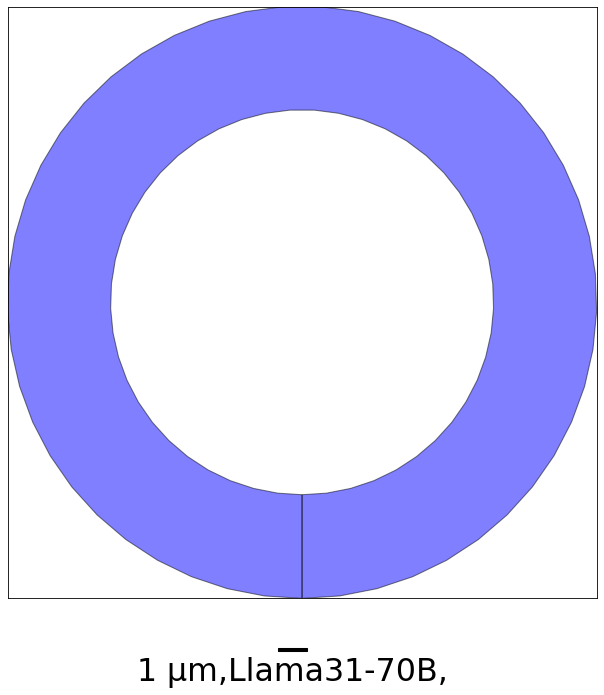} & \includegraphics[width=0.13\textwidth]{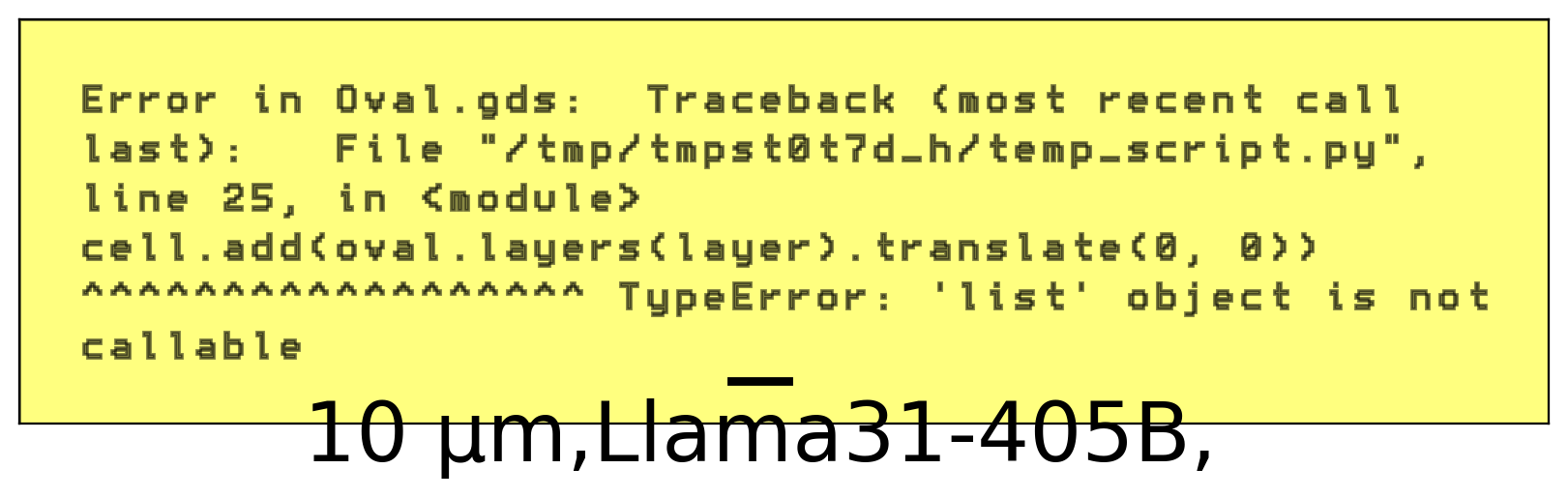} & \includegraphics[width=0.13\textwidth]{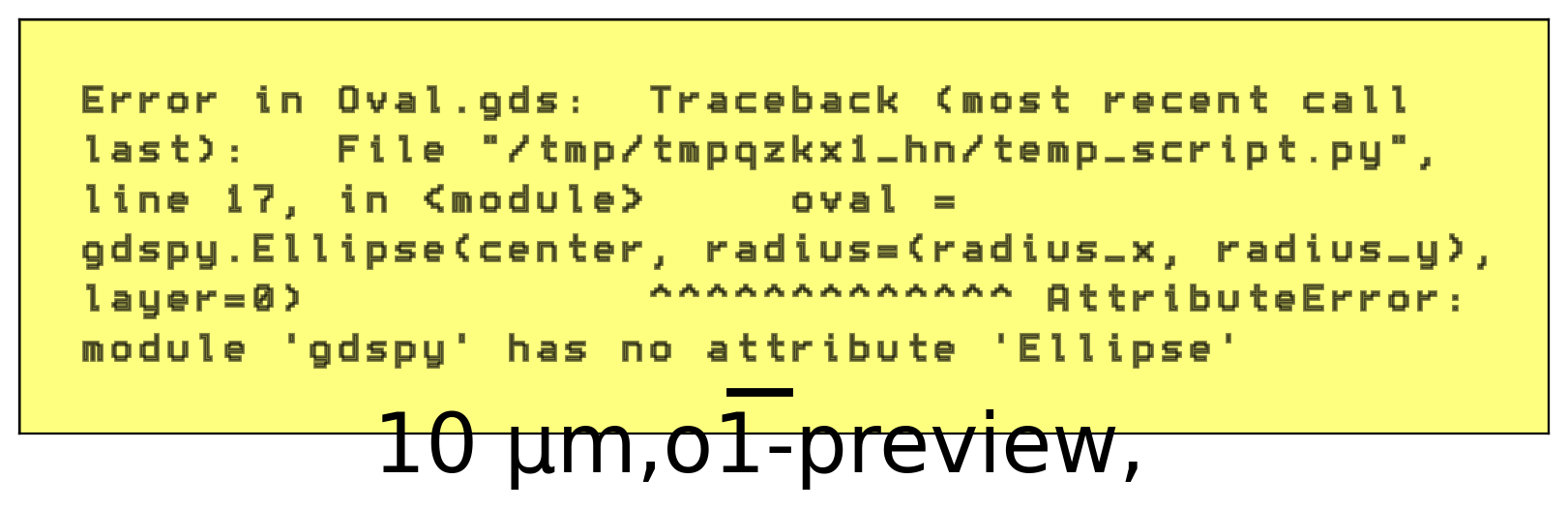} \\
    \begin{tabular}{@{}c@{}}Single LLM \\ Baseline \\ Run 2\end{tabular} & \includegraphics[width=0.13\textwidth]{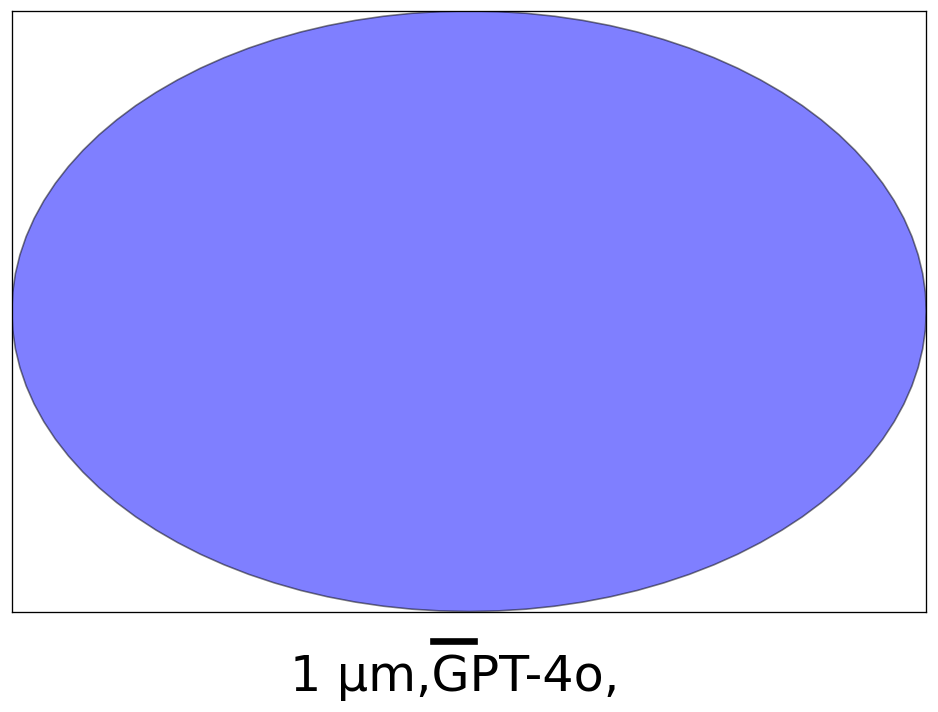} & \includegraphics[width=0.13\textwidth]{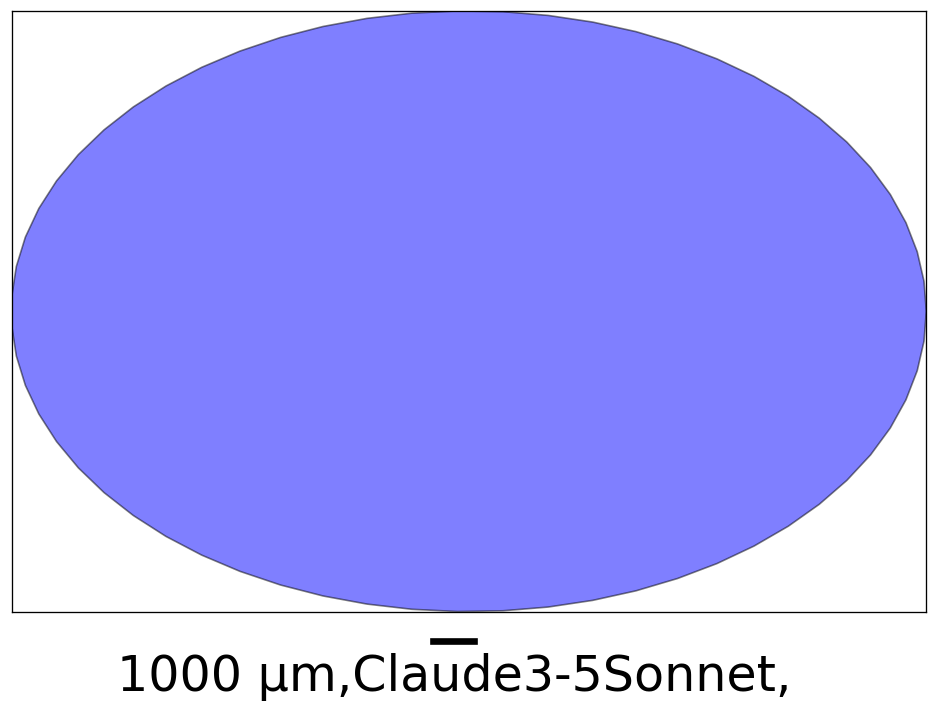} & \includegraphics[width=0.13\textwidth]{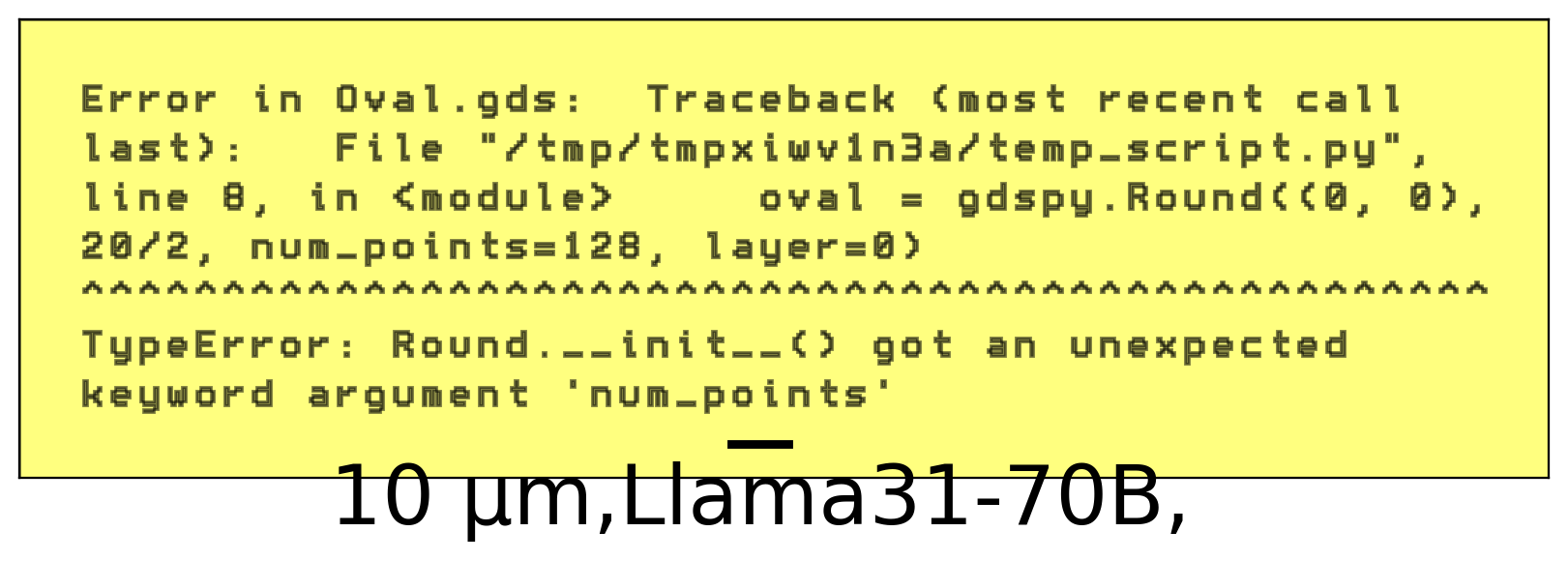} & \includegraphics[width=0.13\textwidth]{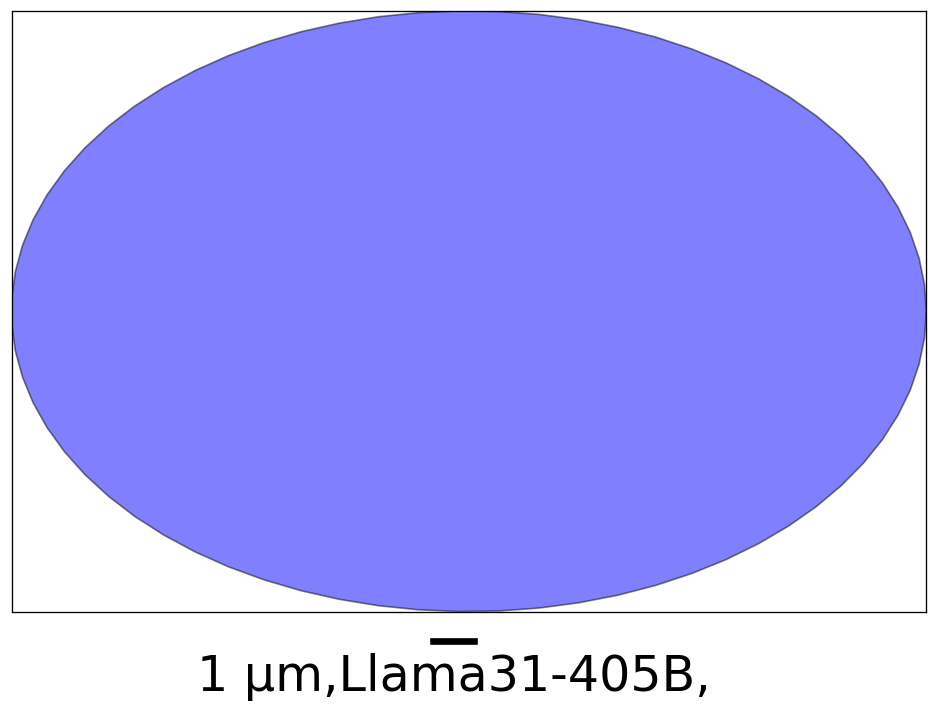} & \includegraphics[width=0.13\textwidth]{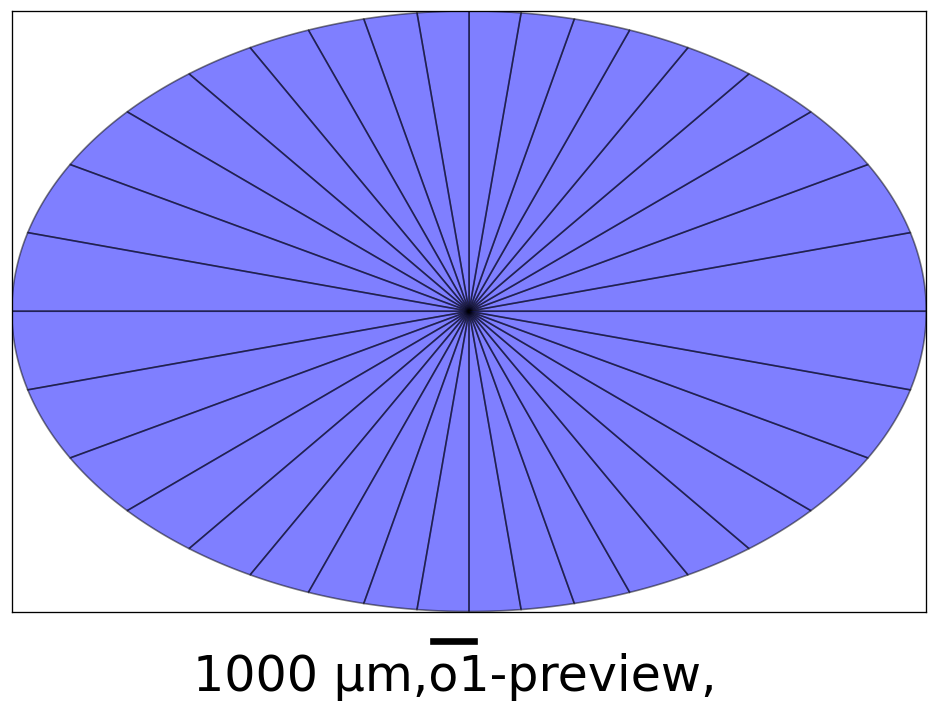} \\
    \begin{tabular}{@{}c@{}}Single LLM \\ Baseline \\ Run 3\end{tabular} & \includegraphics[width=0.13\textwidth]{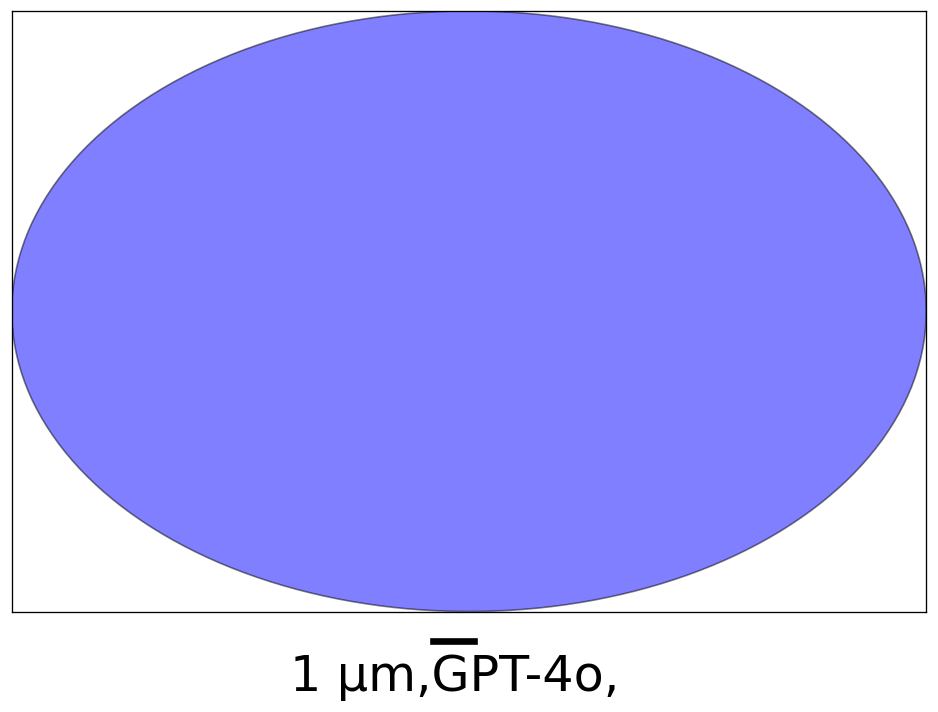} & \includegraphics[width=0.13\textwidth]{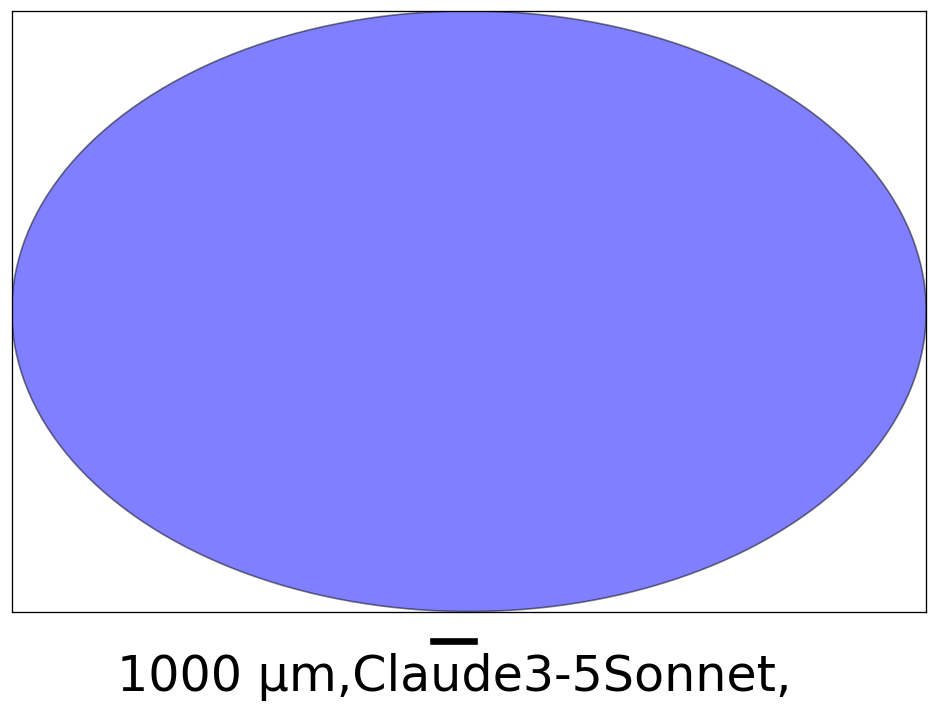} & \includegraphics[width=0.13\textwidth]{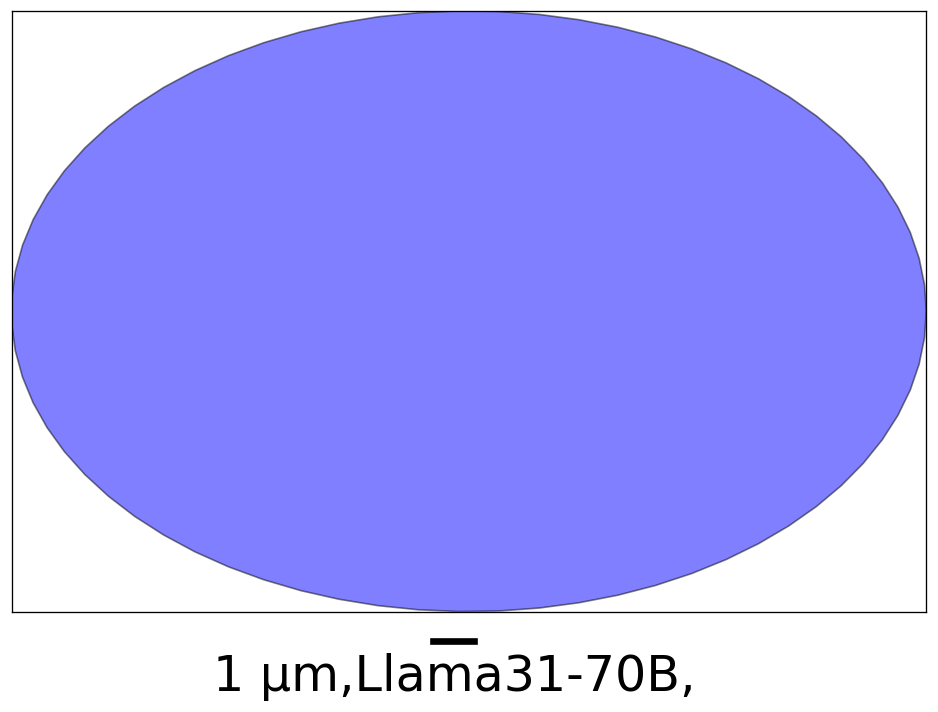} & \includegraphics[width=0.13\textwidth]{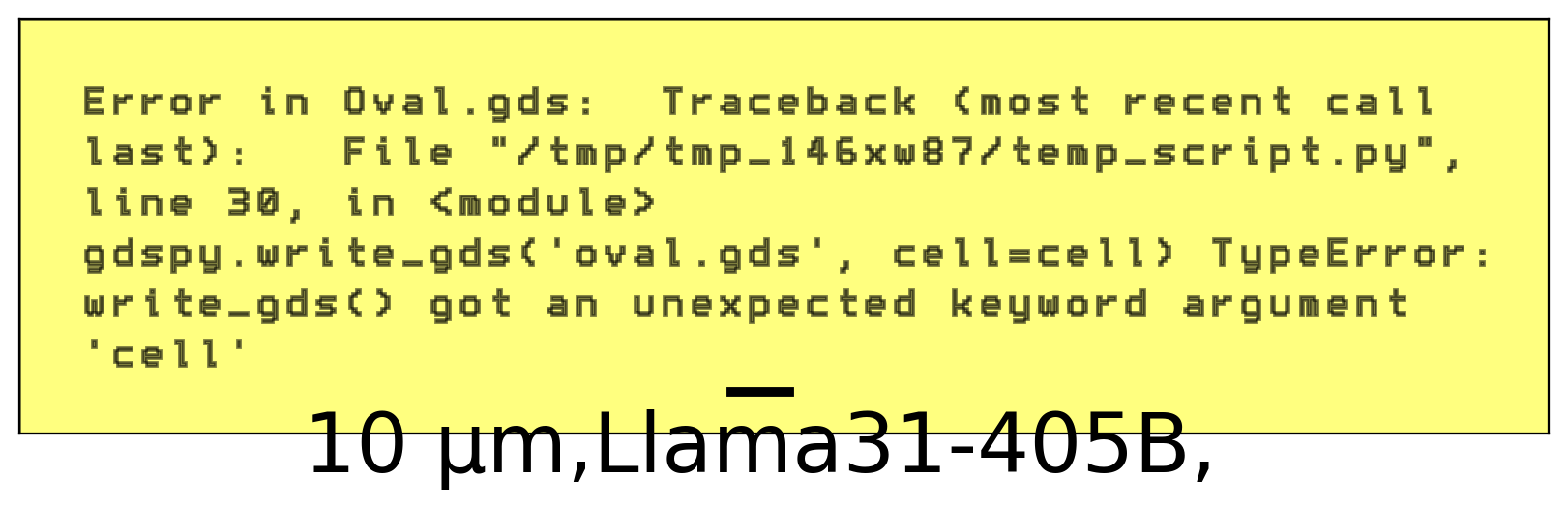} & \includegraphics[width=0.13\textwidth]{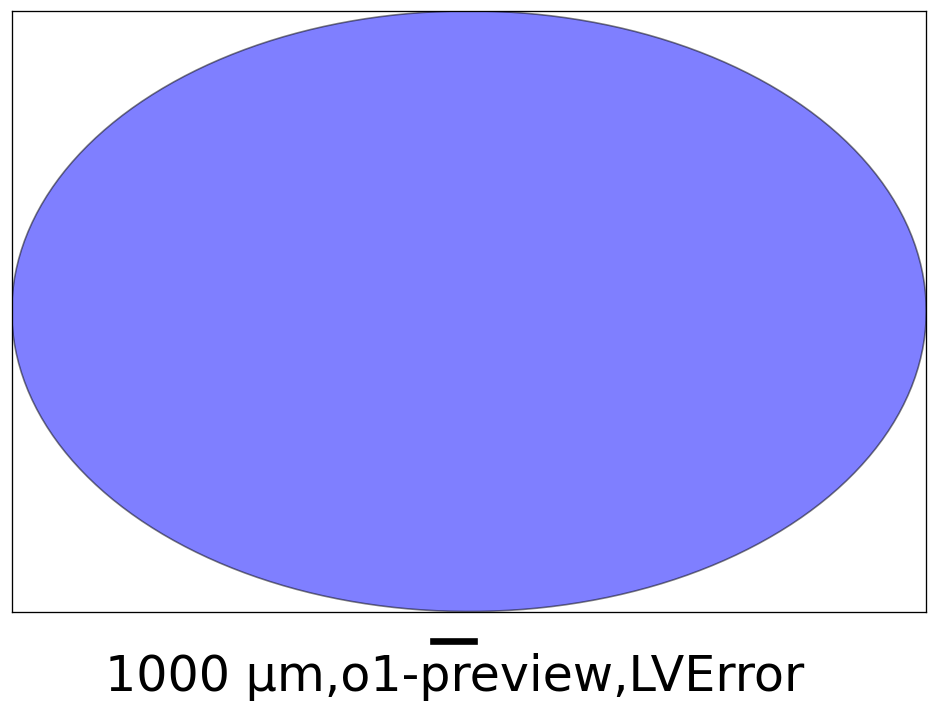} \\
    \begin{tabular}{@{}c@{}}Single LLM \\ Baseline \\ Run 4\end{tabular} & \includegraphics[width=0.13\textwidth]{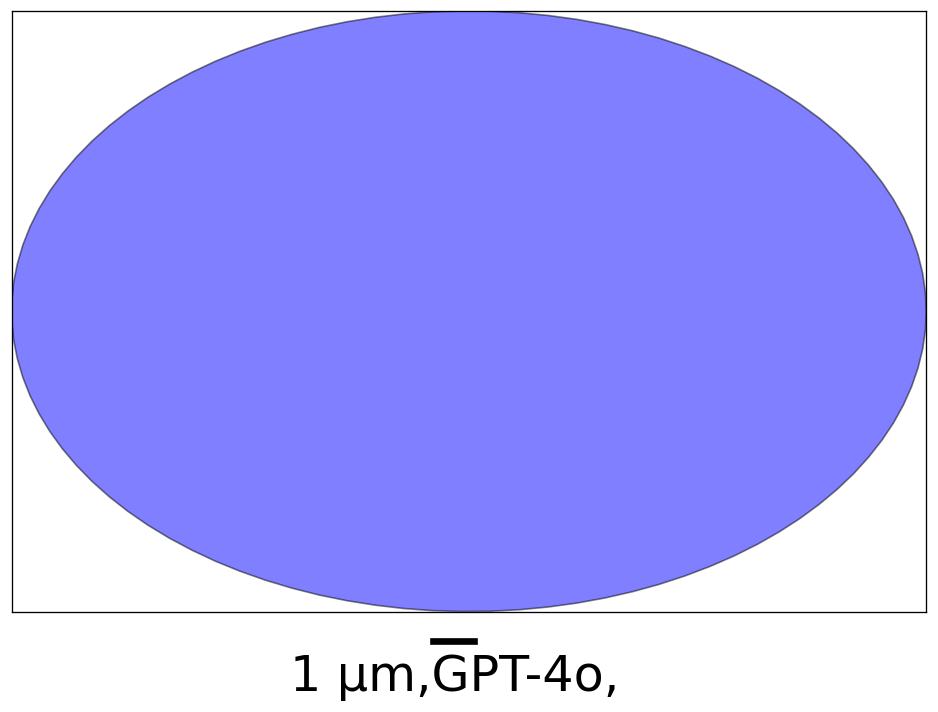} & \includegraphics[width=0.13\textwidth]{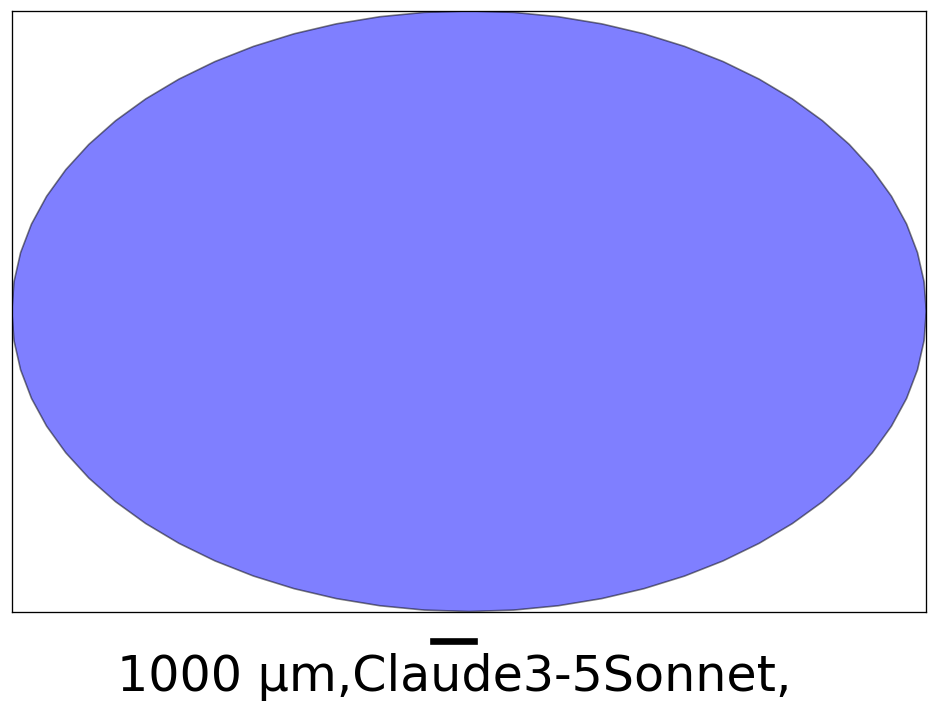} & \includegraphics[width=0.13\textwidth]{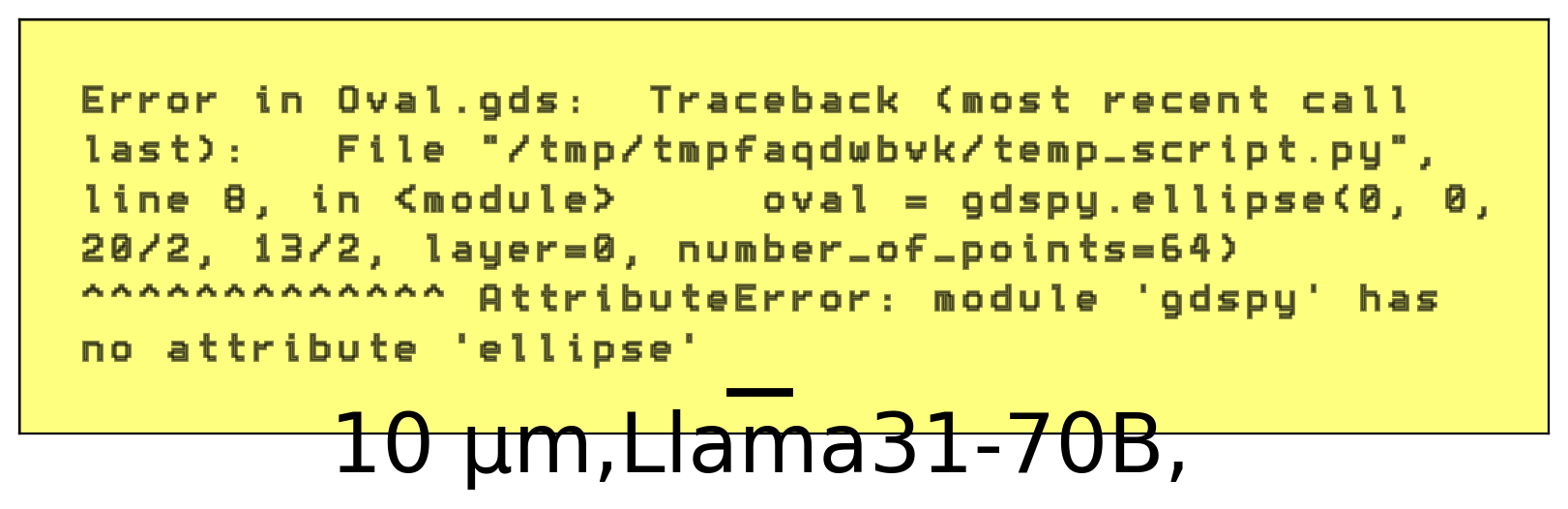} & \includegraphics[width=0.13\textwidth]{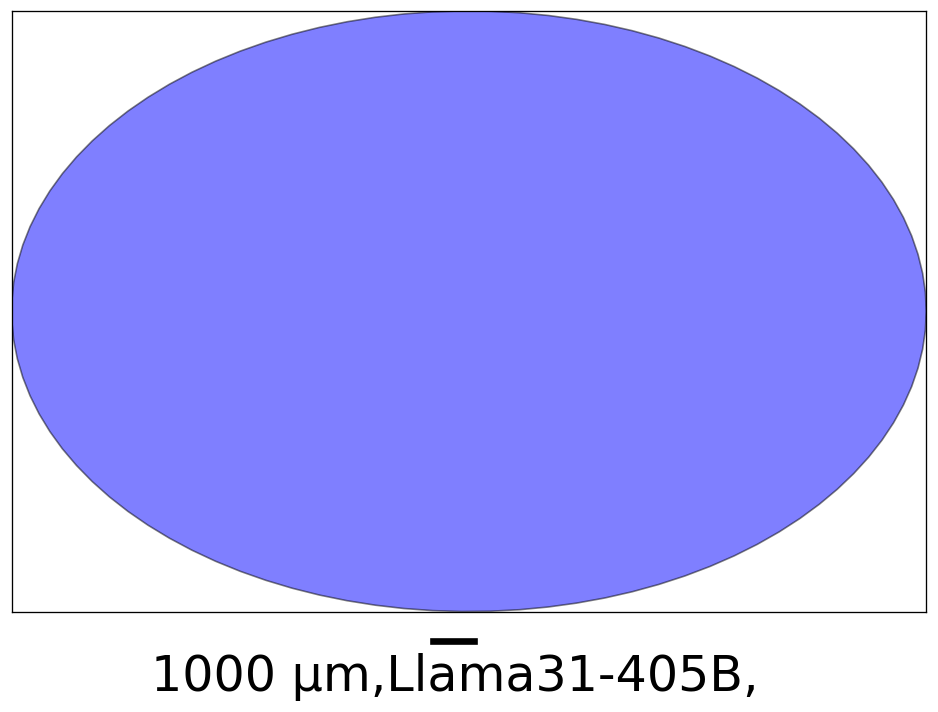} & \includegraphics[width=0.13\textwidth]{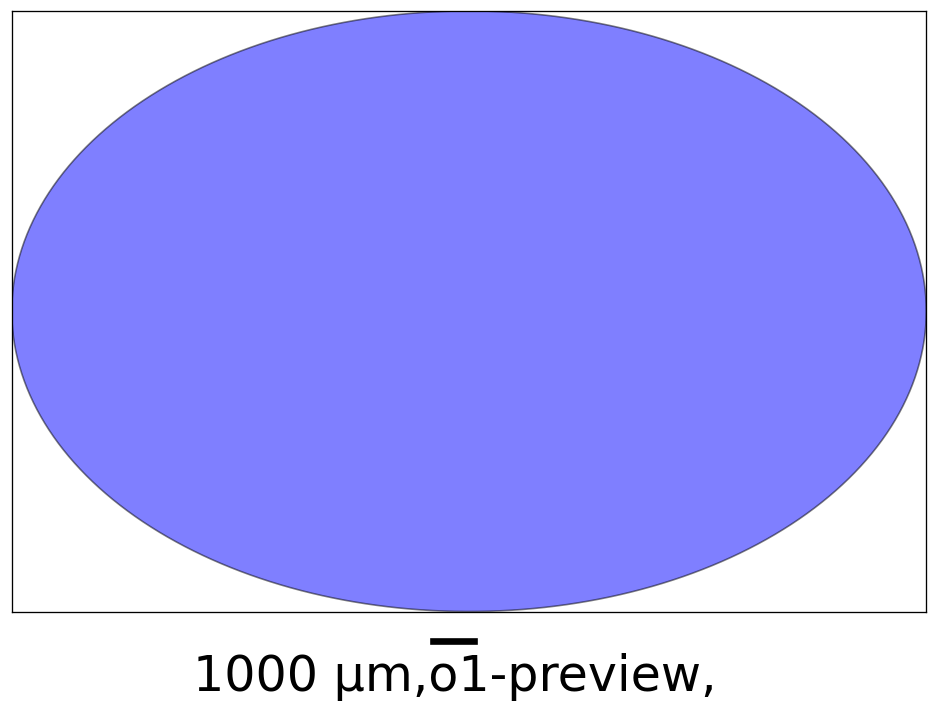} \\
    \begin{tabular}{@{}c@{}}Single LLM \\ Baseline \\ Run 5\end{tabular} & \includegraphics[width=0.13\textwidth]{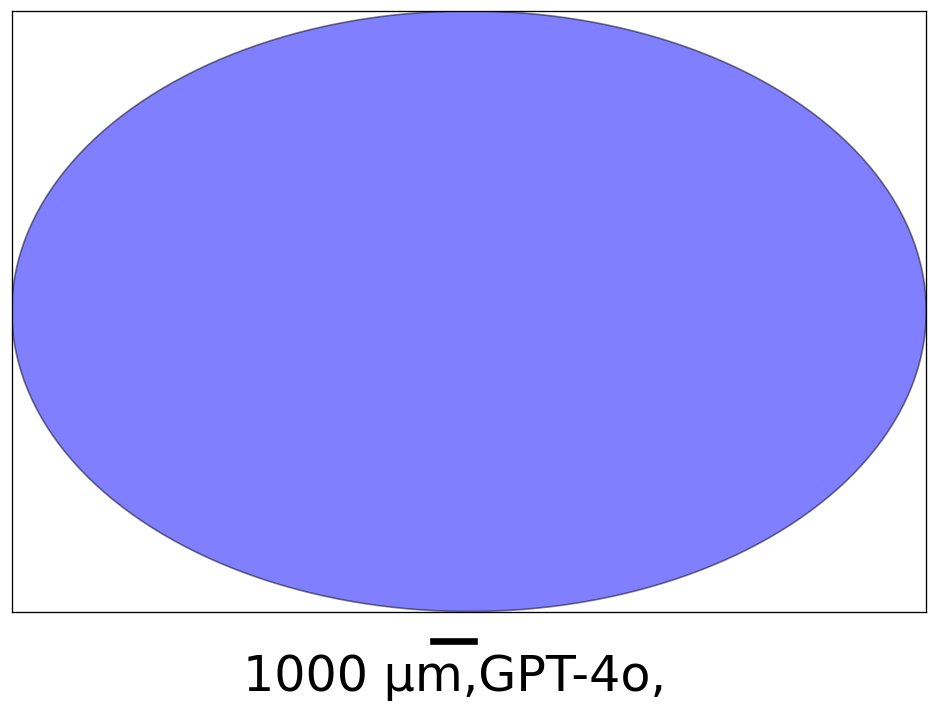} & \includegraphics[width=0.13\textwidth]{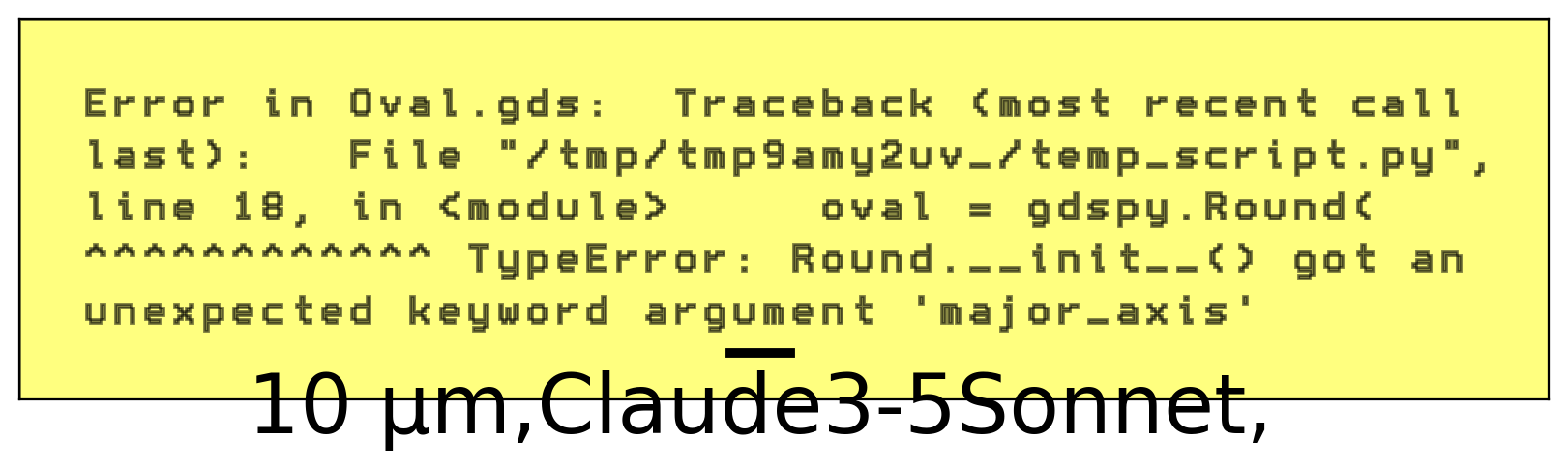} & \includegraphics[width=0.13\textwidth]{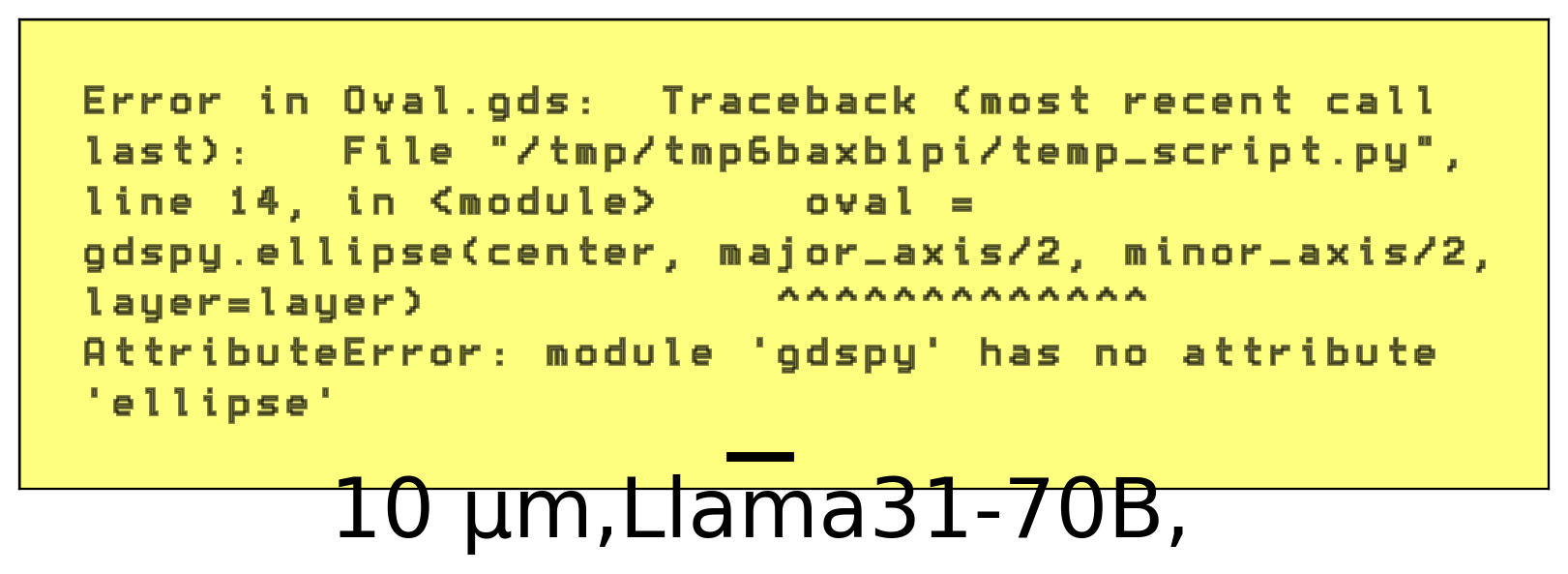} & \includegraphics[width=0.13\textwidth]{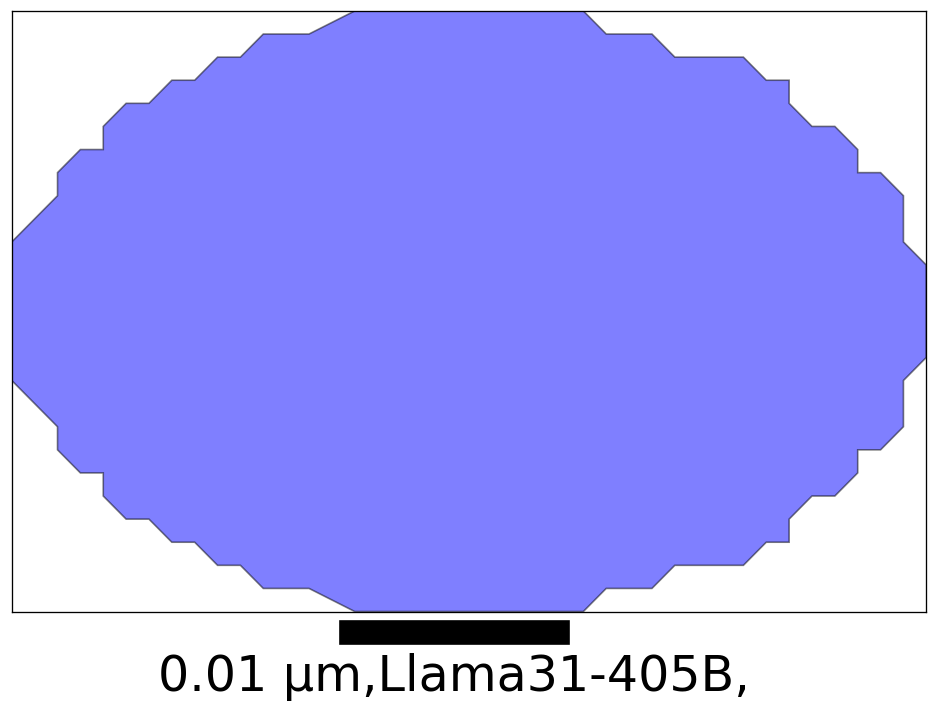} & \includegraphics[width=0.13\textwidth]{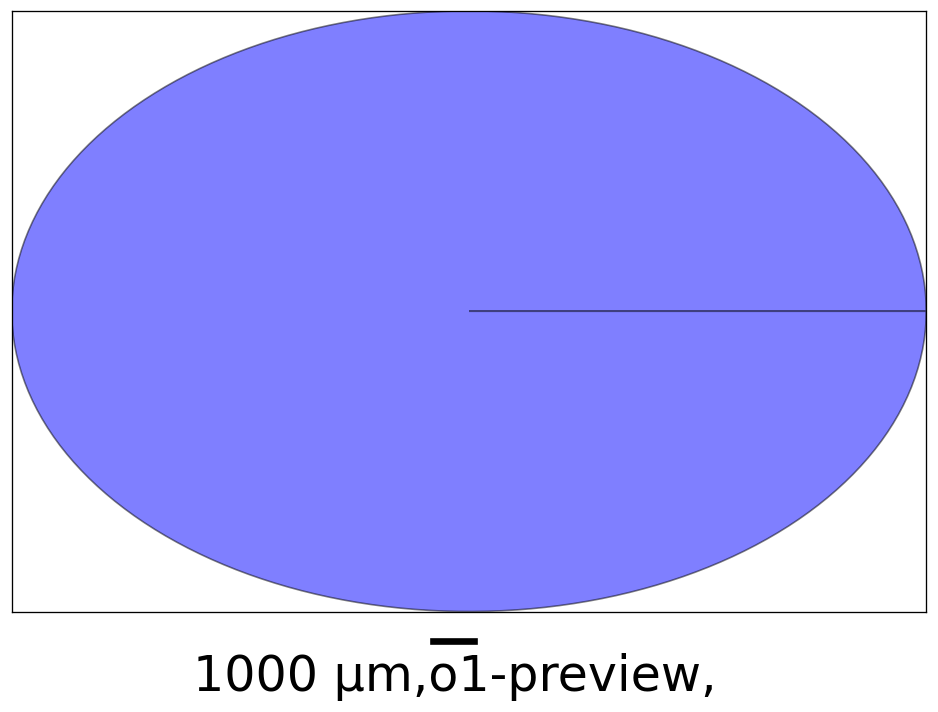} \\
    \bottomrule
  \end{tabularx}
\end{table}

\clearpage
\begin{table}[p]
  \caption{Arrow Task Question: Generate an Arrow pointing to the right with length 10 mm, make the body 1/3 width of the head, start at 0,0.}
  \label{table:arrow}
  \centering
  \begin{tabularx}{0.9\textwidth}{@{}XXXXXX@{}}
    \toprule
    \begin{tabular}{@{}c@{}}Ground Truth \\ \includegraphics[width=0.13\textwidth]{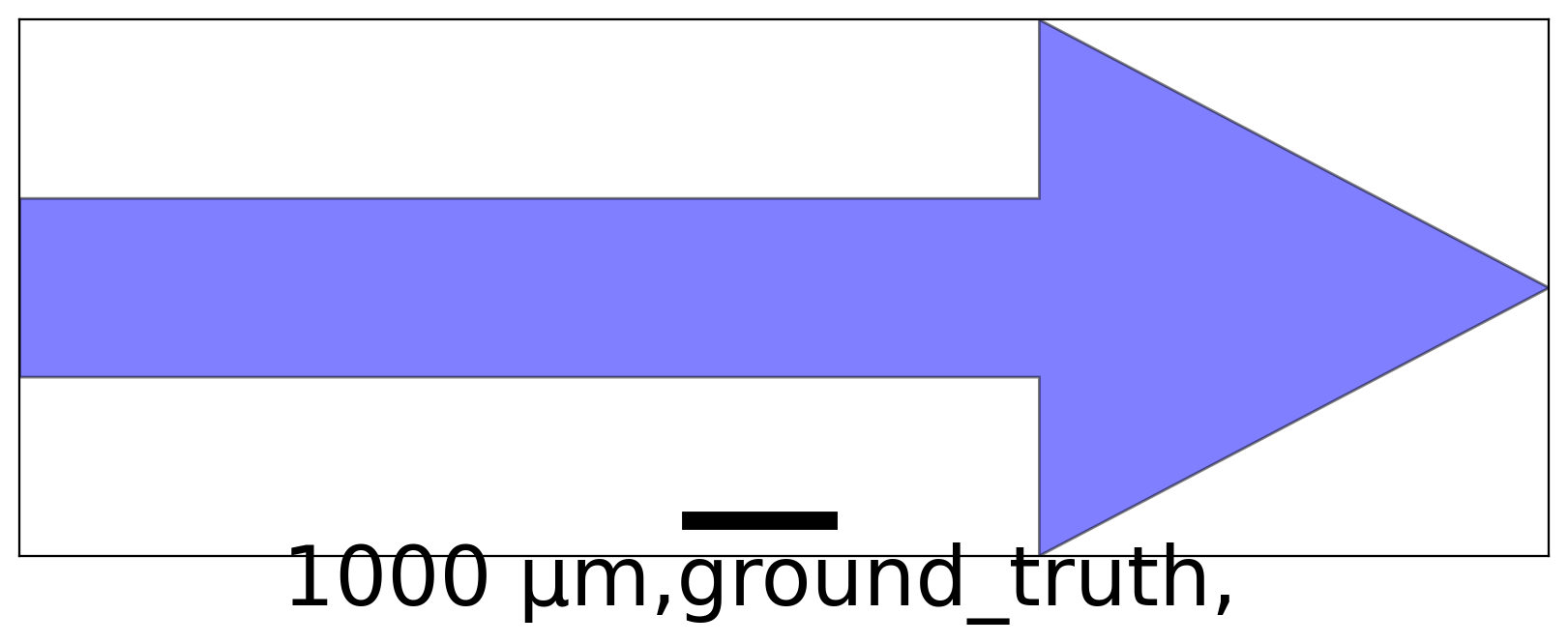}\end{tabular} & GPT-4o & Claude-3.5 & Llama-3-70B & Llama-3-405B & o1-preview \\
    \midrule
    SOLOMON & \includegraphics[width=0.13\textwidth]{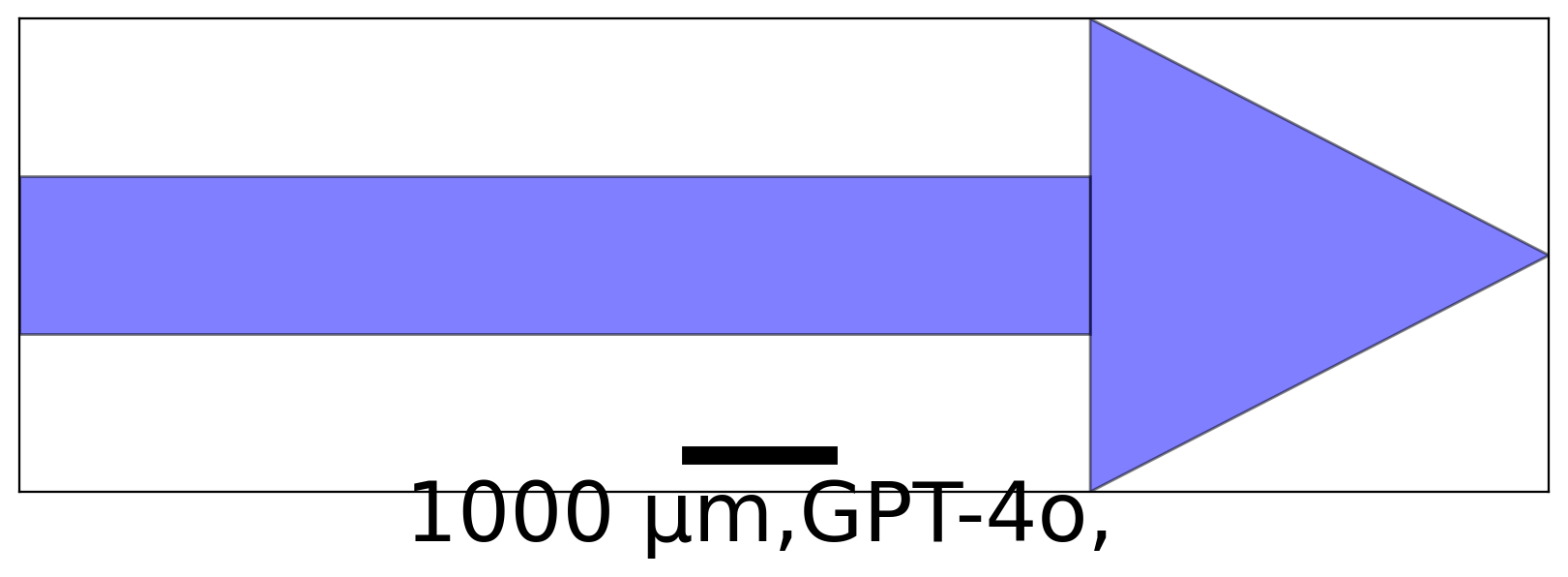} & \includegraphics[width=0.13\textwidth]{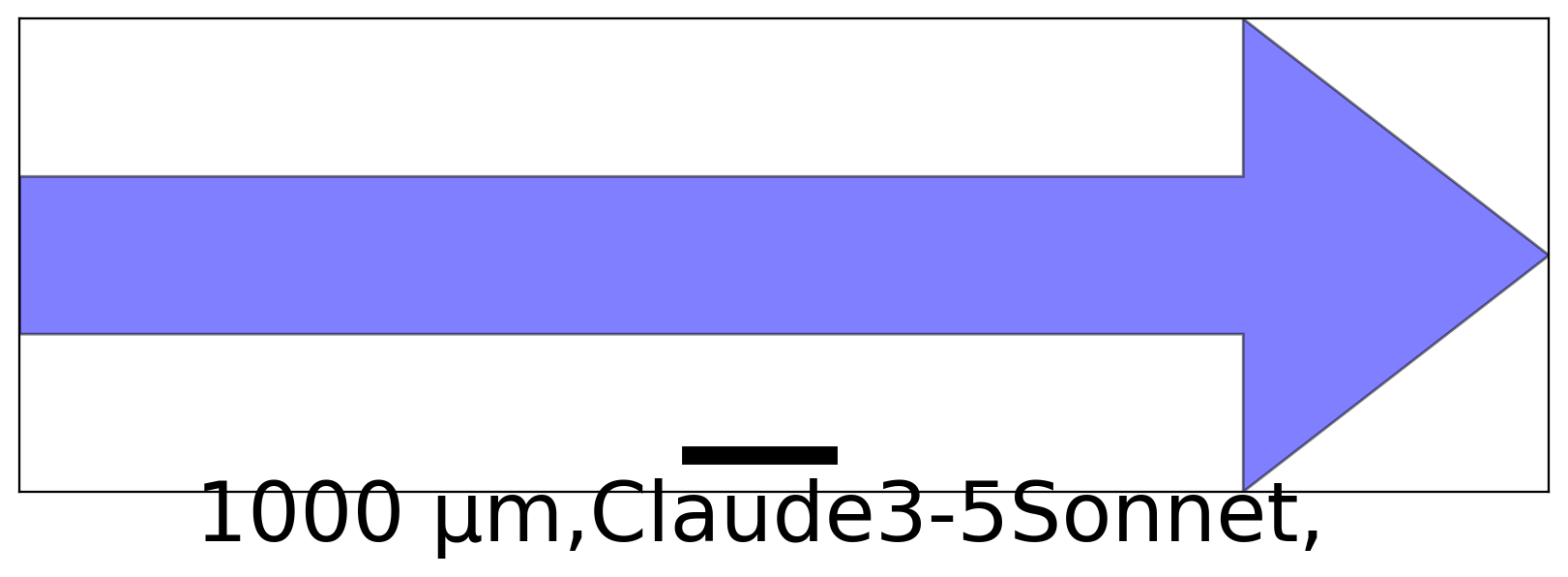} & \includegraphics[width=0.13\textwidth]{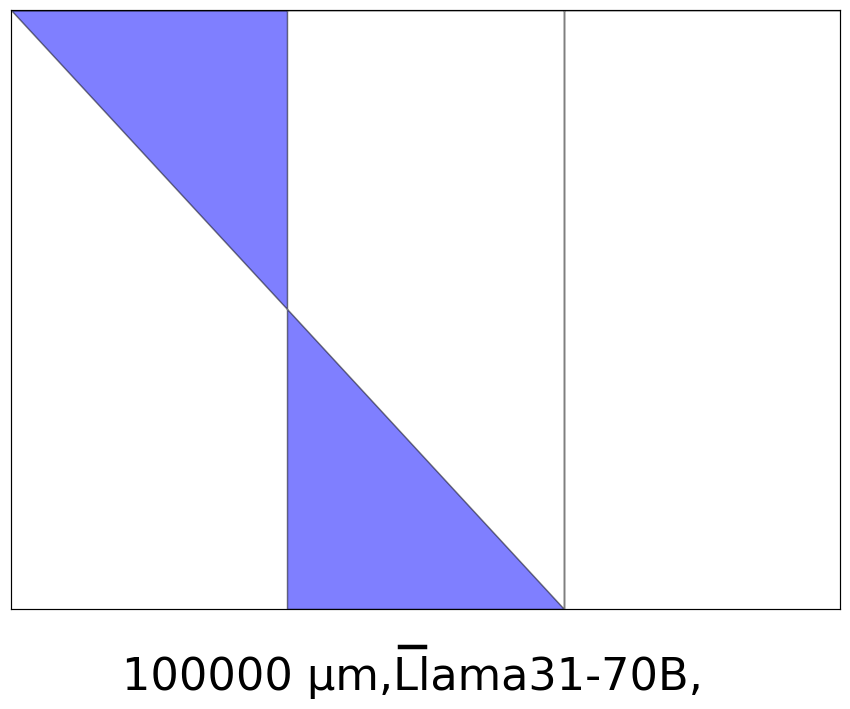} & \includegraphics[width=0.13\textwidth]{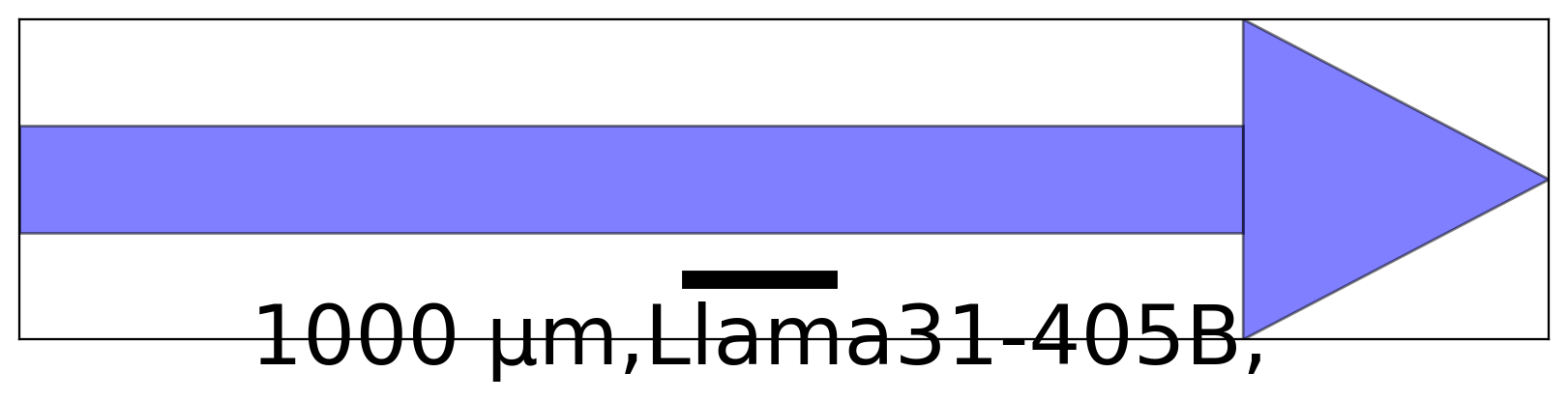} &  \\
    \begin{tabular}{@{}c@{}}Single LLM \\ Baseline \\ Run 1\end{tabular} & \includegraphics[width=0.13\textwidth]{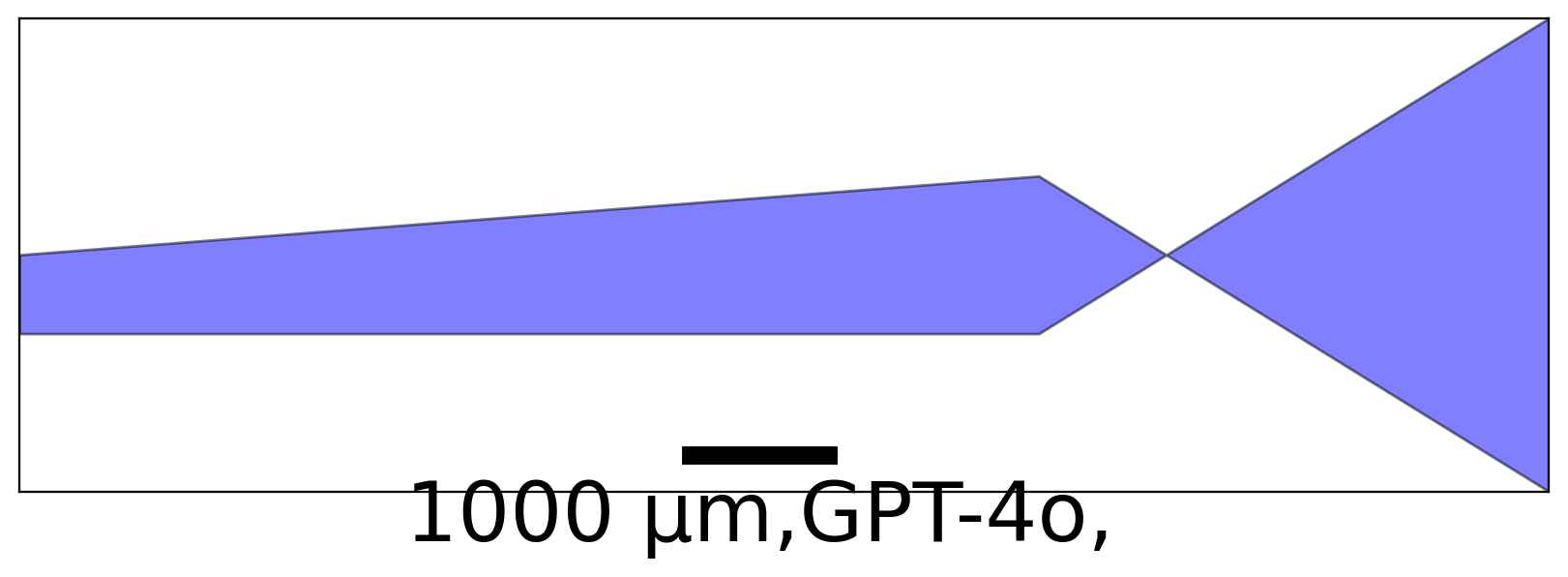} & \includegraphics[width=0.13\textwidth]{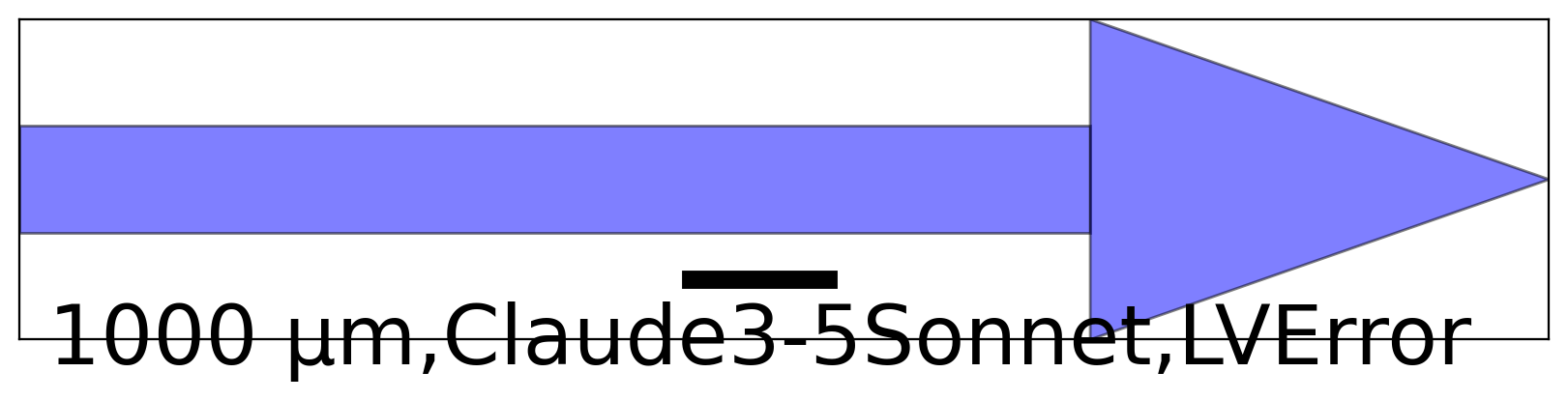} & \includegraphics[width=0.13\textwidth]{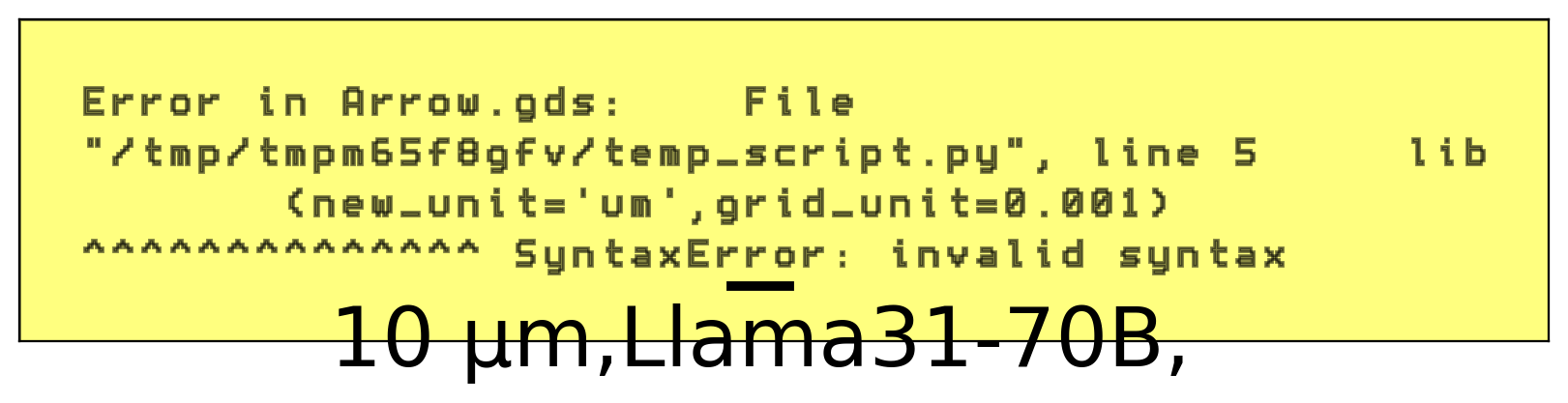} & \includegraphics[width=0.13\textwidth]{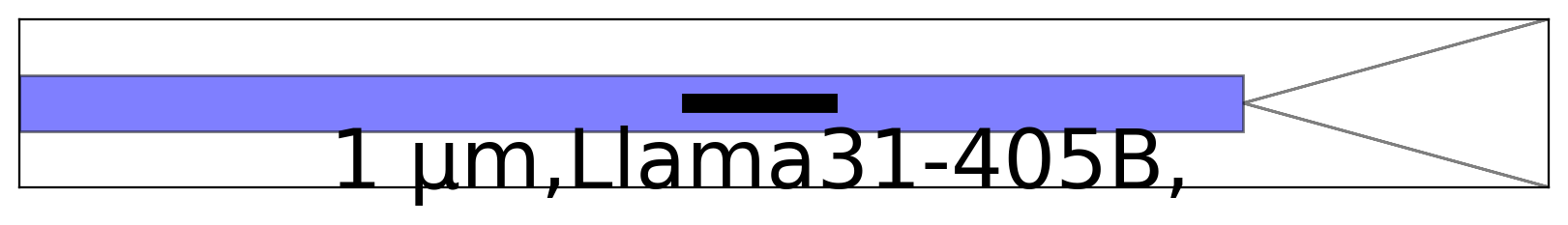} & \includegraphics[width=0.13\textwidth]{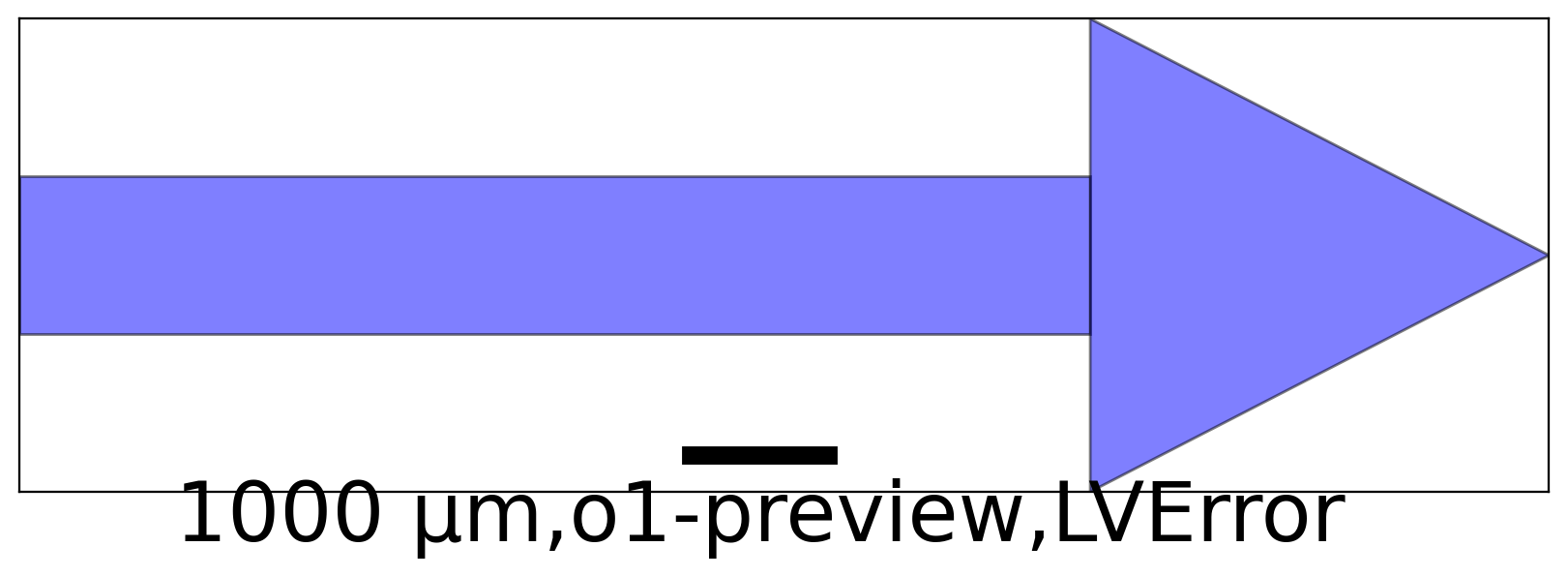} \\
    \begin{tabular}{@{}c@{}}Single LLM \\ Baseline \\ Run 2\end{tabular} & \includegraphics[width=0.13\textwidth]{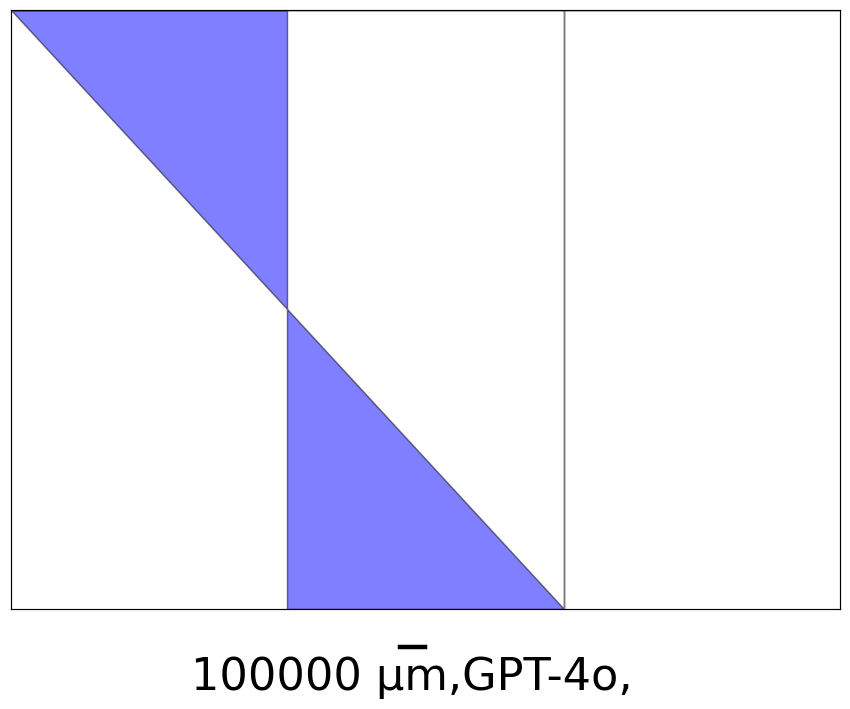} & \includegraphics[width=0.13\textwidth]{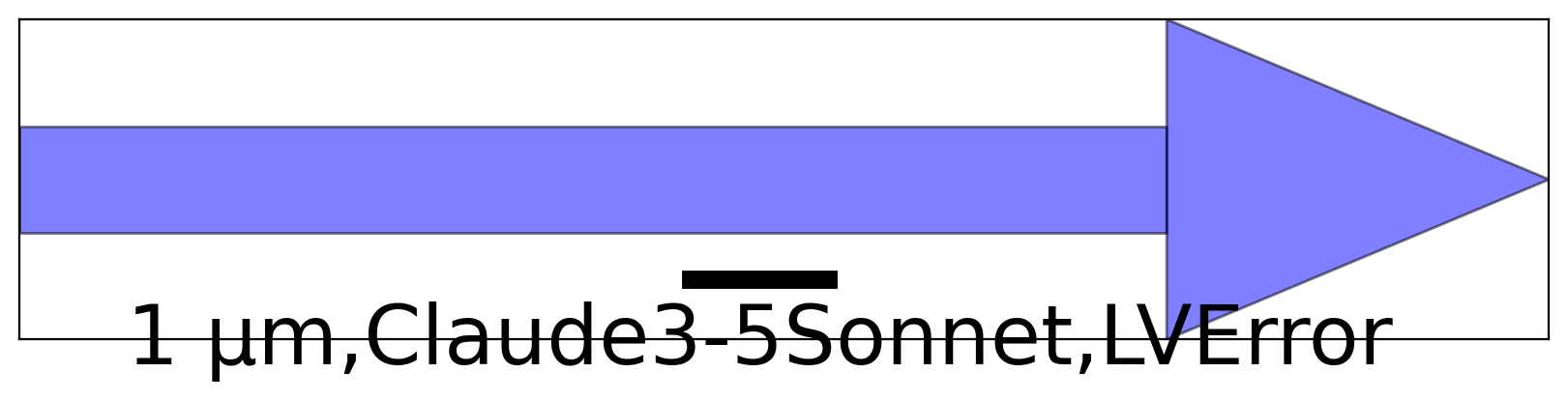} & \includegraphics[width=0.13\textwidth]{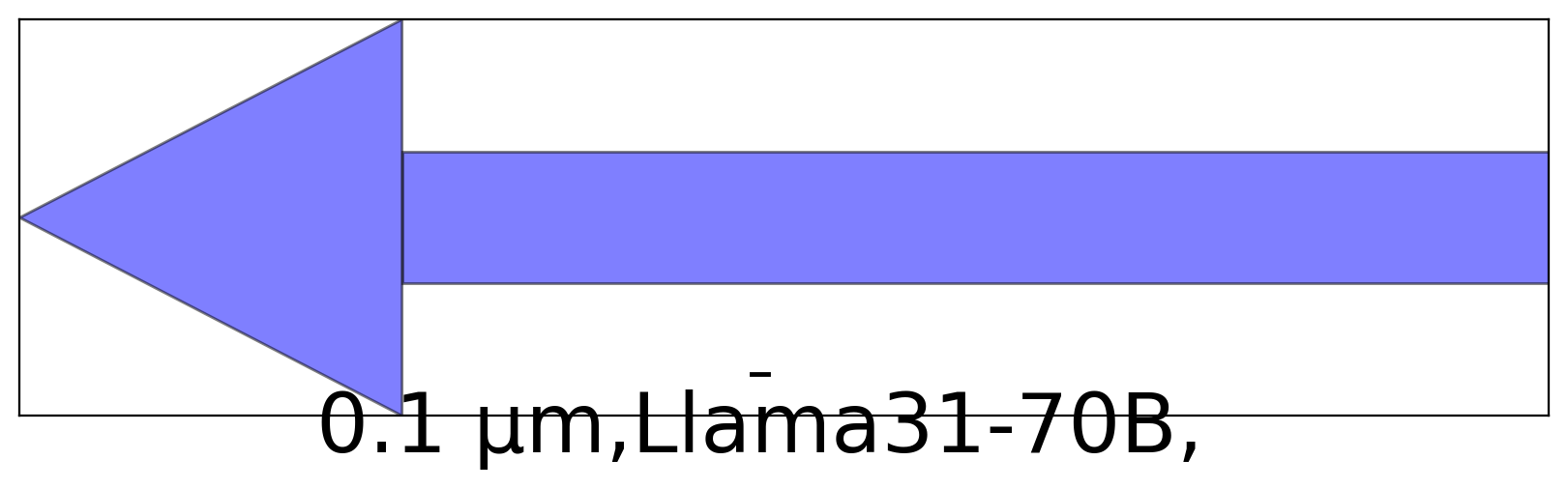} & \includegraphics[width=0.13\textwidth]{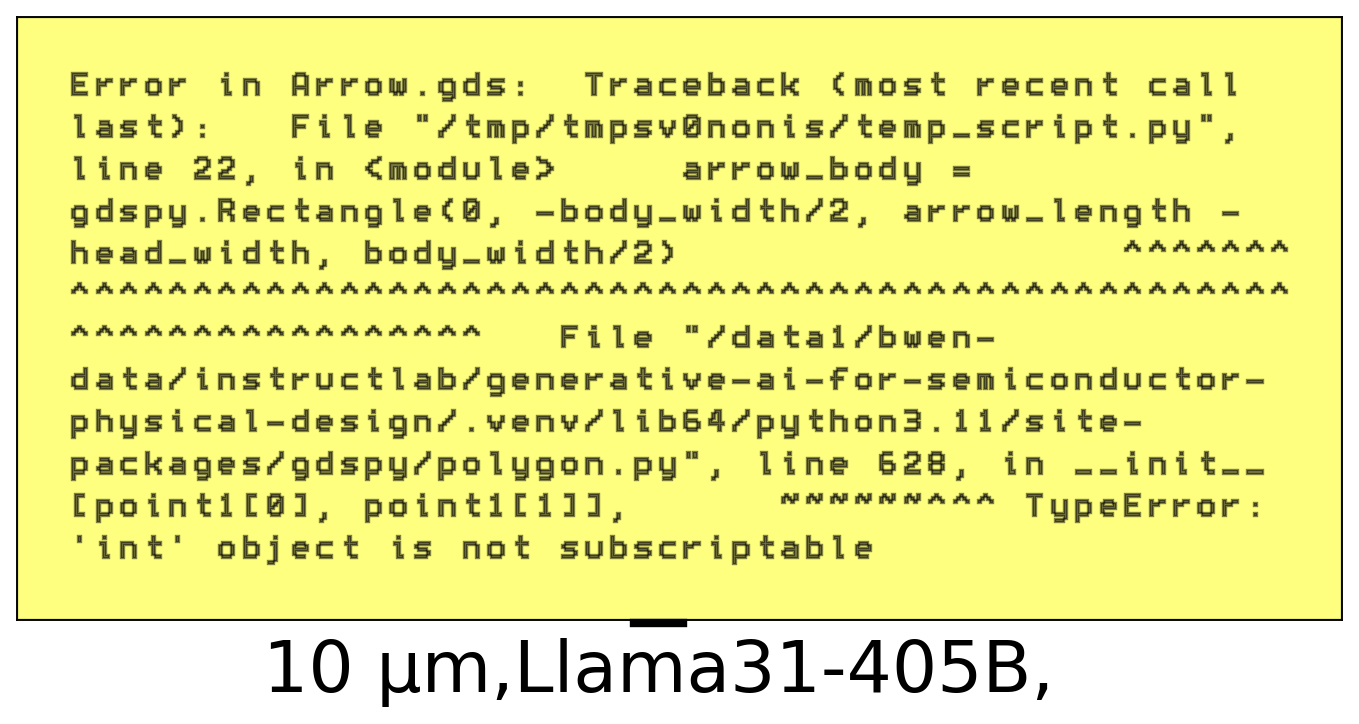} & \includegraphics[width=0.13\textwidth]{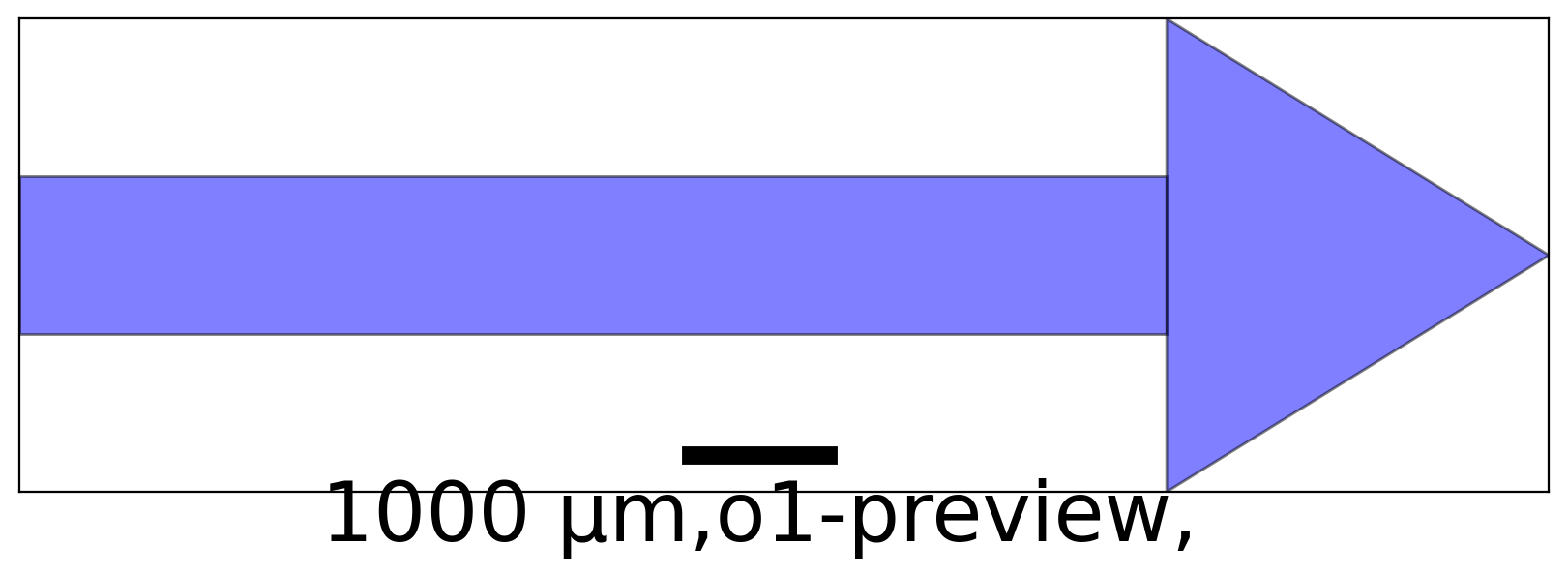} \\
    \begin{tabular}{@{}c@{}}Single LLM \\ Baseline \\ Run 3\end{tabular} & \includegraphics[width=0.13\textwidth]{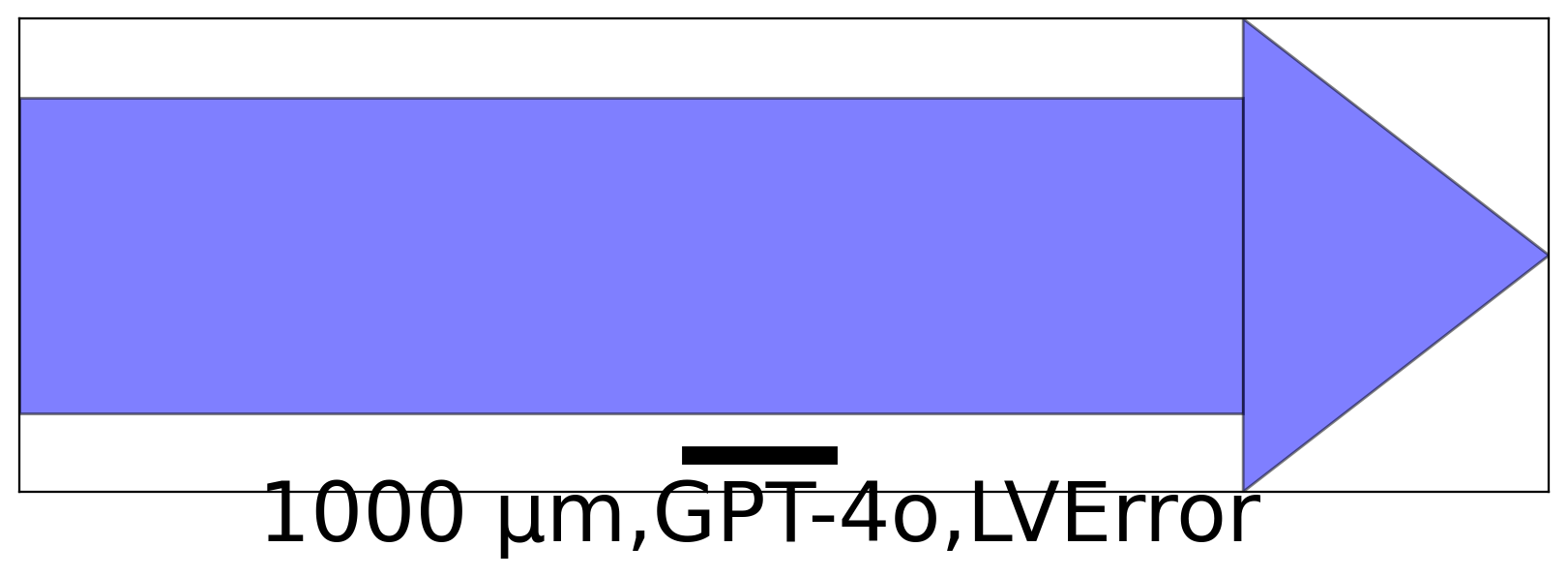} & \includegraphics[width=0.13\textwidth]{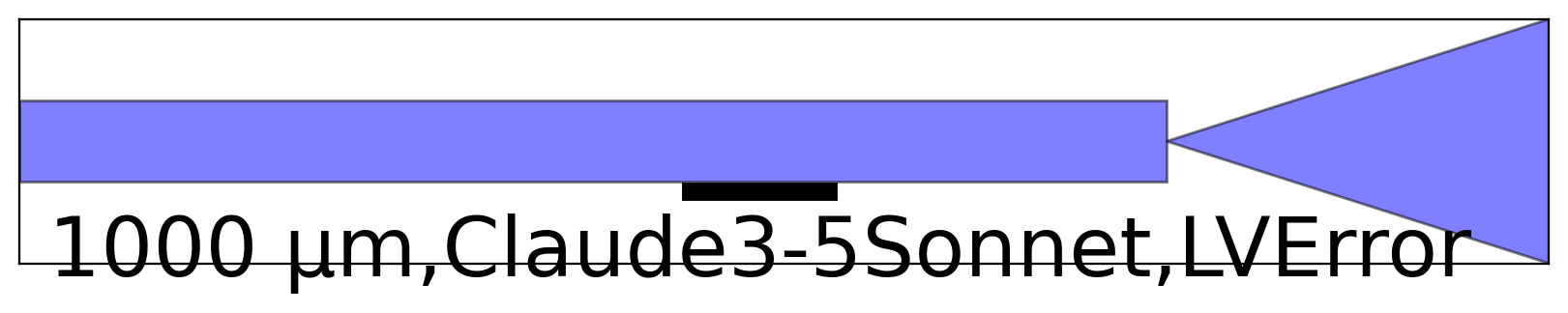} & \includegraphics[width=0.13\textwidth]{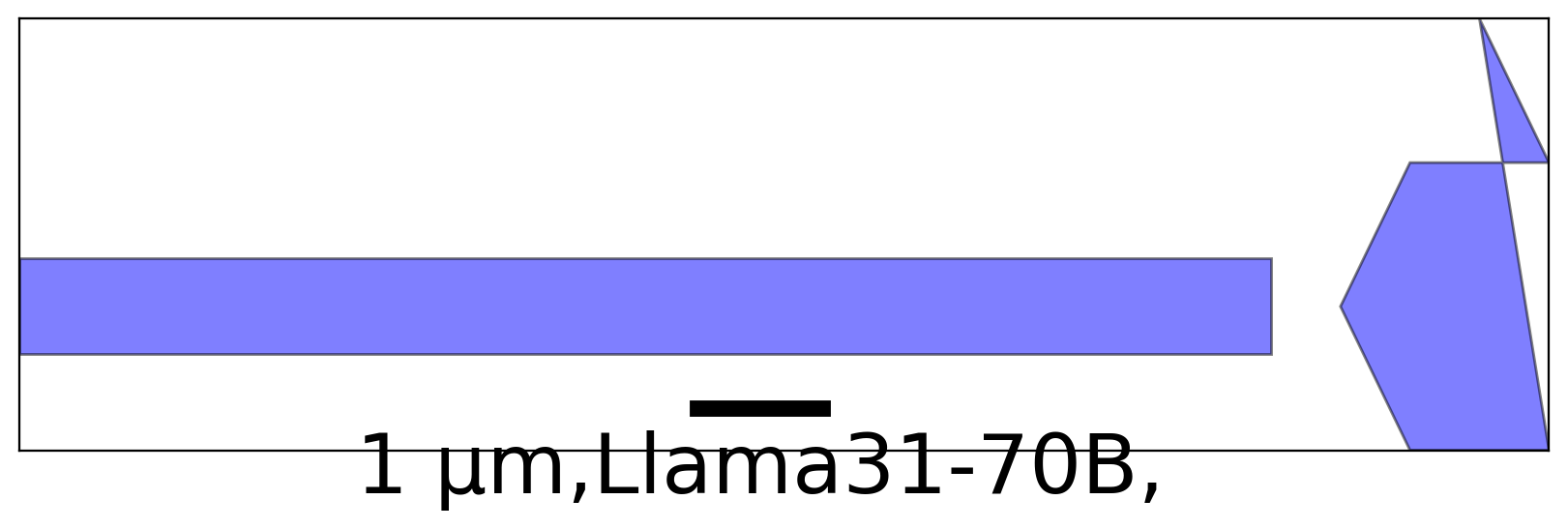} & \includegraphics[width=0.13\textwidth]{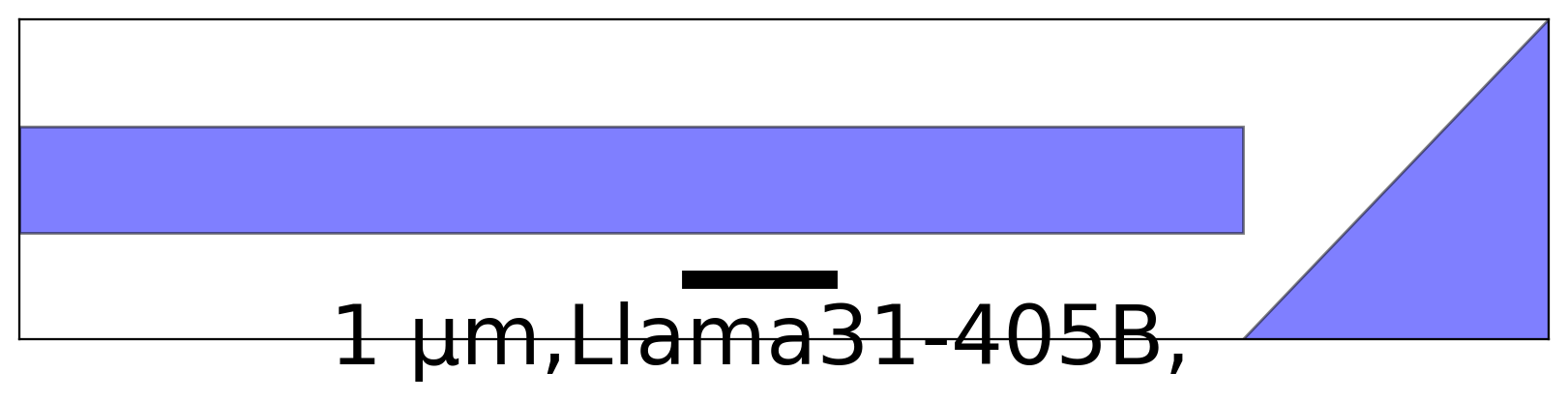} & \includegraphics[width=0.13\textwidth]{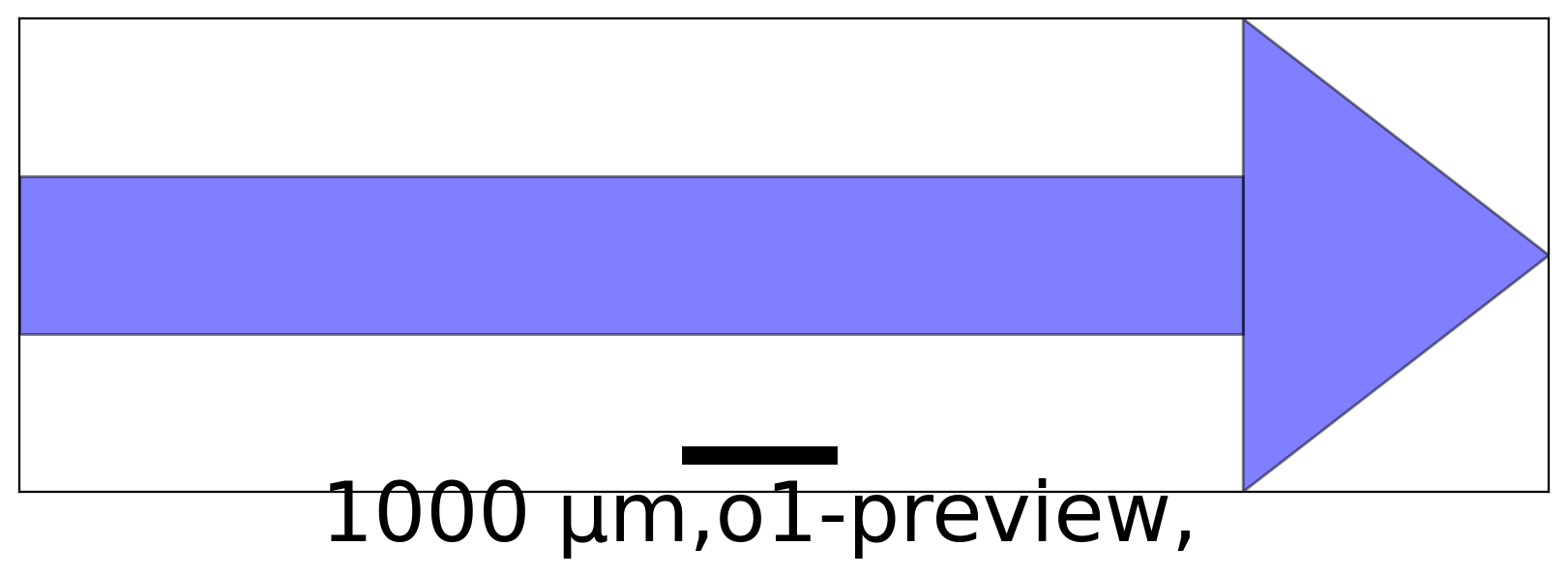} \\
    \begin{tabular}{@{}c@{}}Single LLM \\ Baseline \\ Run 4\end{tabular} & \includegraphics[width=0.13\textwidth]{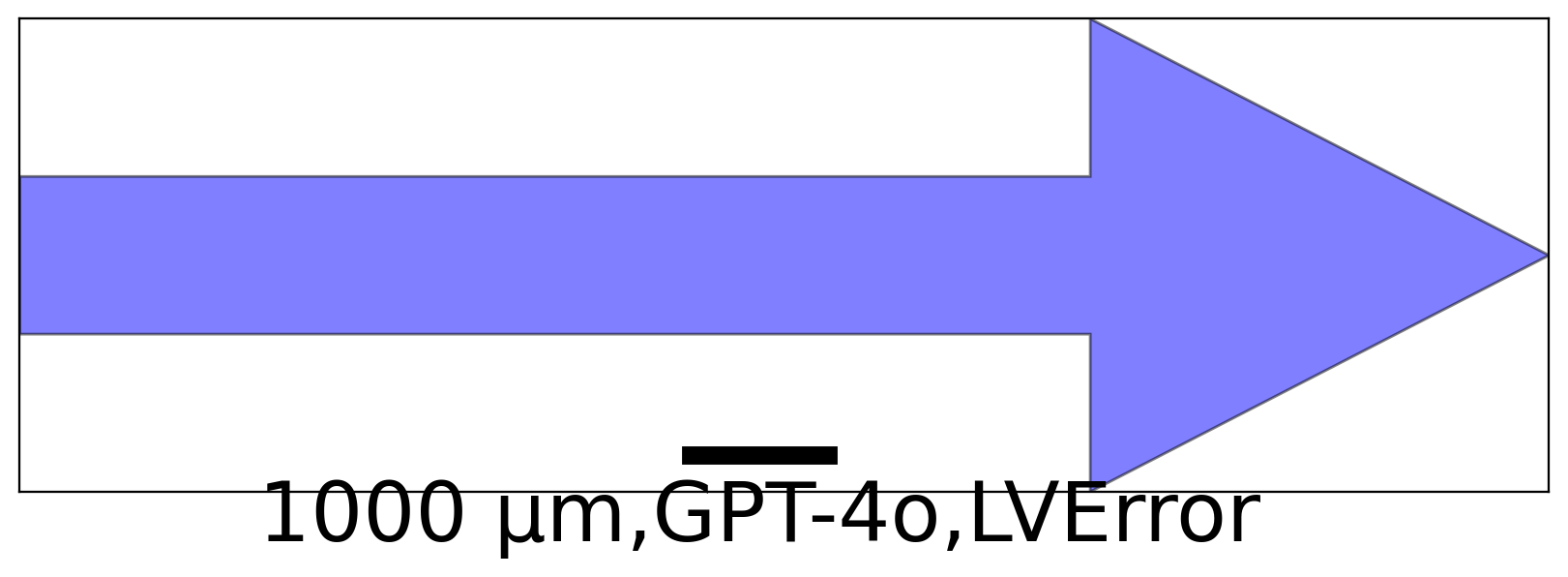} & \includegraphics[width=0.13\textwidth]{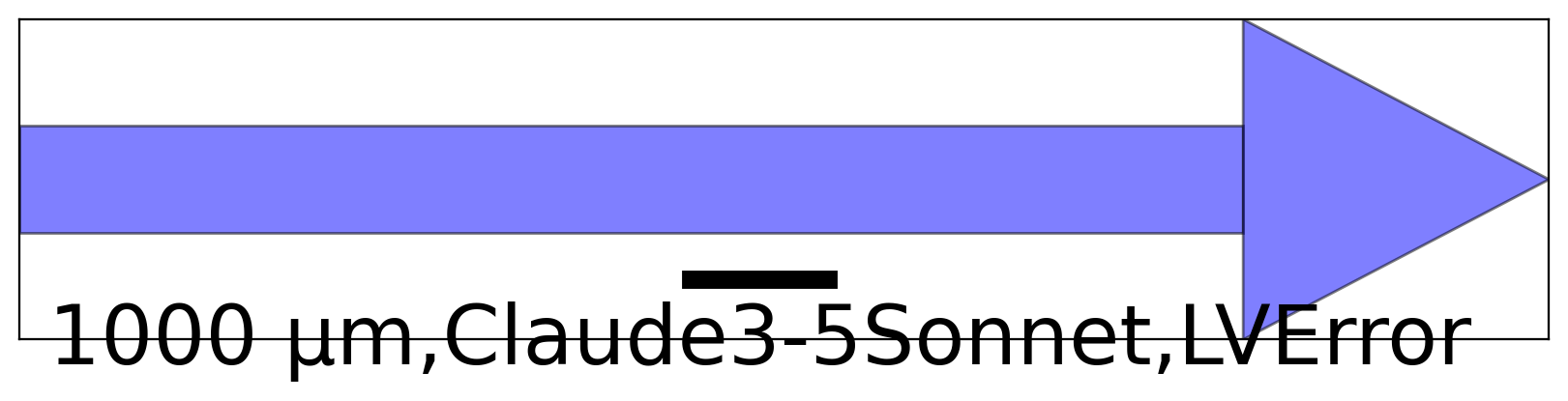} & \includegraphics[width=0.13\textwidth]{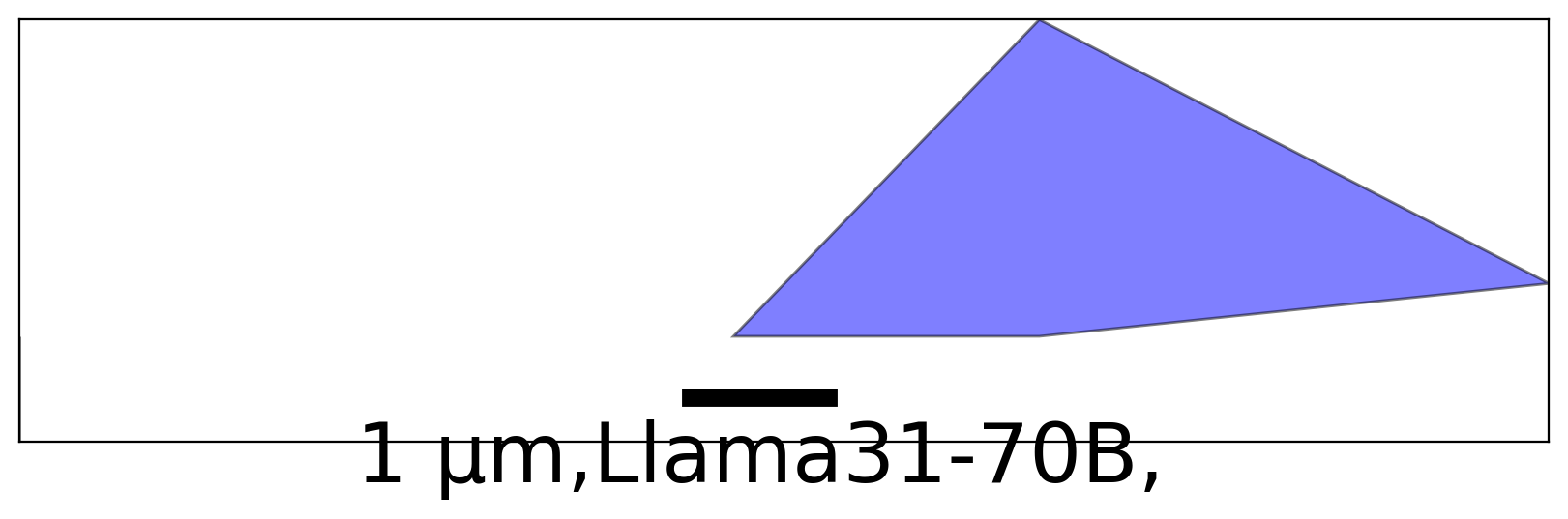} & \includegraphics[width=0.13\textwidth]{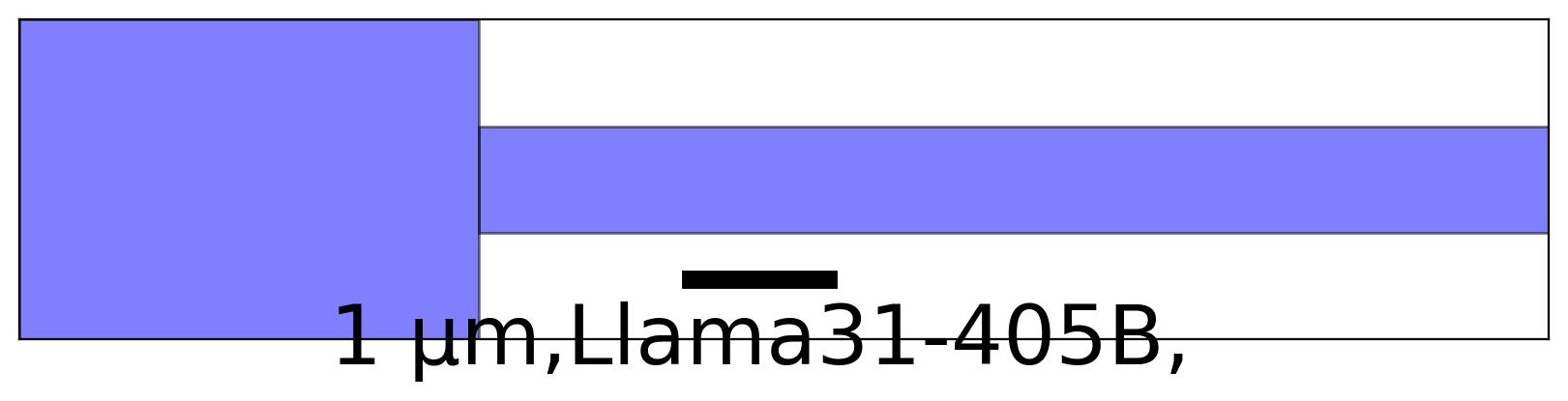} & \includegraphics[width=0.13\textwidth]{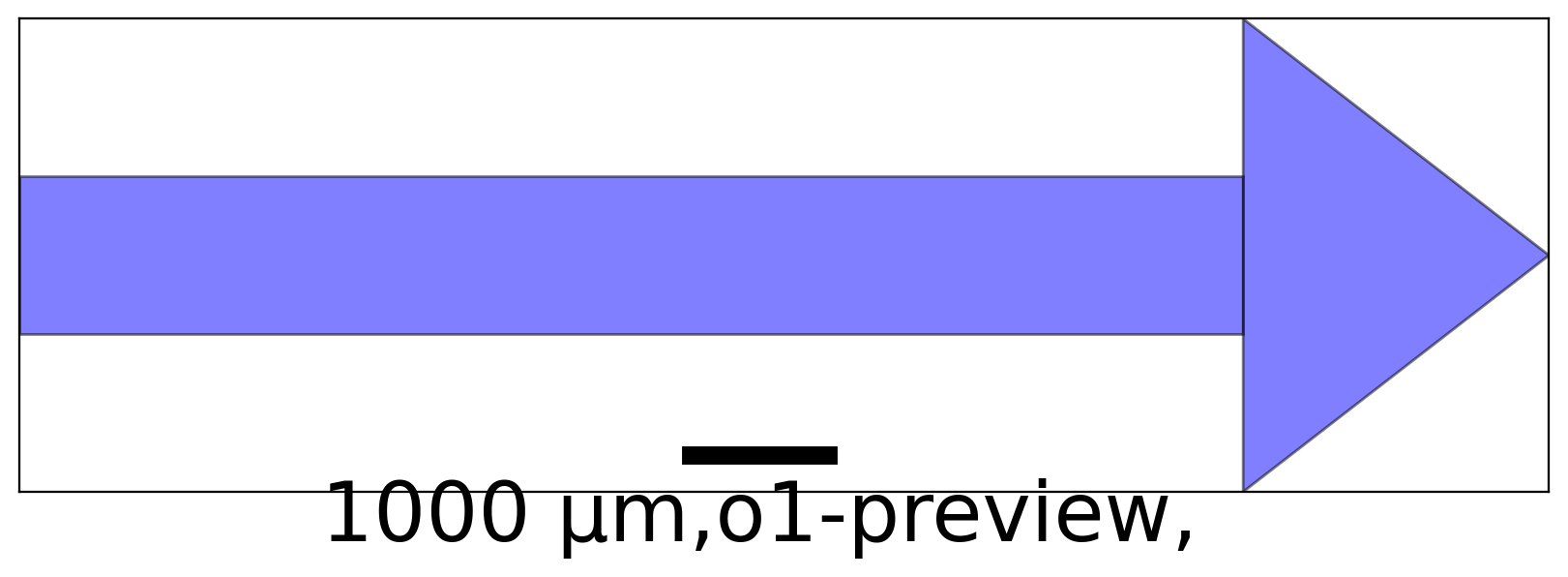} \\
    \begin{tabular}{@{}c@{}}Single LLM \\ Baseline \\ Run 5\end{tabular} & \includegraphics[width=0.13\textwidth]{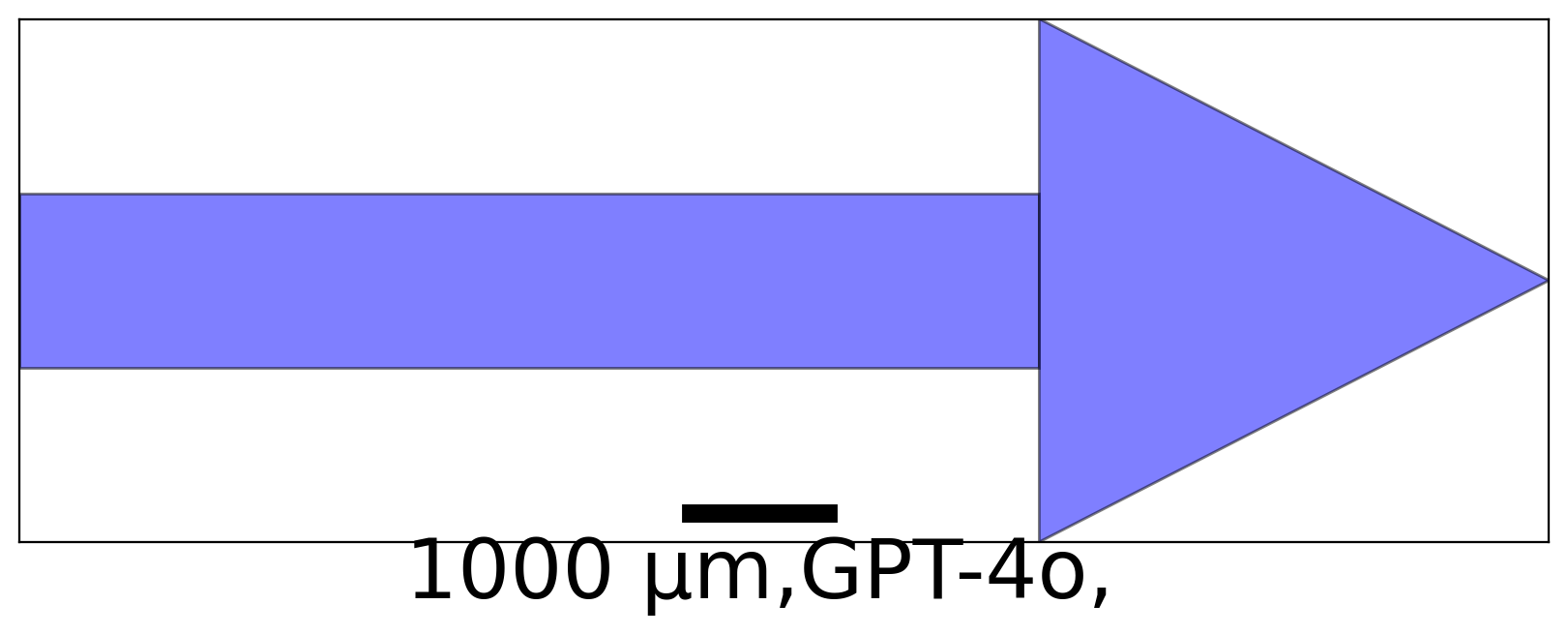} & \includegraphics[width=0.13\textwidth]{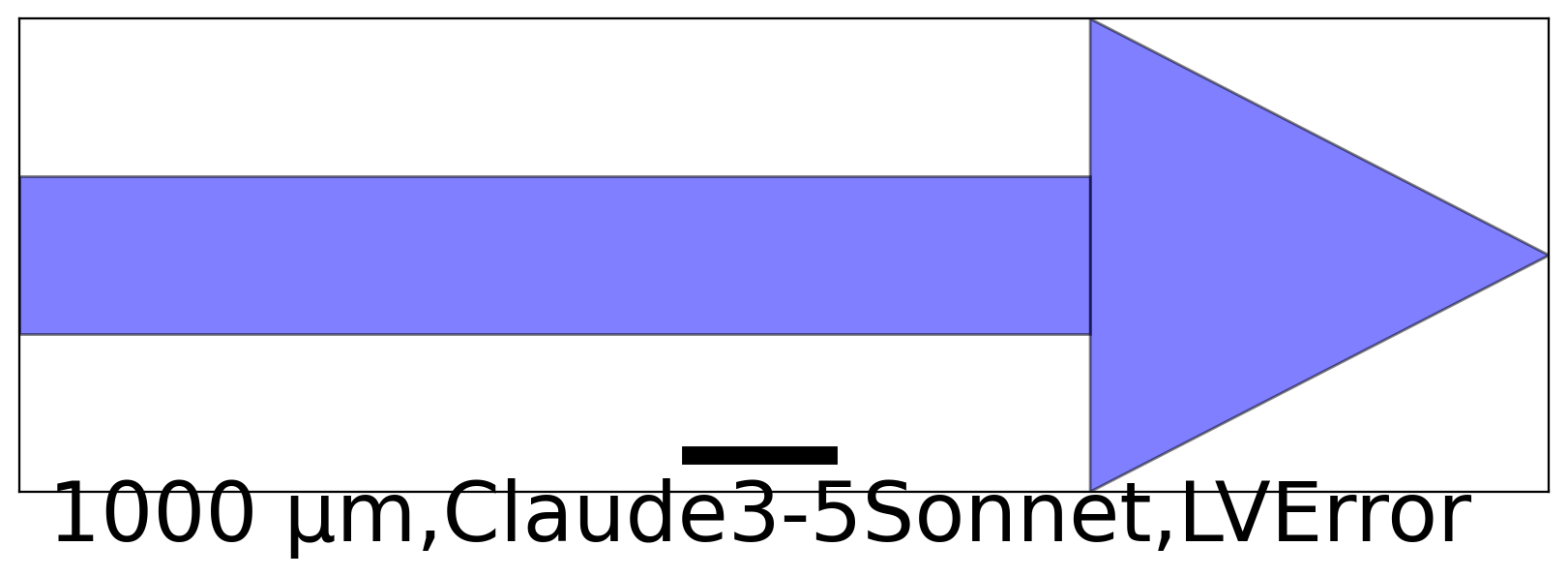} & \includegraphics[width=0.13\textwidth]{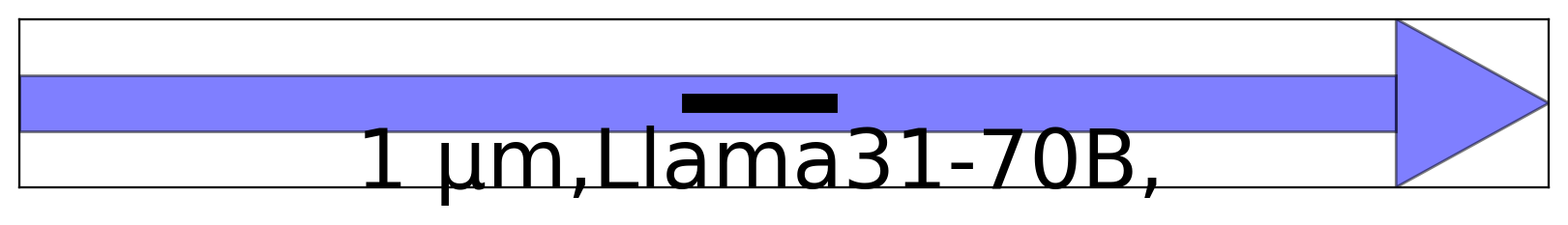} & \includegraphics[width=0.13\textwidth]{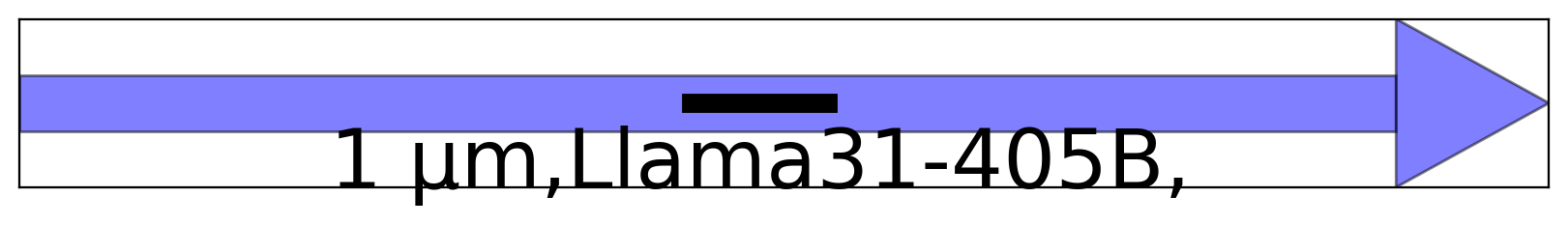} & \includegraphics[width=0.13\textwidth]{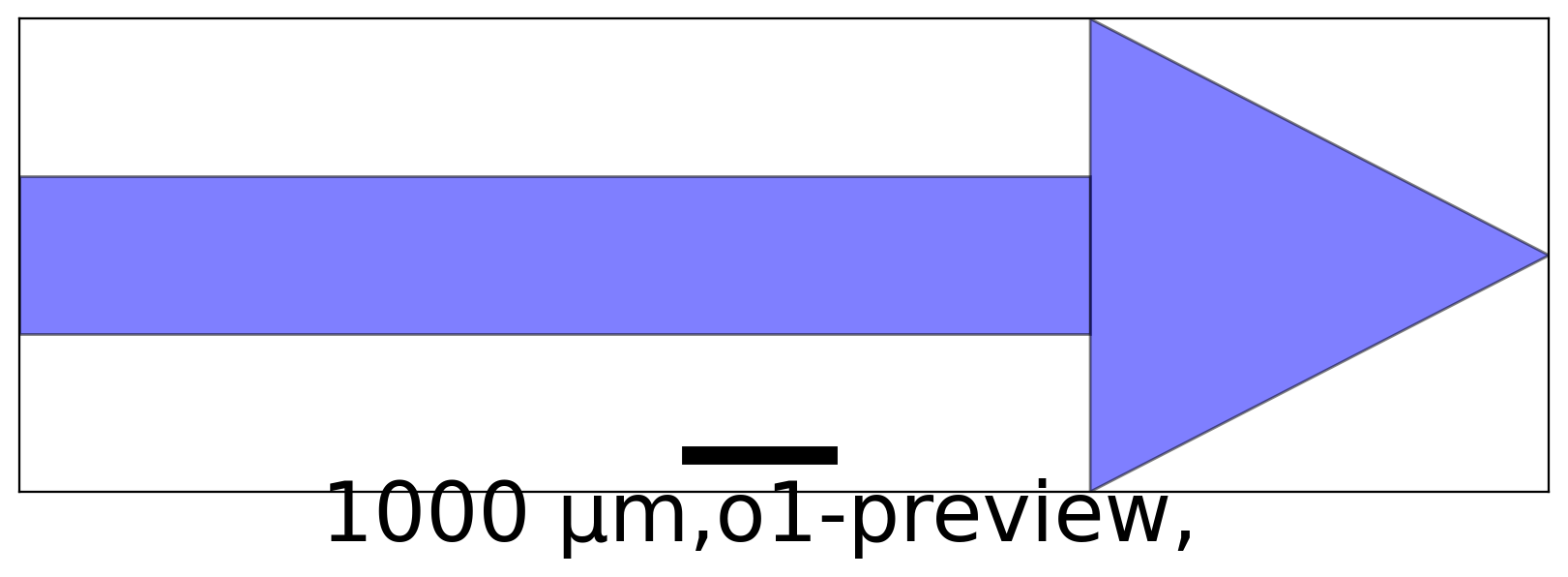} \\
    \bottomrule
  \end{tabularx}
\end{table}

\clearpage
\begin{table}[p]
  \caption{BasicLayout Task Question: 1. Draw a rectangular active region with dimensions 10 µm x 5 µm.
2. Place a polysilicon gate that crosses the active region vertically at its center, with a width of 1 µm.
3. Add two square contact holes, each 1 µm x 1 µm, positioned 1 µm away from the gate on either side along the active region.}
  \label{table:basiclayout}
  \centering
  \begin{tabularx}{0.9\textwidth}{@{}XXXXXX@{}}
    \toprule
    \begin{tabular}{@{}c@{}}Ground Truth \\ \includegraphics[width=0.13\textwidth]{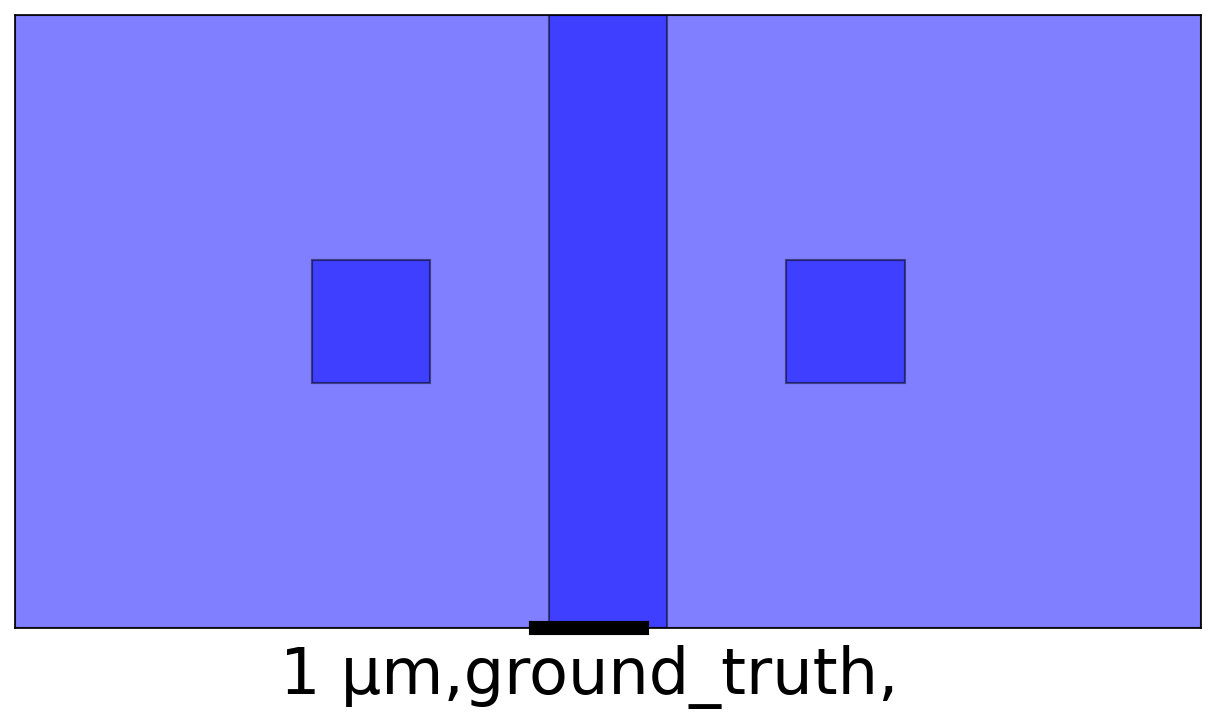}\end{tabular} & GPT-4o & Claude-3.5 & Llama-3-70B & Llama-3-405B & o1-preview \\
    \midrule
    SOLOMON & \includegraphics[width=0.13\textwidth]{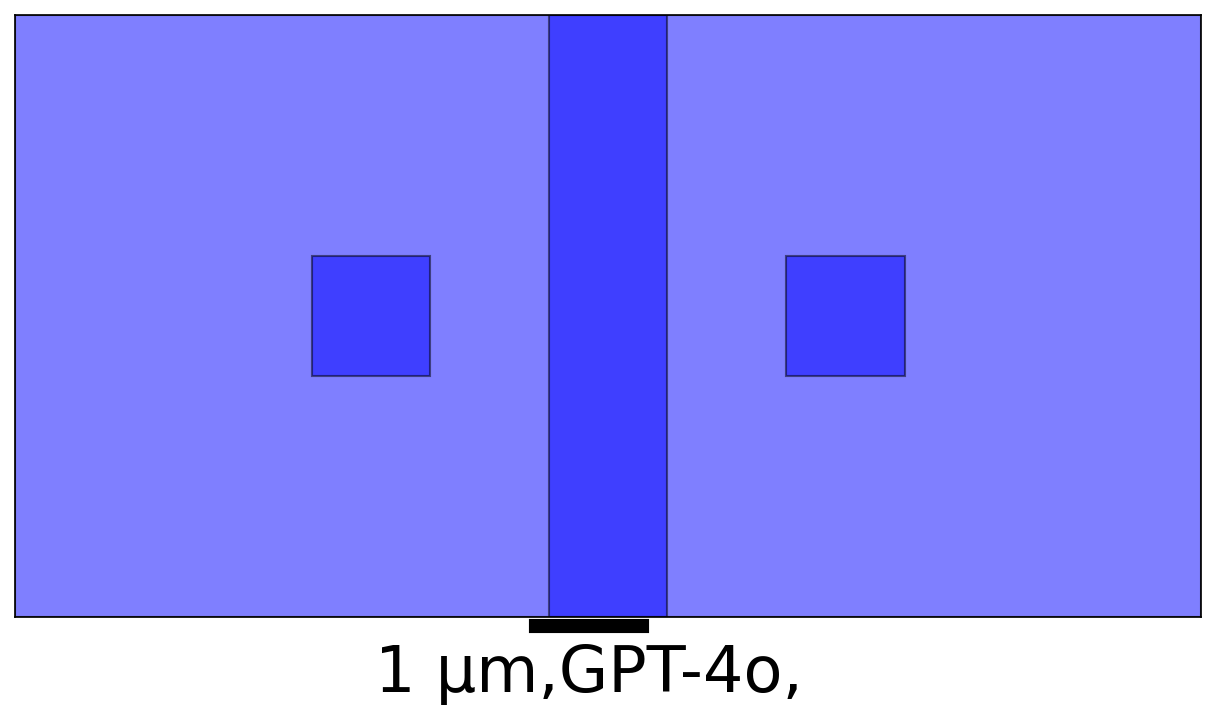} & \includegraphics[width=0.13\textwidth]{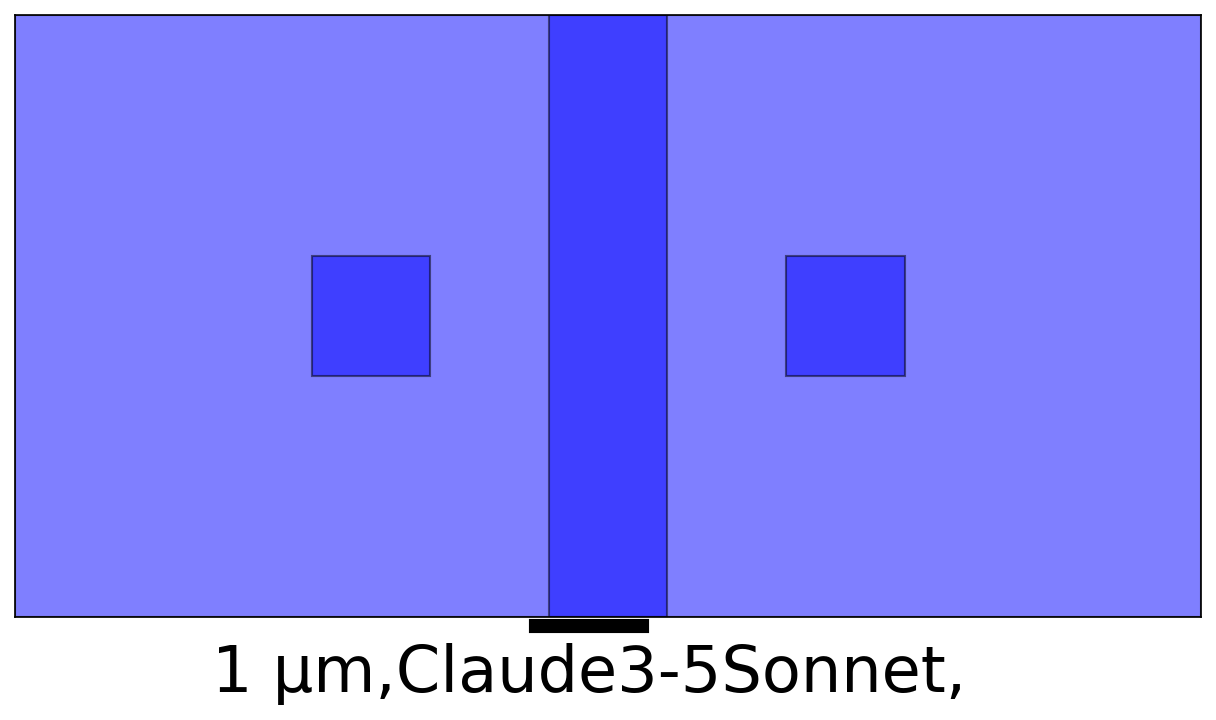} & \includegraphics[width=0.13\textwidth]{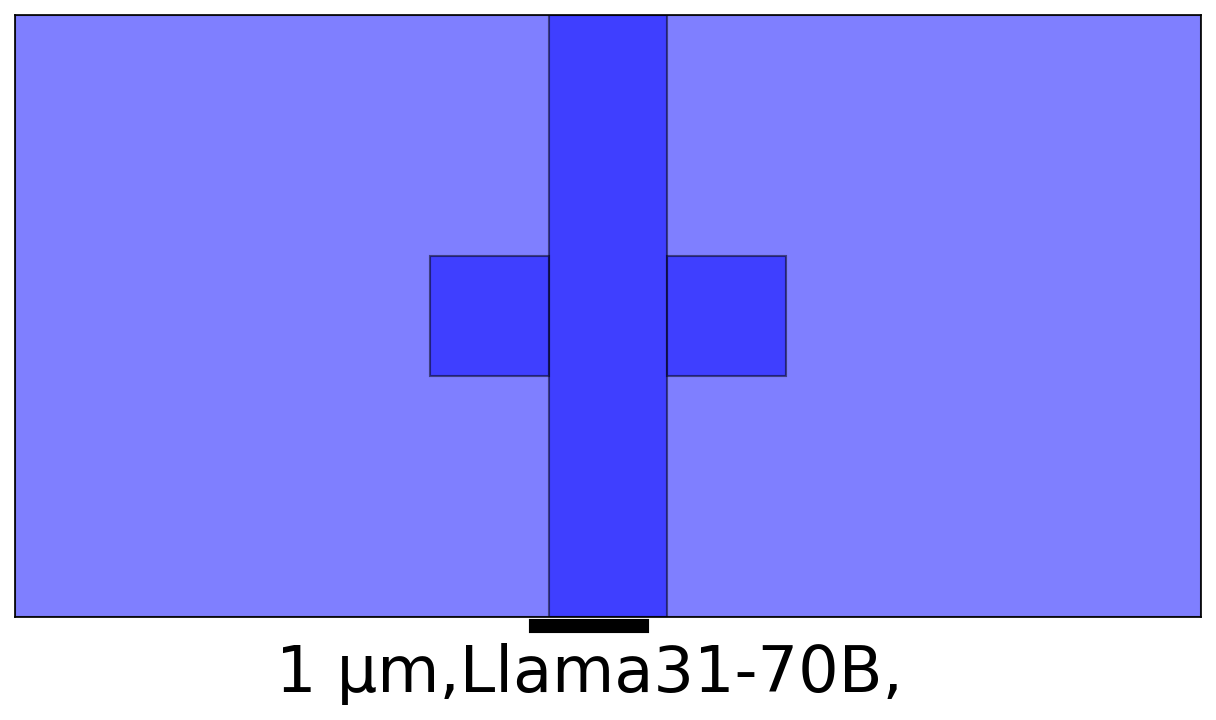} & \includegraphics[width=0.13\textwidth]{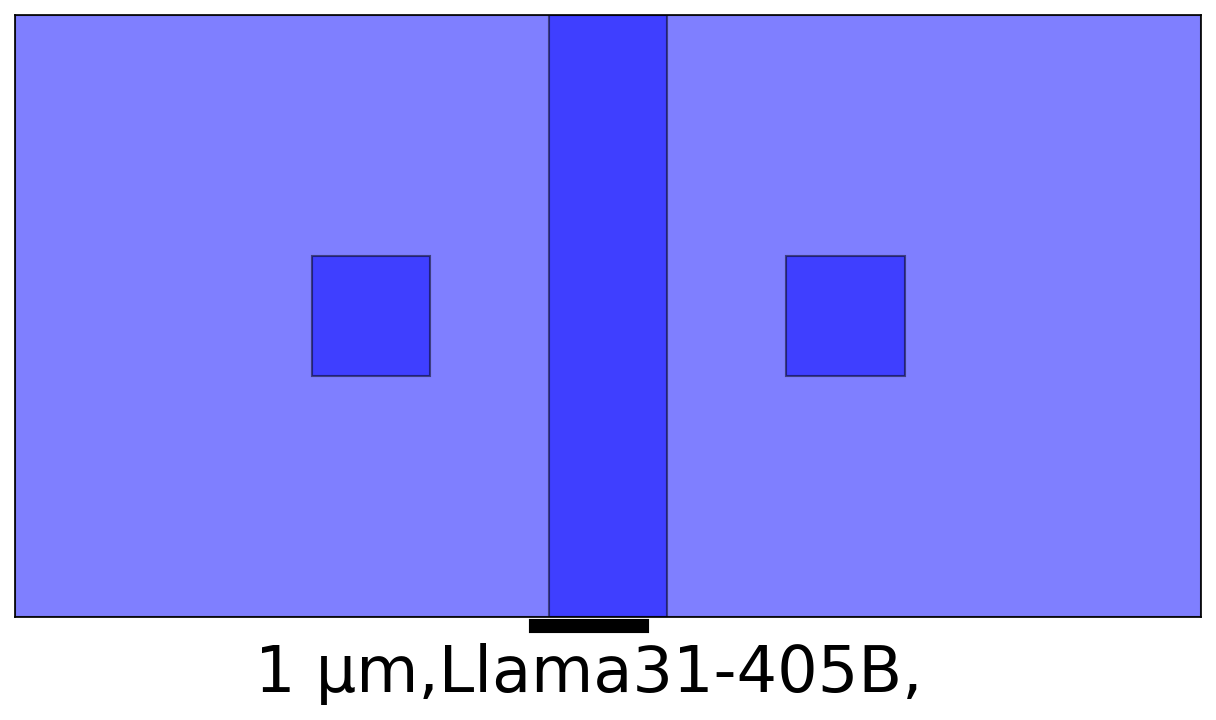} &  \\
    \begin{tabular}{@{}c@{}}Single LLM \\ Baseline \\ Run 1\end{tabular} & \includegraphics[width=0.13\textwidth]{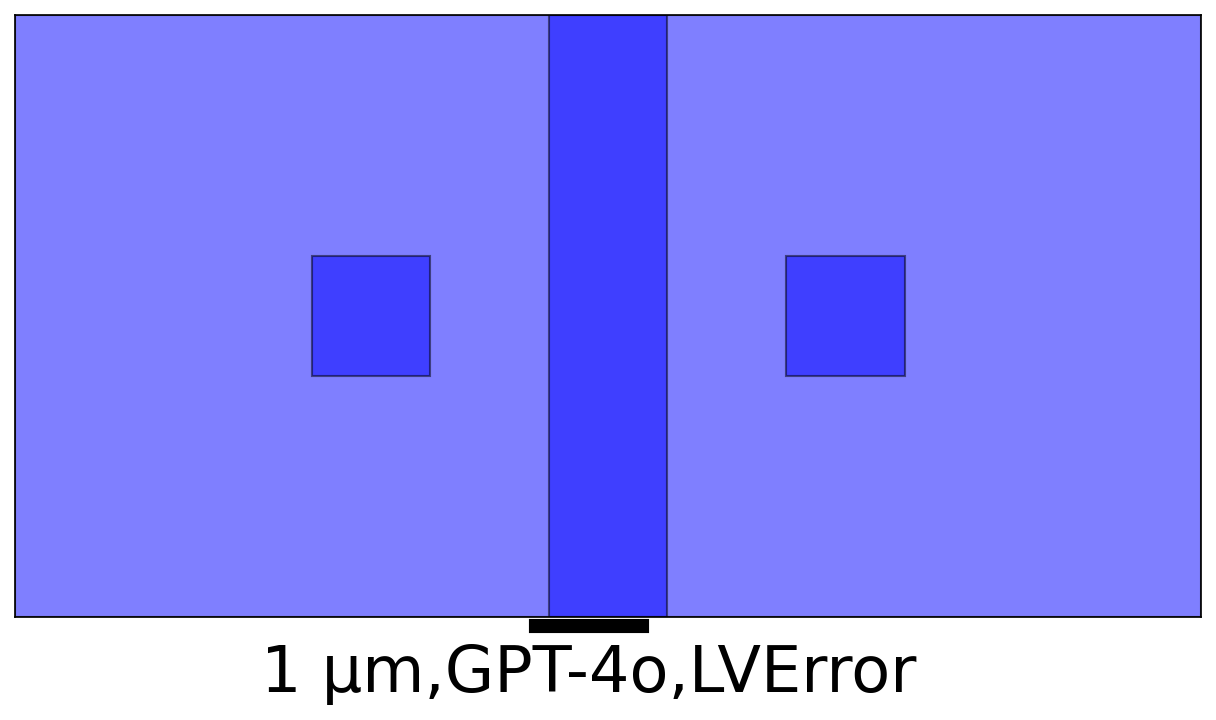} & \includegraphics[width=0.13\textwidth]{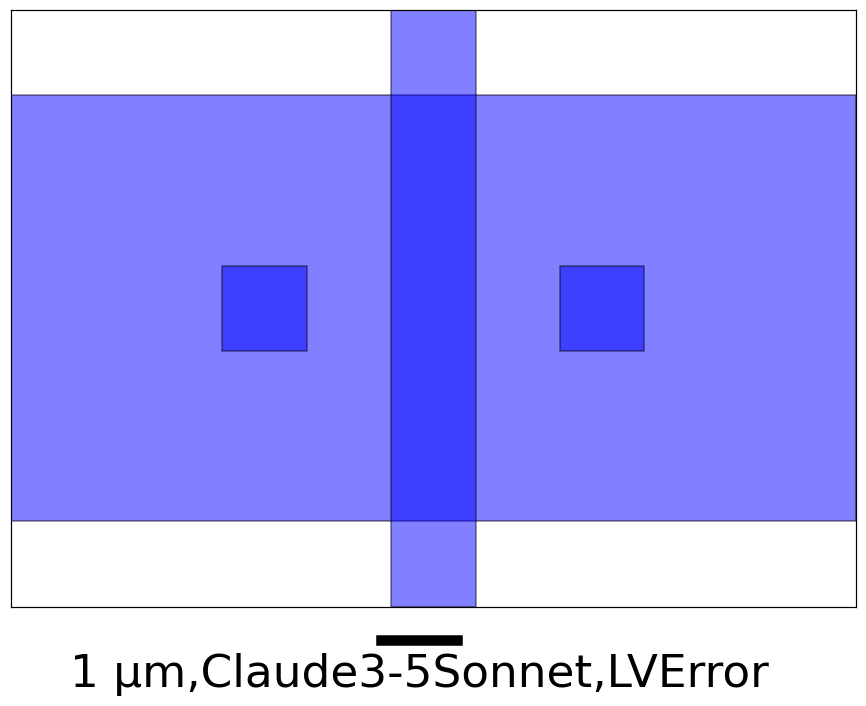} & \includegraphics[width=0.13\textwidth]{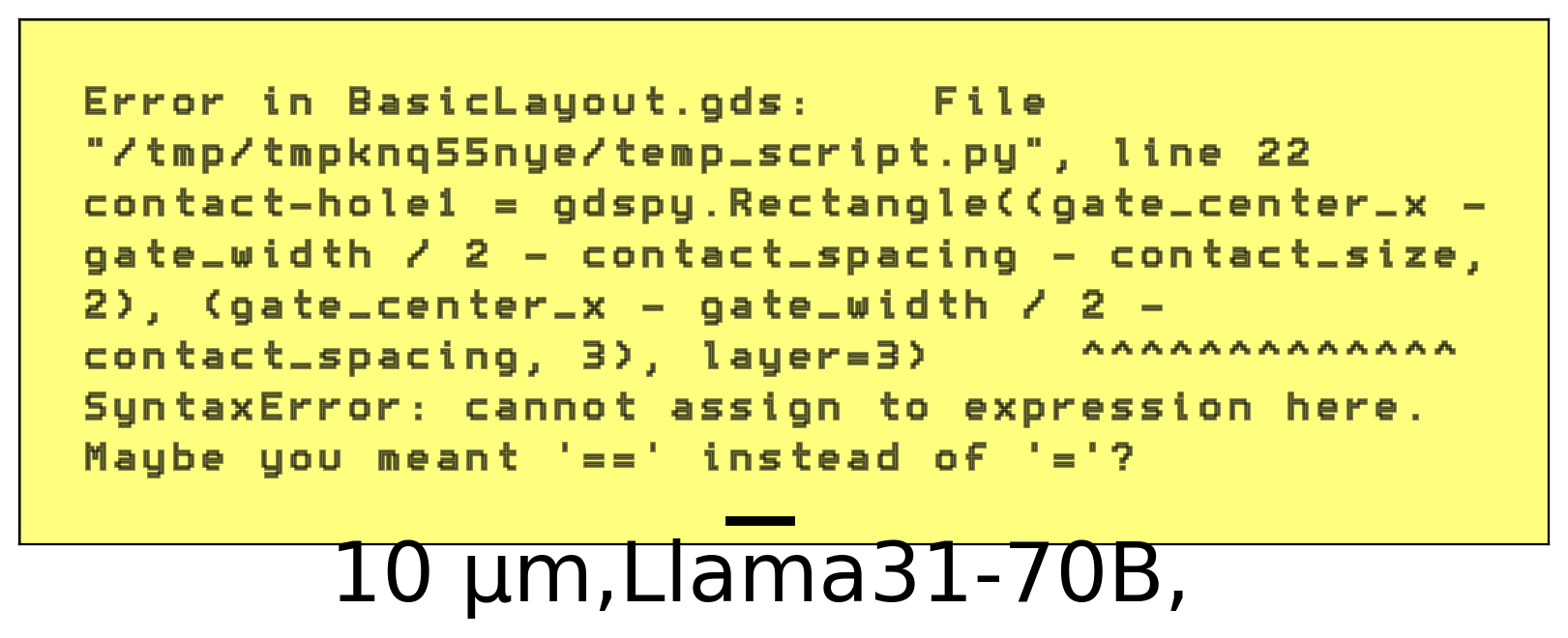} & \includegraphics[width=0.13\textwidth]{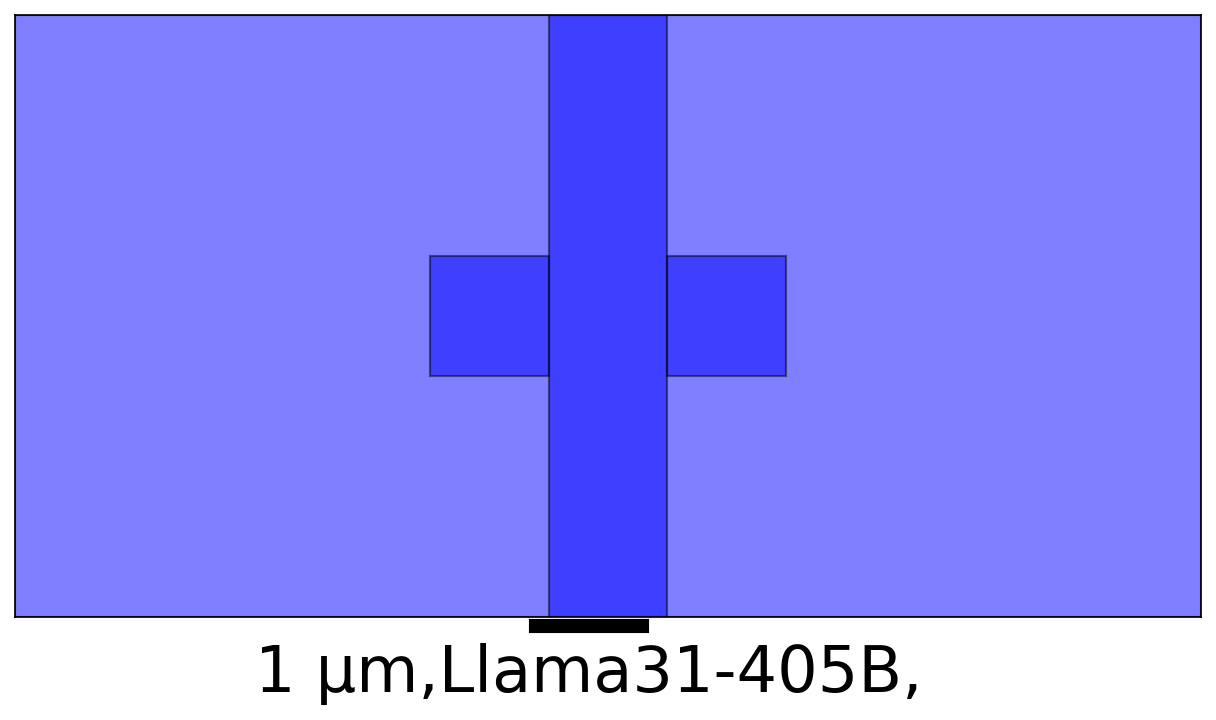} & \includegraphics[width=0.13\textwidth]{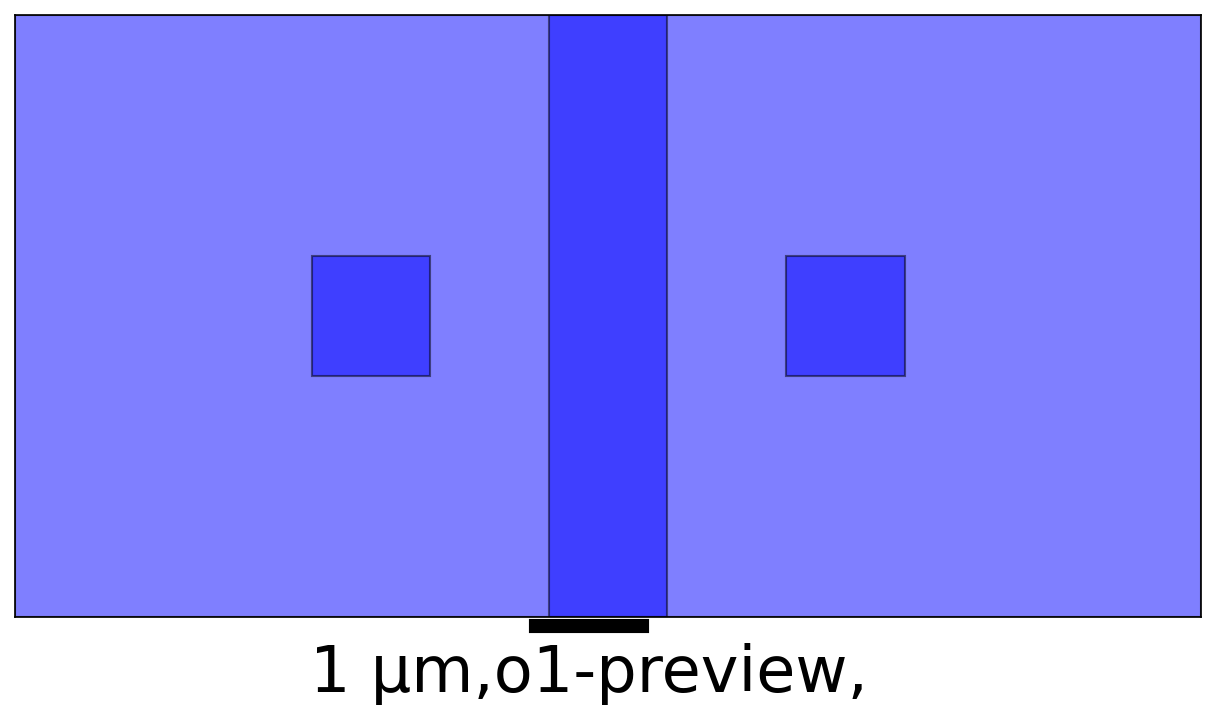} \\
    \begin{tabular}{@{}c@{}}Single LLM \\ Baseline \\ Run 2\end{tabular} & \includegraphics[width=0.13\textwidth]{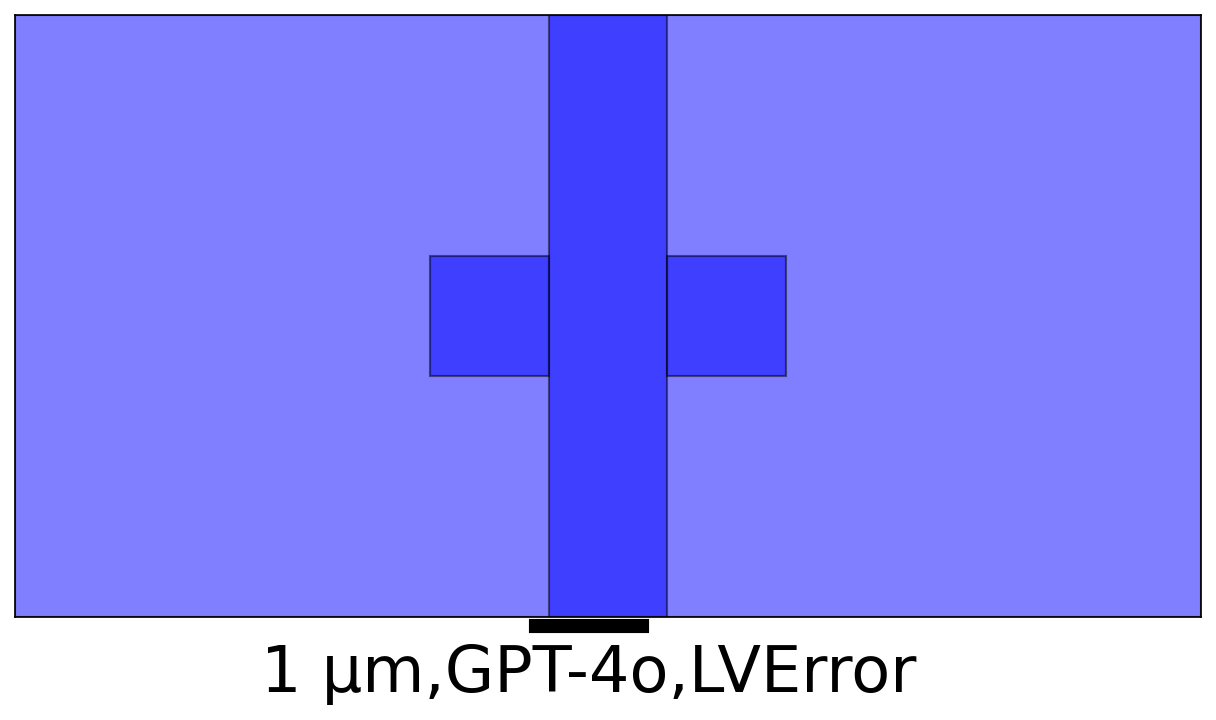} & \includegraphics[width=0.13\textwidth]{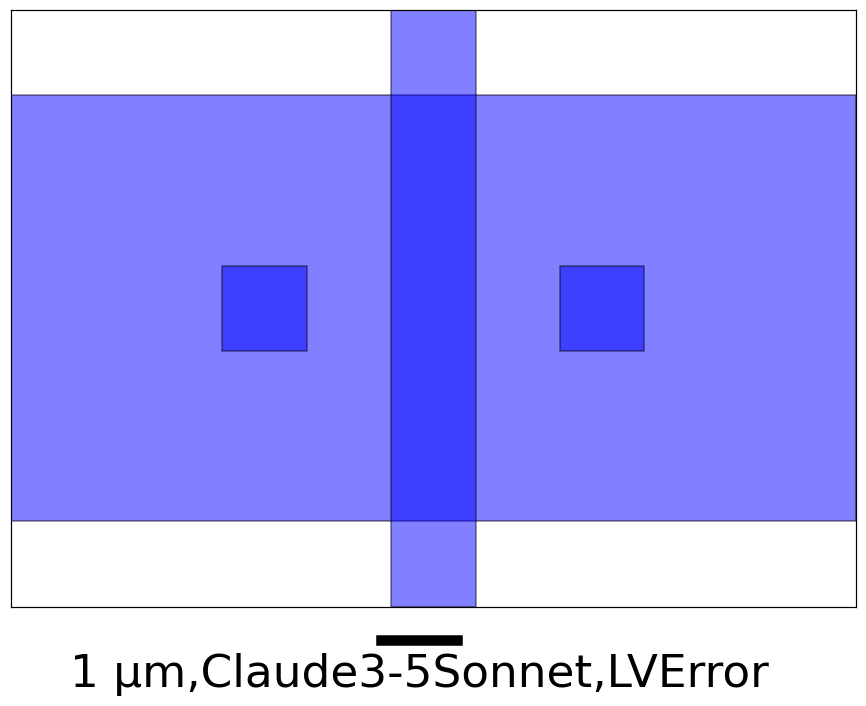} & \includegraphics[width=0.13\textwidth]{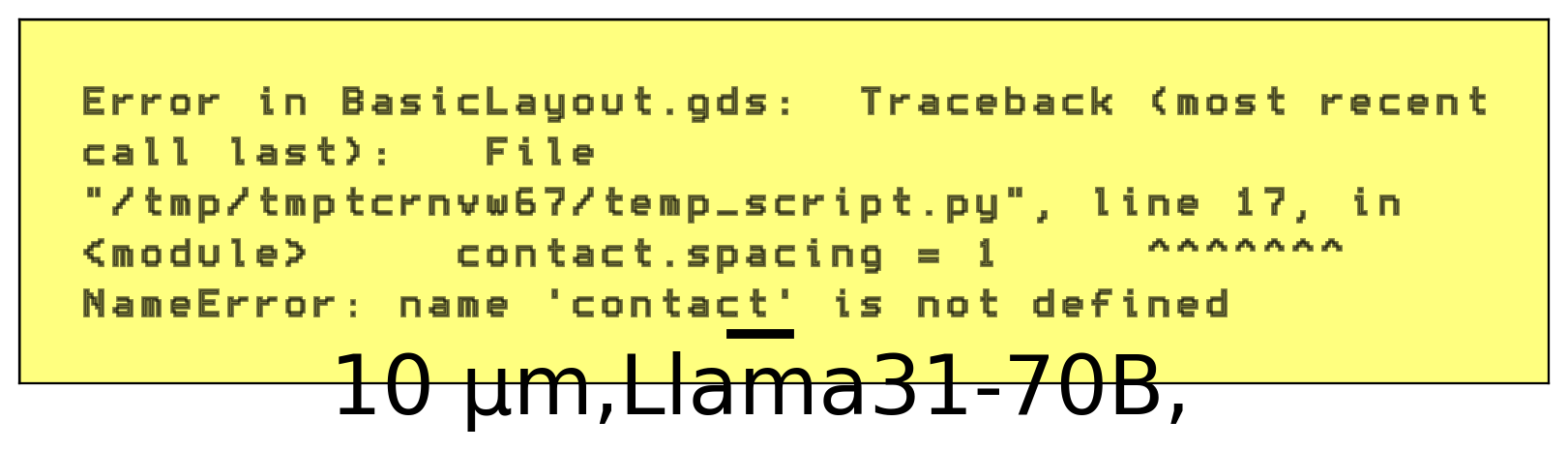} & \includegraphics[width=0.13\textwidth]{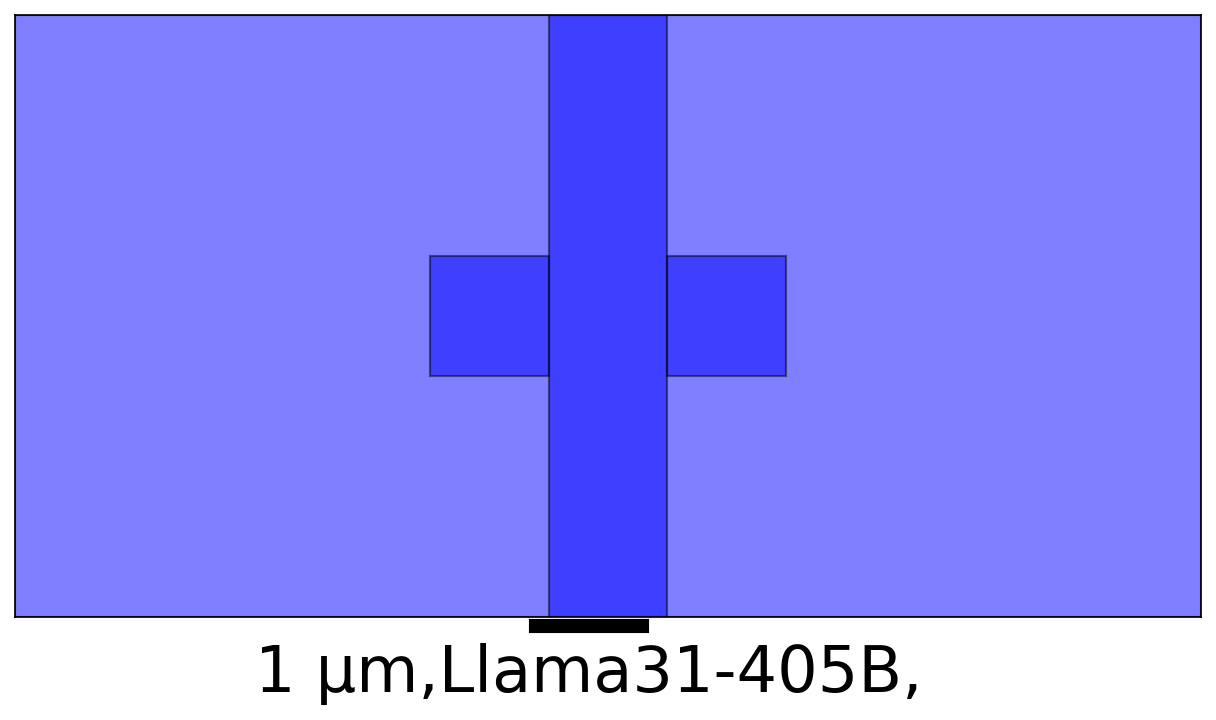} & \includegraphics[width=0.13\textwidth]{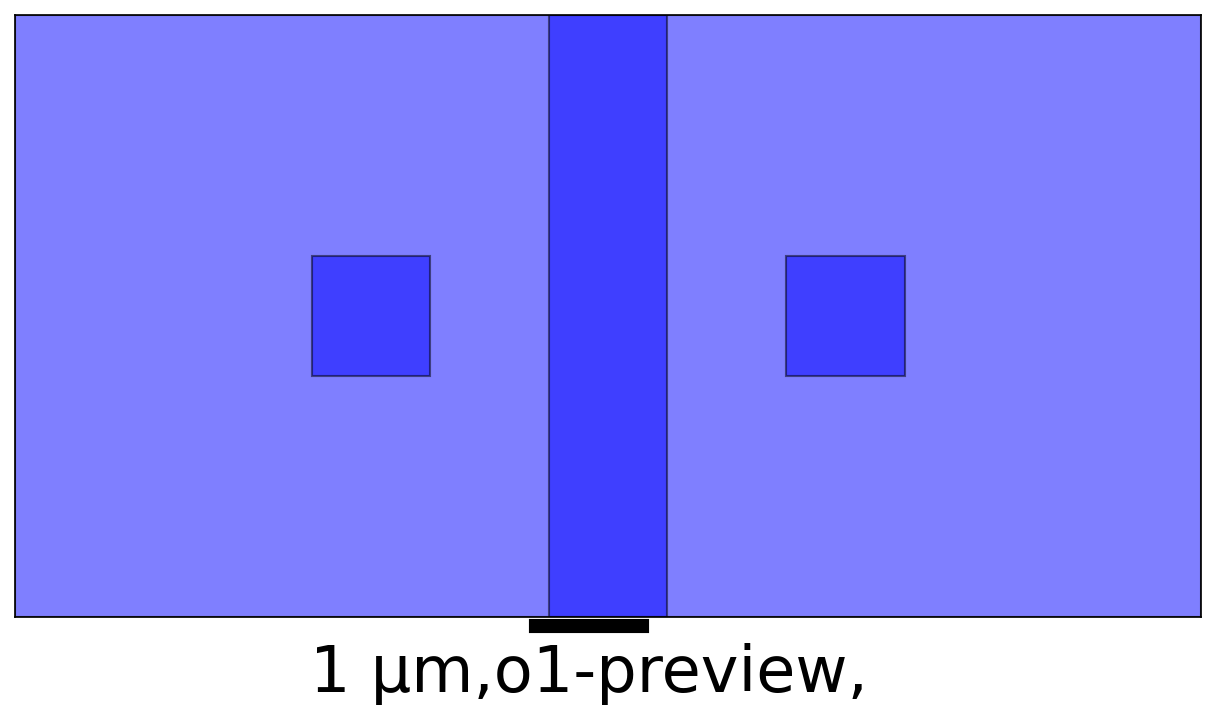} \\
    \begin{tabular}{@{}c@{}}Single LLM \\ Baseline \\ Run 3\end{tabular} & \includegraphics[width=0.13\textwidth]{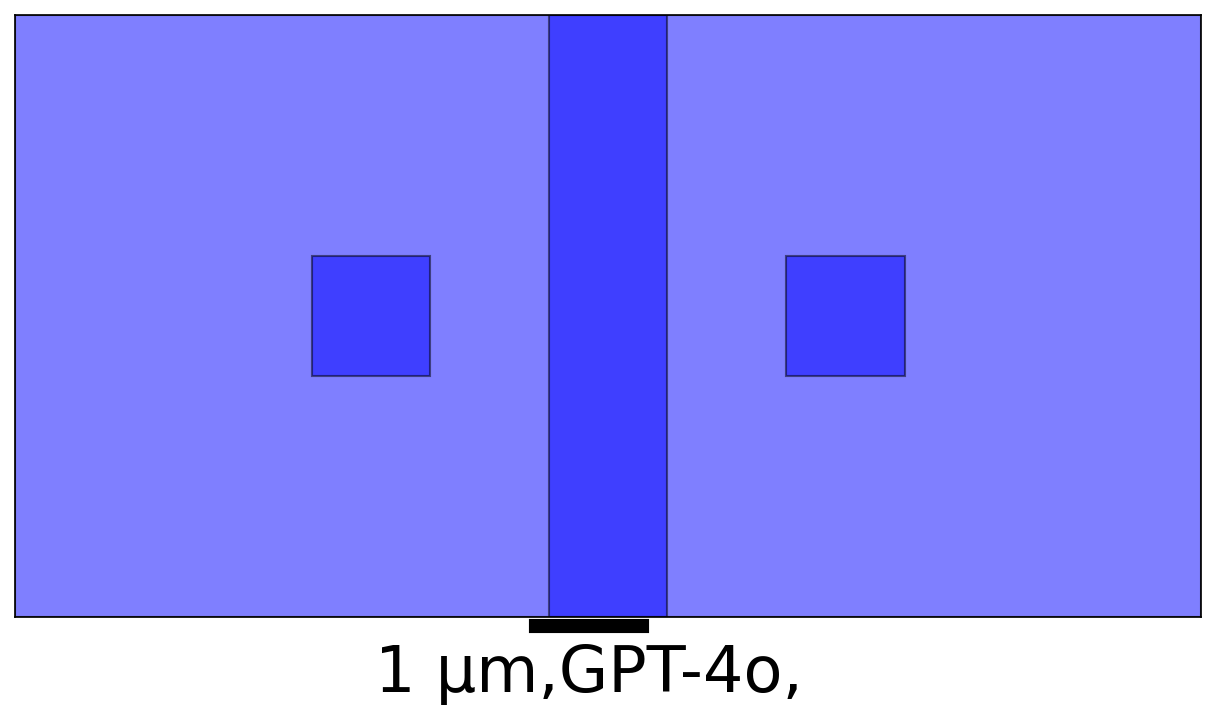} & \includegraphics[width=0.13\textwidth]{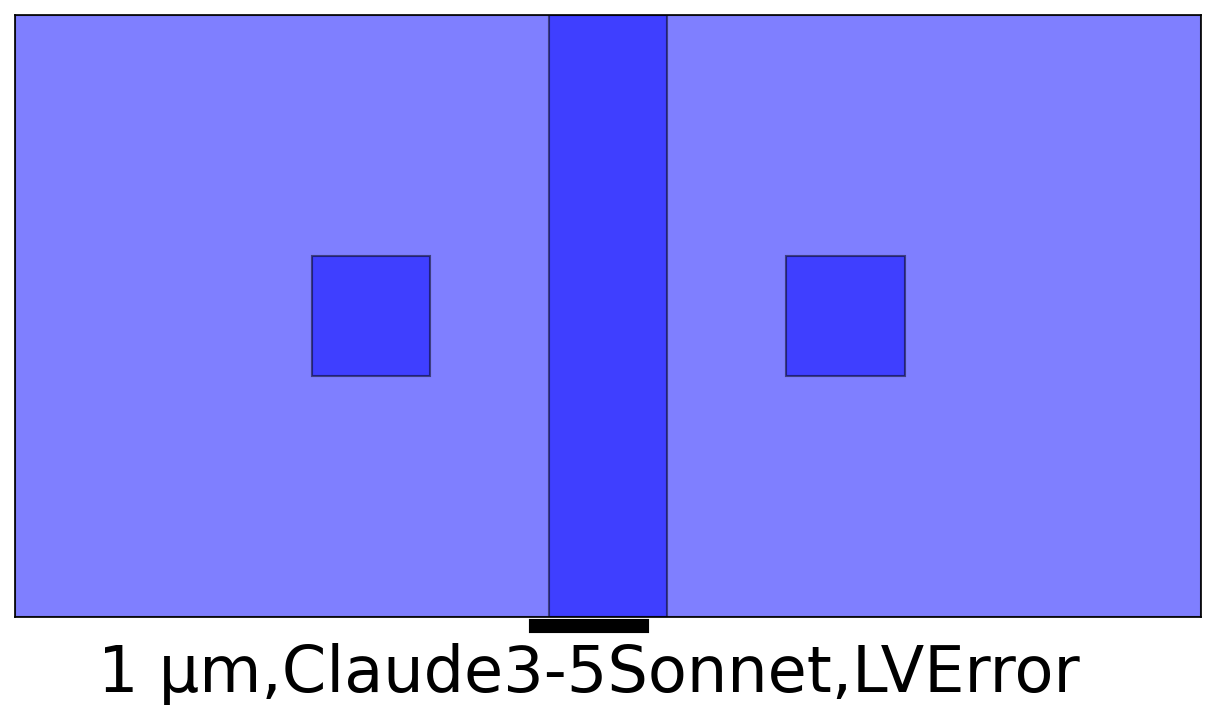} & \includegraphics[width=0.13\textwidth]{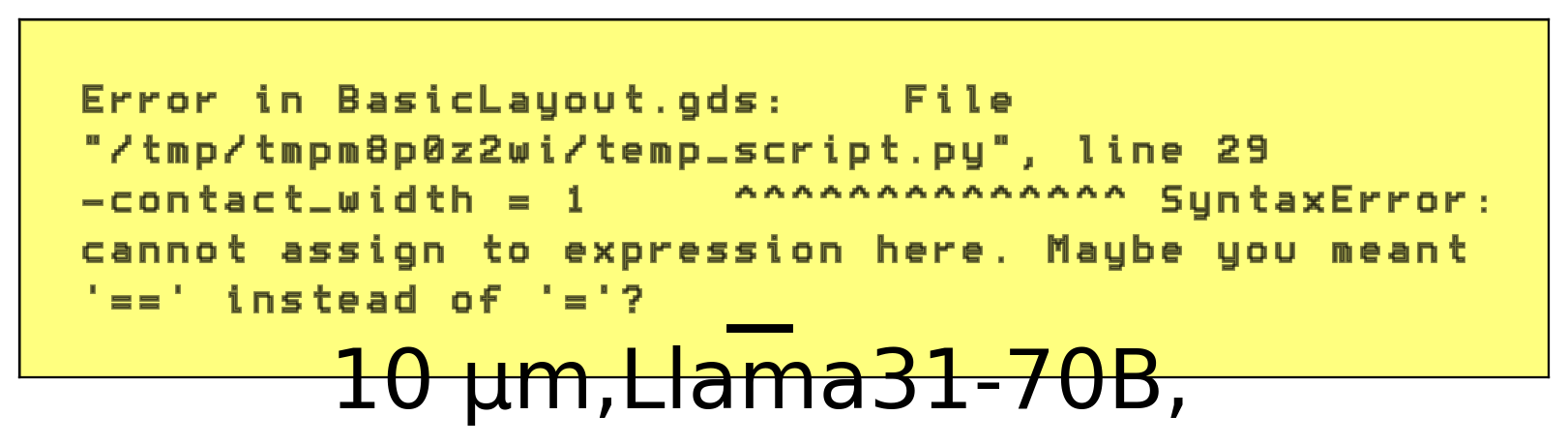} & \includegraphics[width=0.13\textwidth]{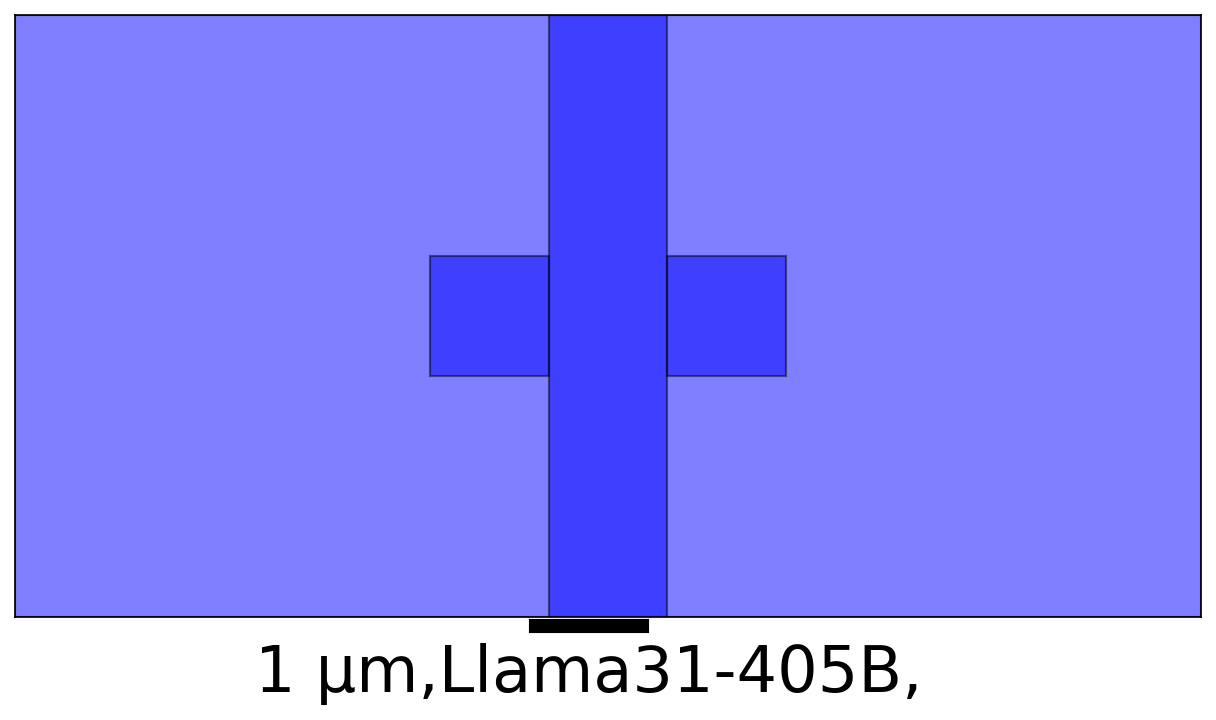} & \includegraphics[width=0.13\textwidth]{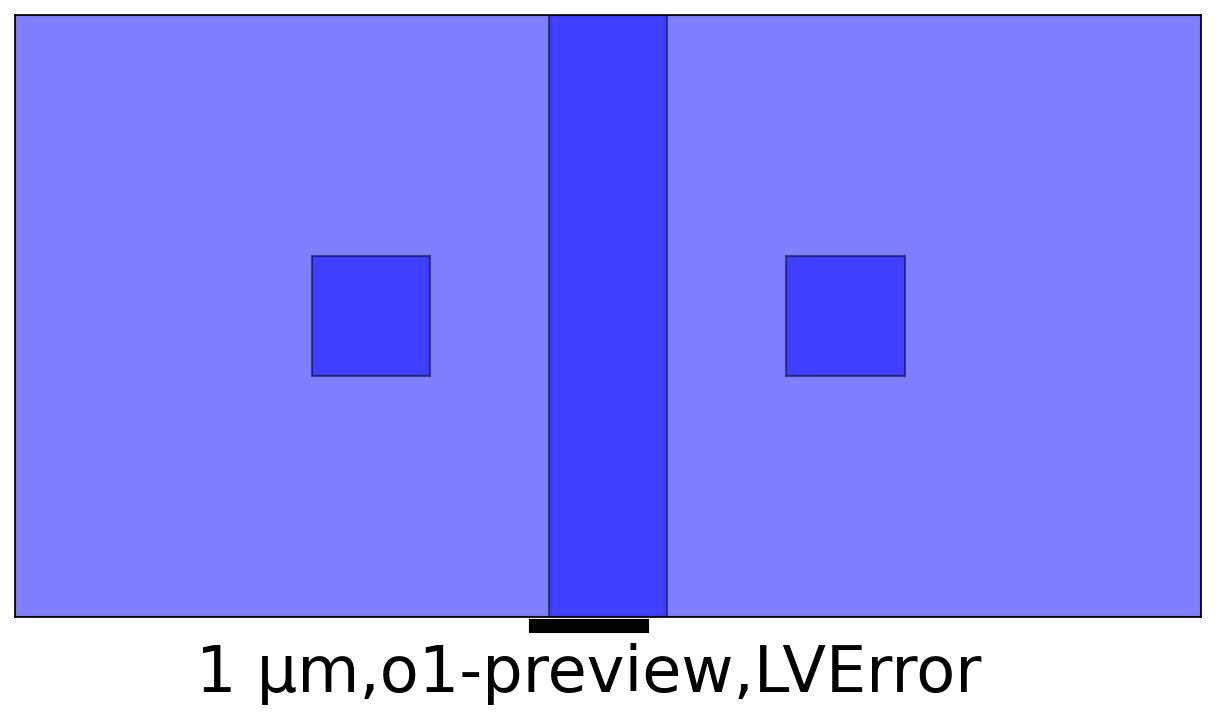} \\
    \begin{tabular}{@{}c@{}}Single LLM \\ Baseline \\ Run 4\end{tabular} & \includegraphics[width=0.13\textwidth]{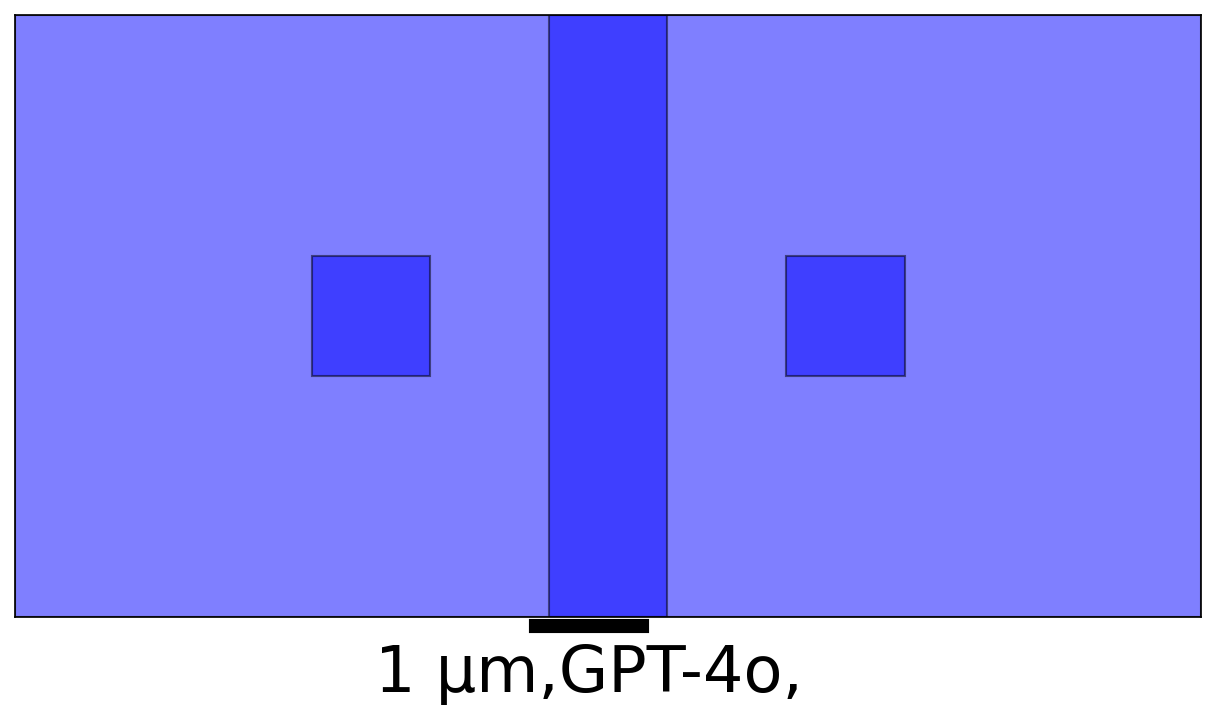} & \includegraphics[width=0.13\textwidth]{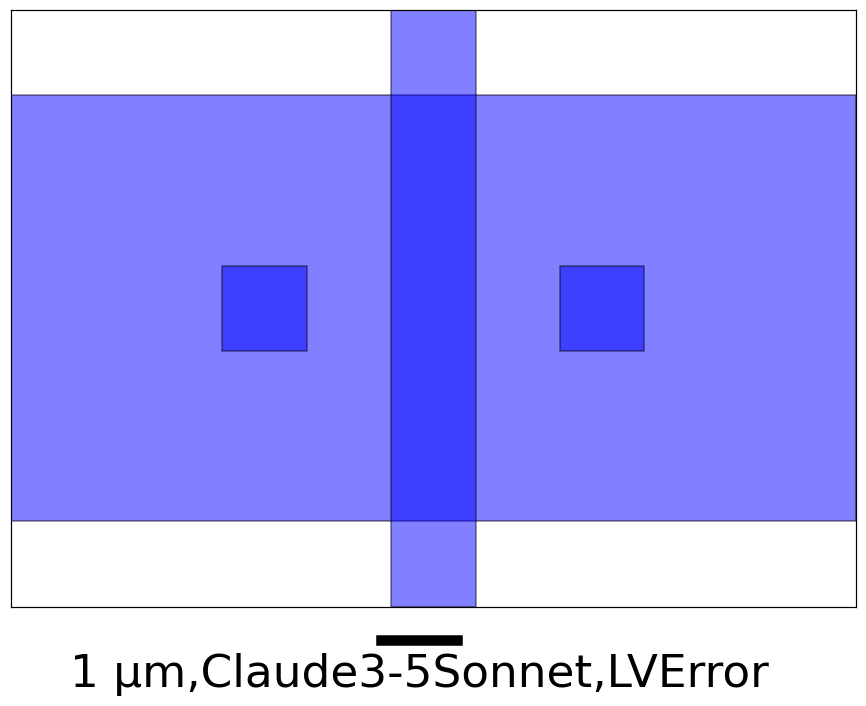} & \includegraphics[width=0.13\textwidth]{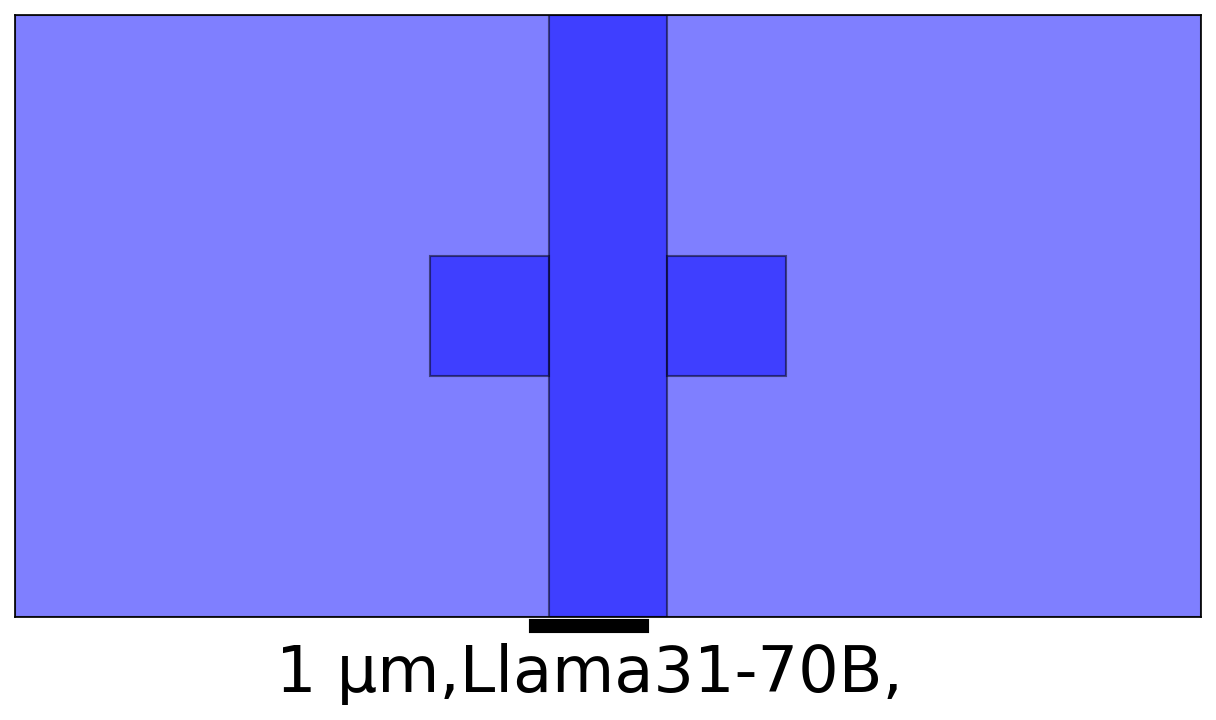} & \includegraphics[width=0.13\textwidth]{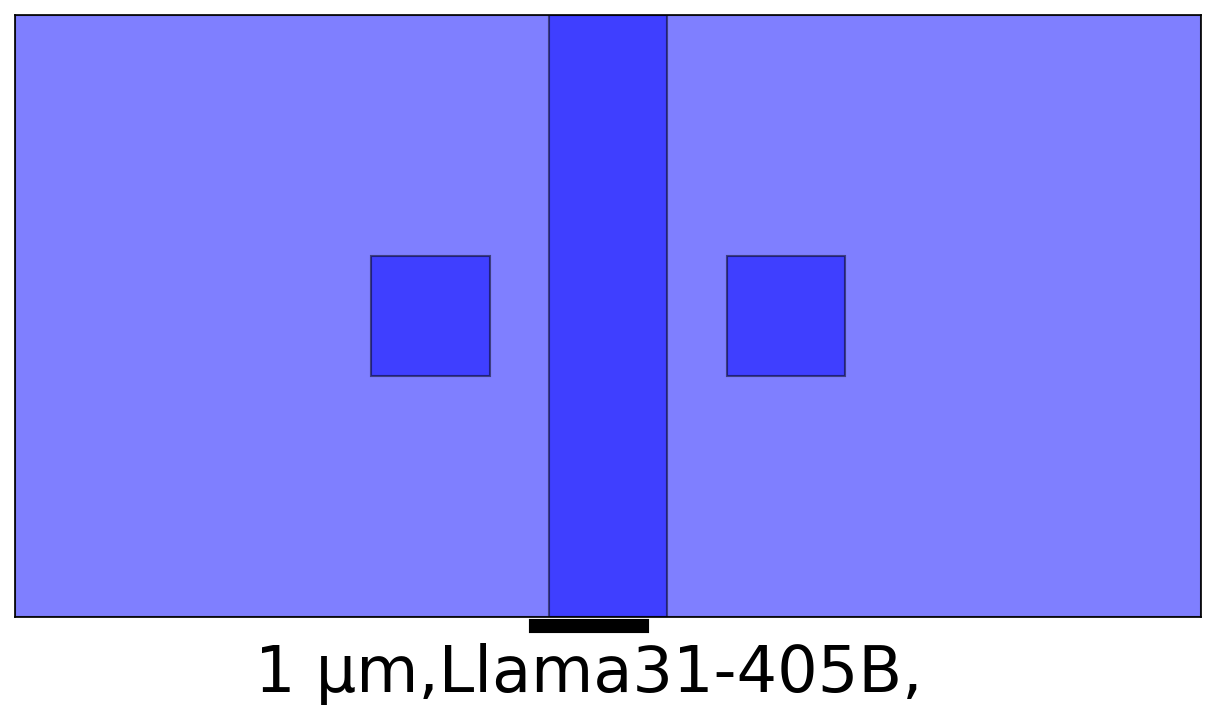} & \includegraphics[width=0.13\textwidth]{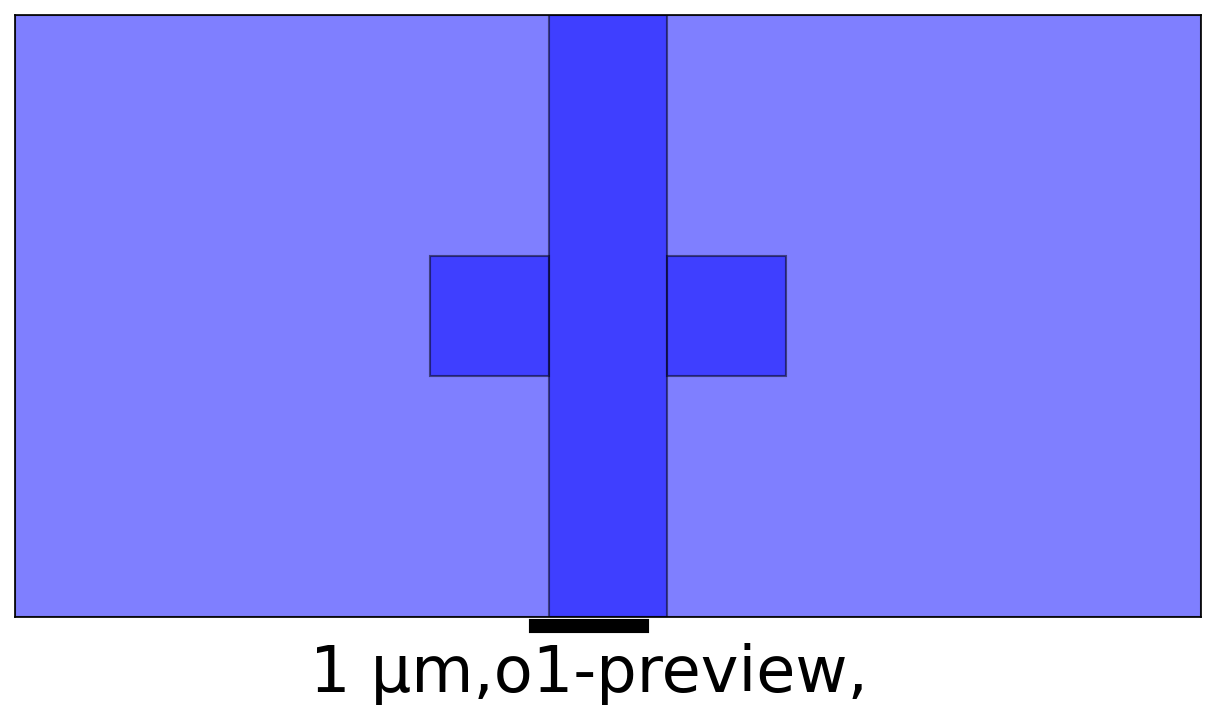} \\
    \begin{tabular}{@{}c@{}}Single LLM \\ Baseline \\ Run 5\end{tabular} & \includegraphics[width=0.13\textwidth]{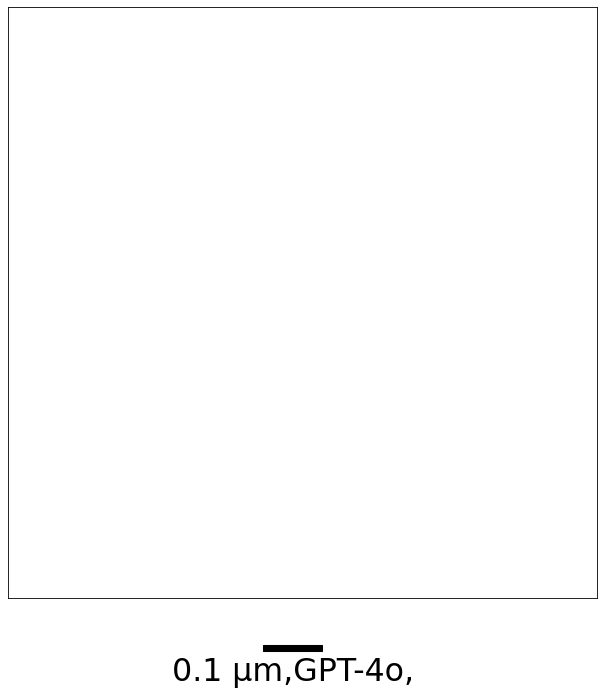} & \includegraphics[width=0.13\textwidth]{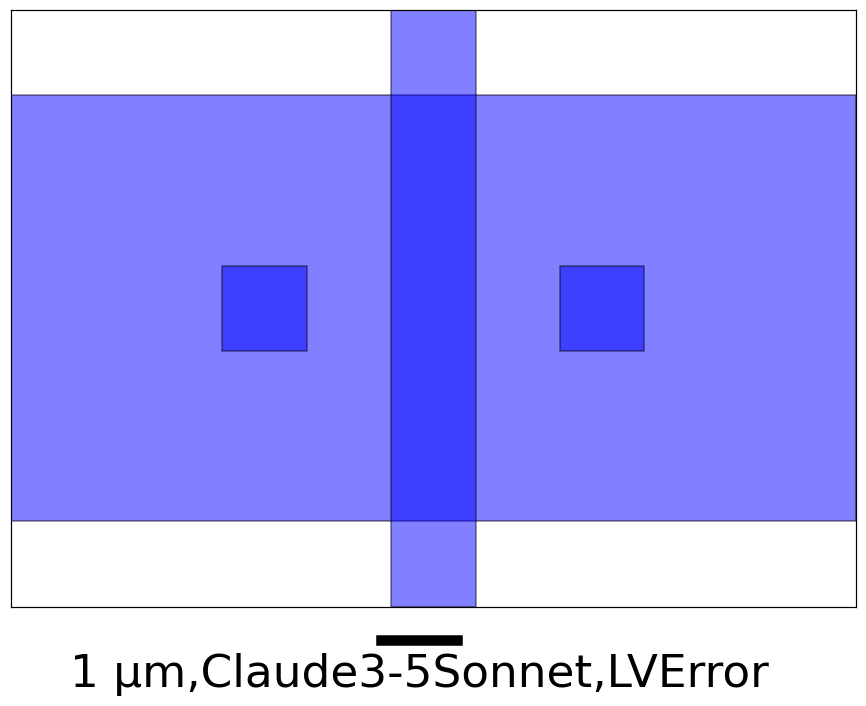} & \includegraphics[width=0.13\textwidth]{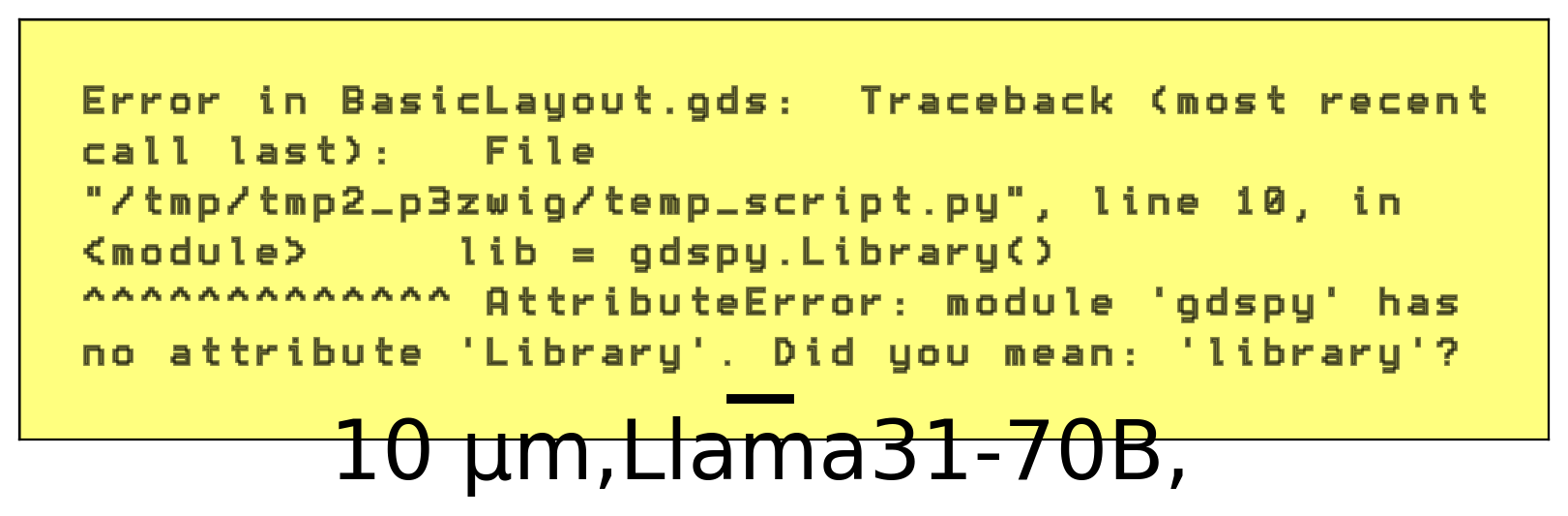} & \includegraphics[width=0.13\textwidth]{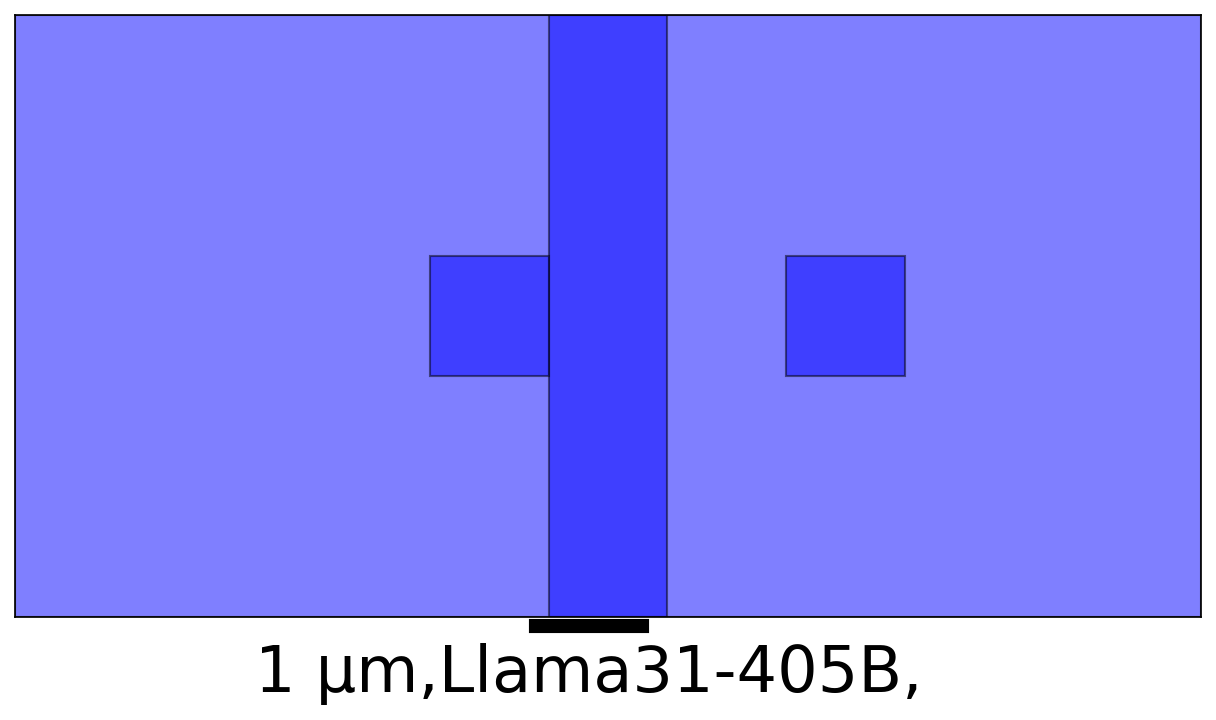} & \includegraphics[width=0.13\textwidth]{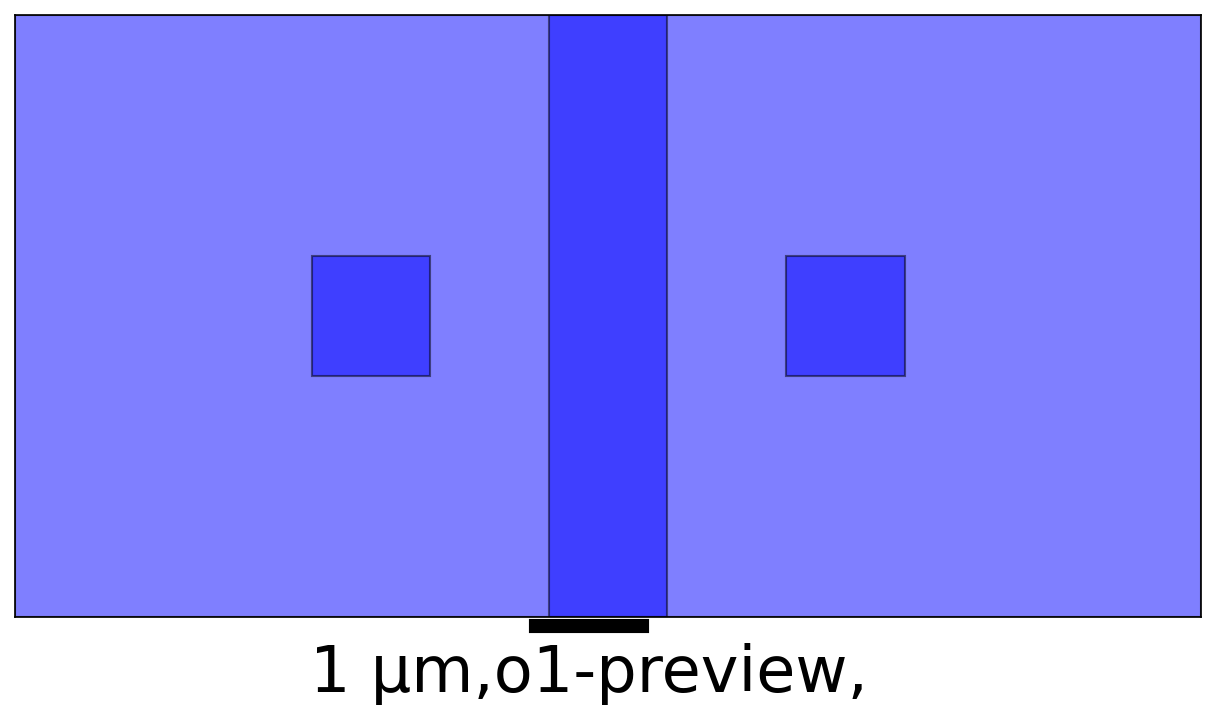} \\
    \bottomrule
  \end{tabularx}
\end{table}

\clearpage
\begin{table}[p]
  \caption{ViaConnection Task Question: Create a design with three layers: via layer (yellow), metal layer (blue), and pad layer (red). The via radius is 10 units, pad radius is 30 units, and metal connection width is 40 units with a total length of 600 units. Position the first via at (50, 150) and the second via at (550, 150). Ensure the metal connection fully covers the vias and leaves a margin of 10 units between the edge of the metal and the pads. Leave a space of 50 units between the vias and the edges of the metal connection.}
  \label{table:viaconnection}
  \centering
  \begin{tabularx}{0.9\textwidth}{@{}XXXXXX@{}}
    \toprule
    \begin{tabular}{@{}c@{}}Ground Truth \\ \includegraphics[width=0.13\textwidth]{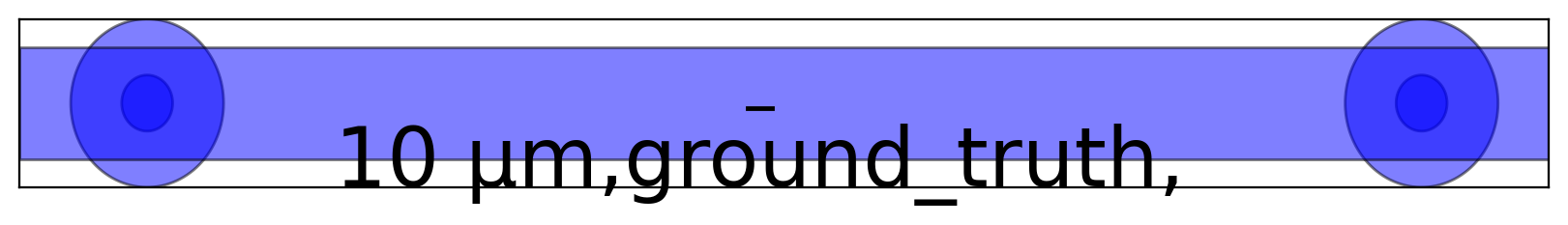}\end{tabular} & GPT-4o & Claude-3.5 & Llama-3-70B & Llama-3-405B & o1-preview \\
    \midrule
    SOLOMON & \includegraphics[width=0.13\textwidth]{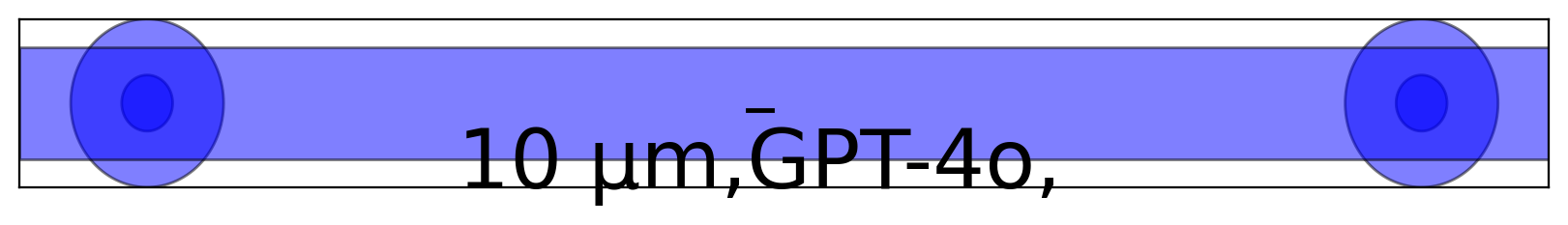} & \includegraphics[width=0.13\textwidth]{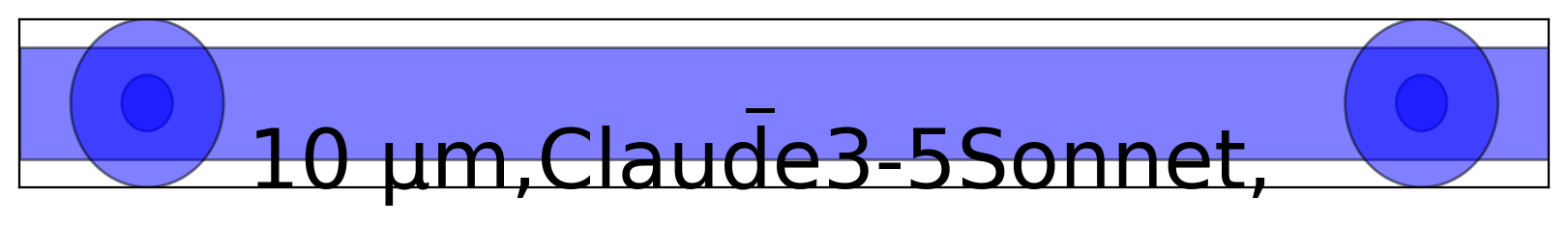} & \includegraphics[width=0.13\textwidth]{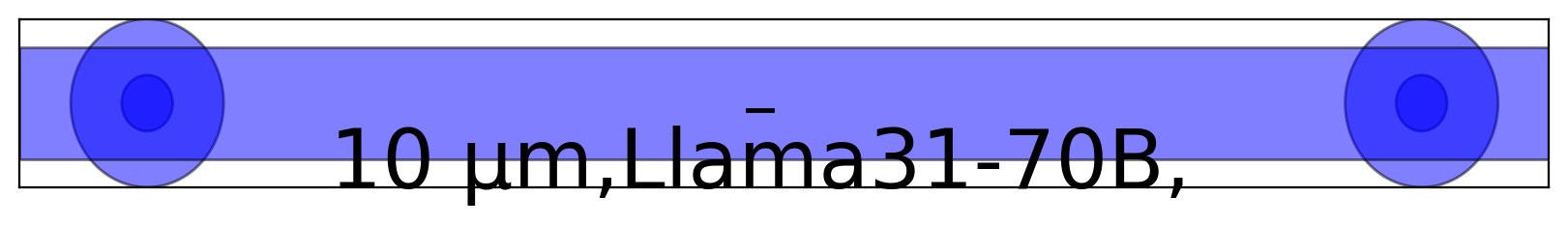} & \includegraphics[width=0.13\textwidth]{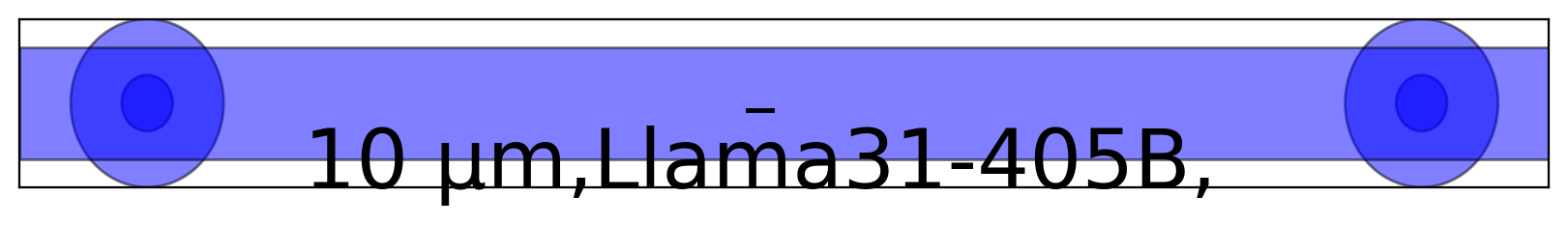} &  \\
    \begin{tabular}{@{}c@{}}Single LLM \\ Baseline \\ Run 1\end{tabular} & \includegraphics[width=0.13\textwidth]{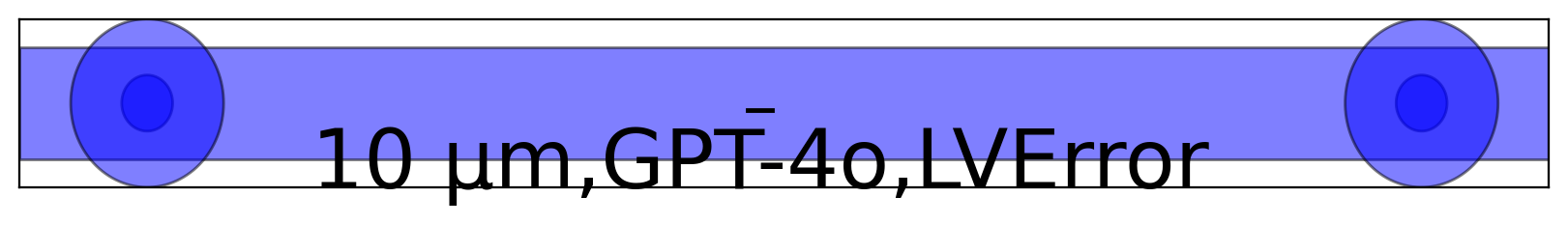} & \includegraphics[width=0.13\textwidth]{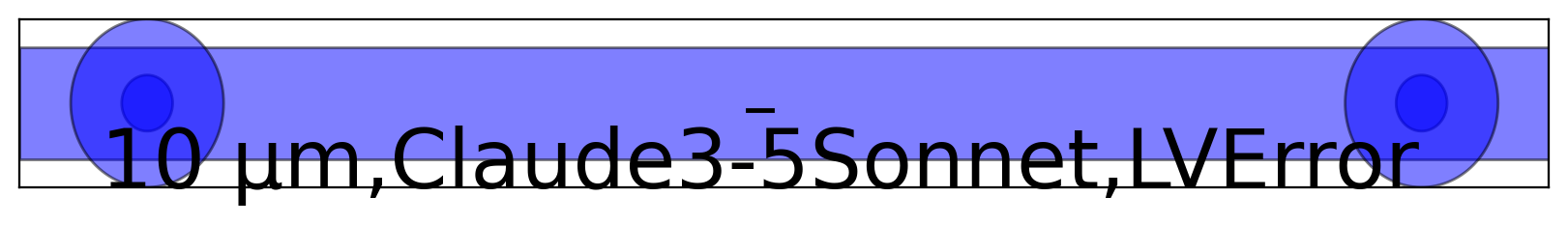} & \includegraphics[width=0.13\textwidth]{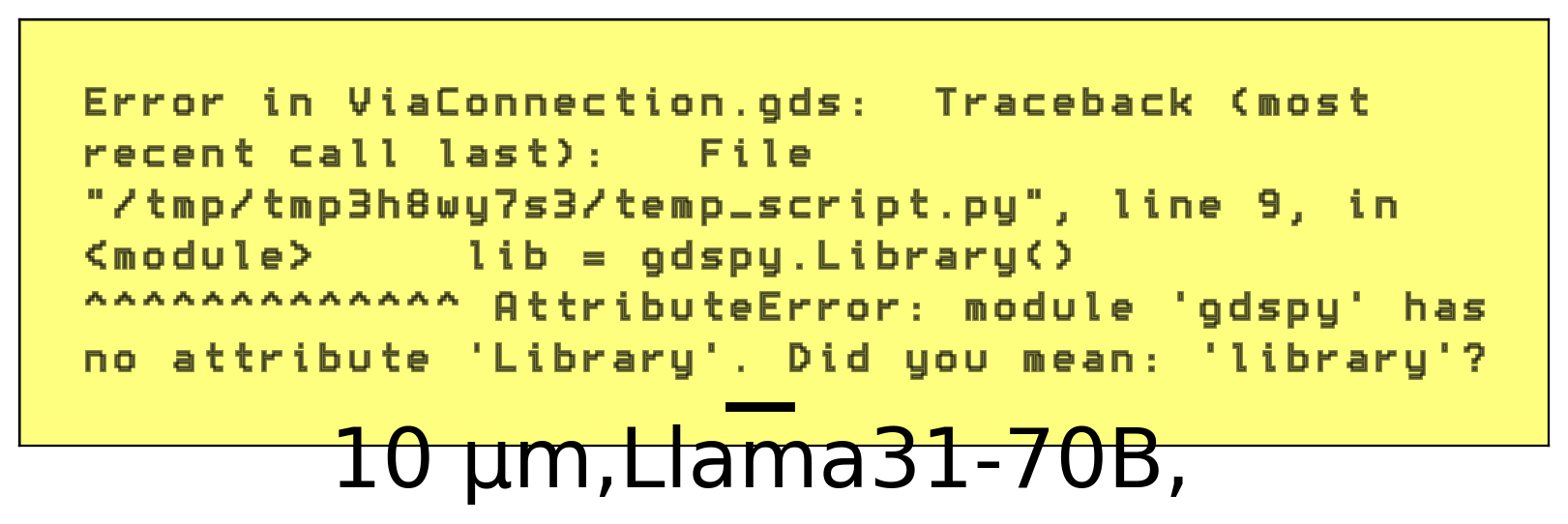} & \includegraphics[width=0.13\textwidth]{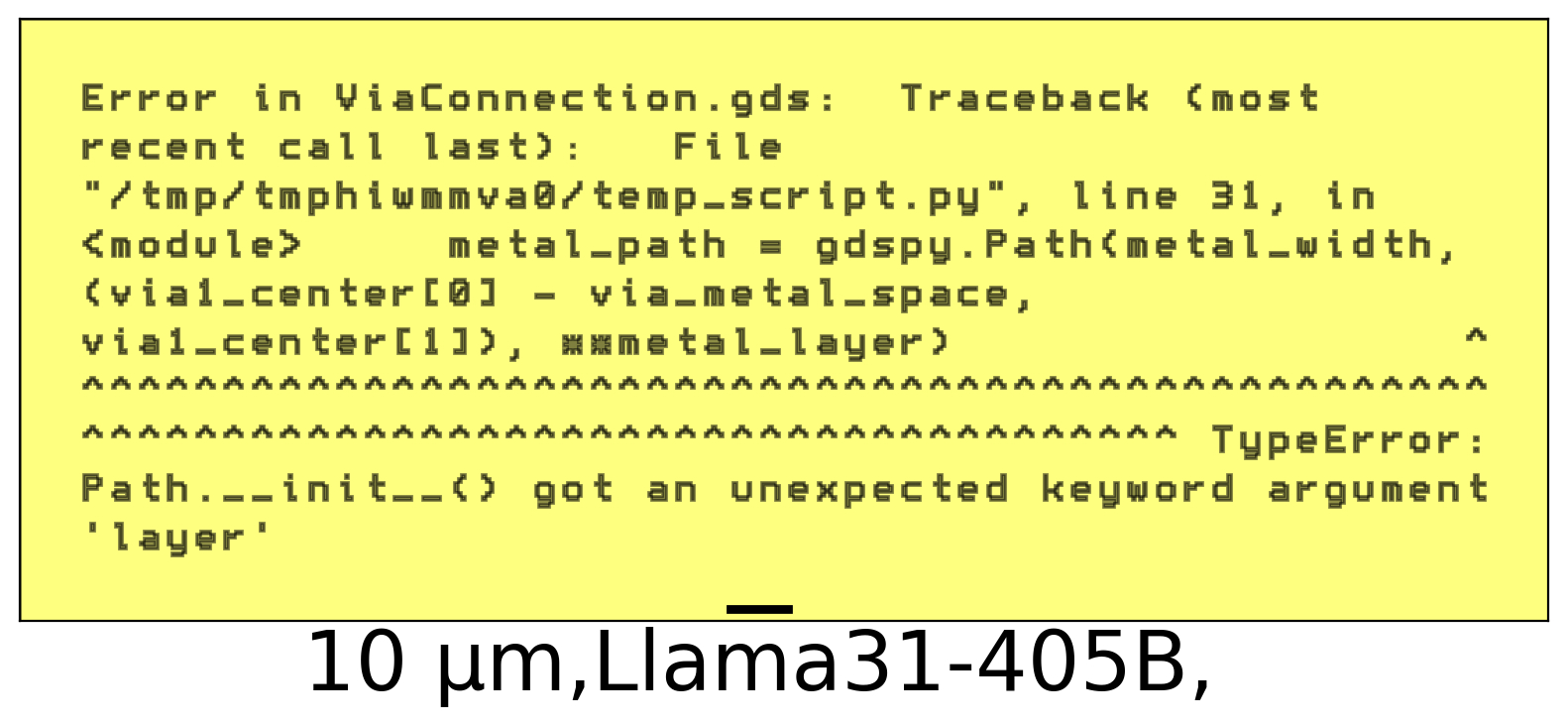} & \includegraphics[width=0.13\textwidth]{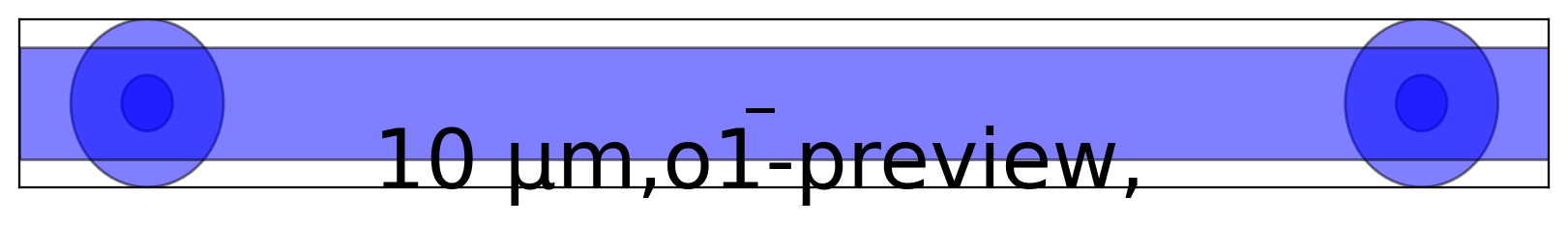} \\
    \begin{tabular}{@{}c@{}}Single LLM \\ Baseline \\ Run 2\end{tabular} & \includegraphics[width=0.13\textwidth]{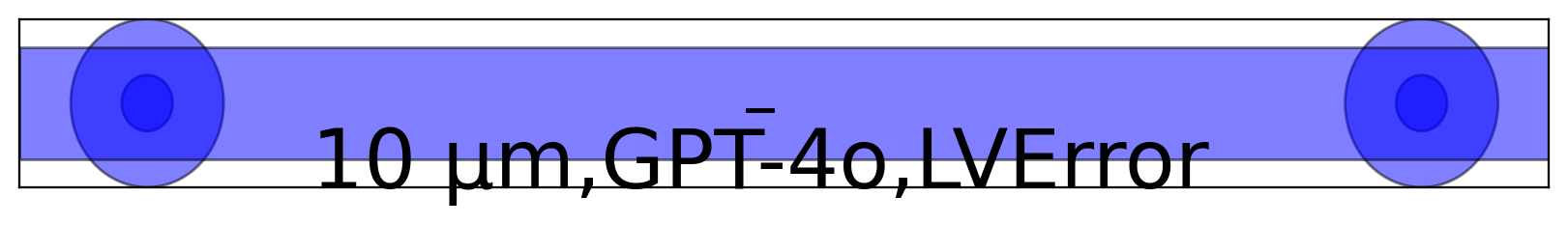} & \includegraphics[width=0.13\textwidth]{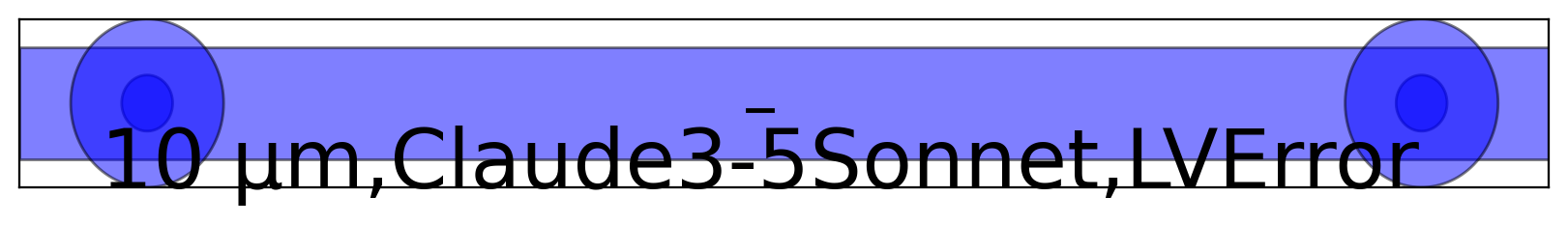} & \includegraphics[width=0.13\textwidth]{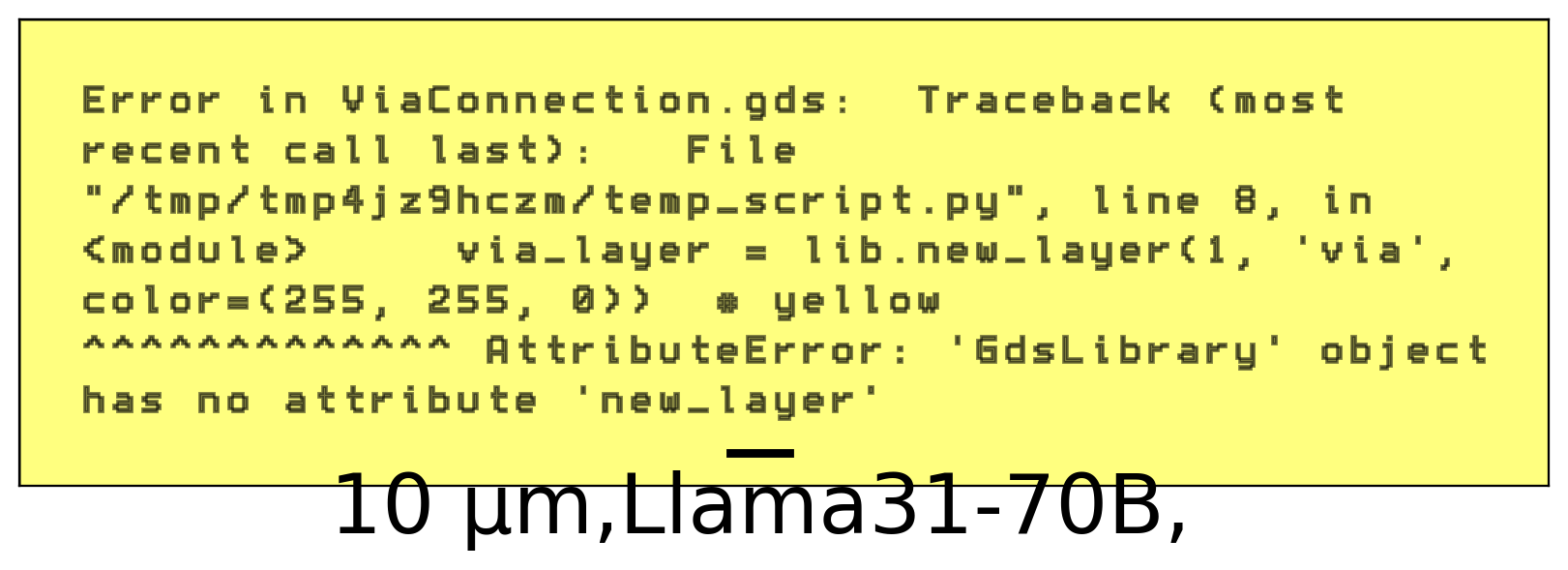} & \includegraphics[width=0.13\textwidth]{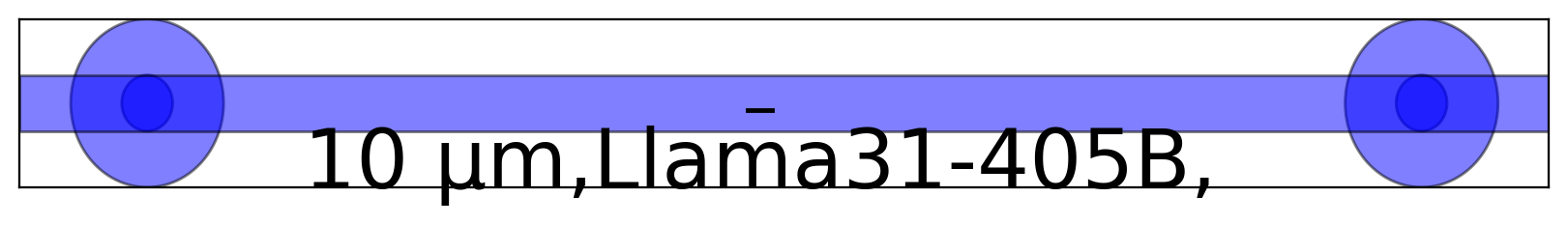} & \includegraphics[width=0.13\textwidth]{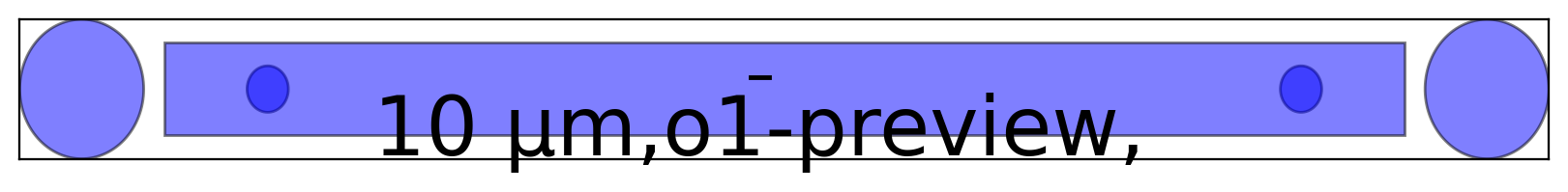} \\
    \begin{tabular}{@{}c@{}}Single LLM \\ Baseline \\ Run 3\end{tabular} & \includegraphics[width=0.13\textwidth]{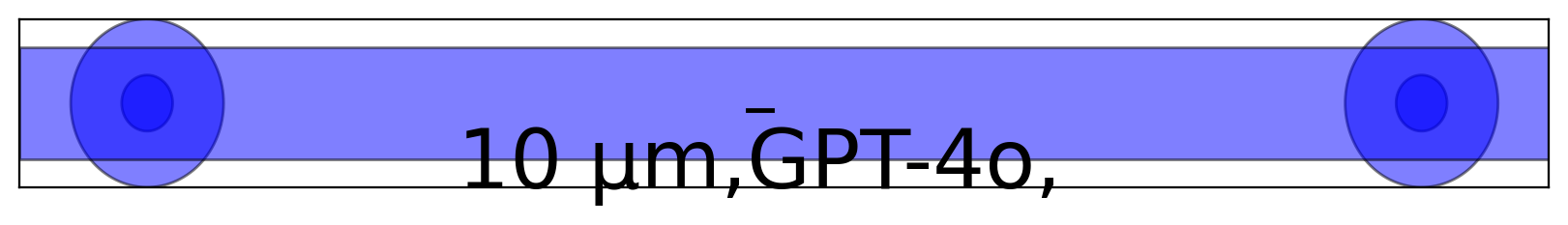} & \includegraphics[width=0.13\textwidth]{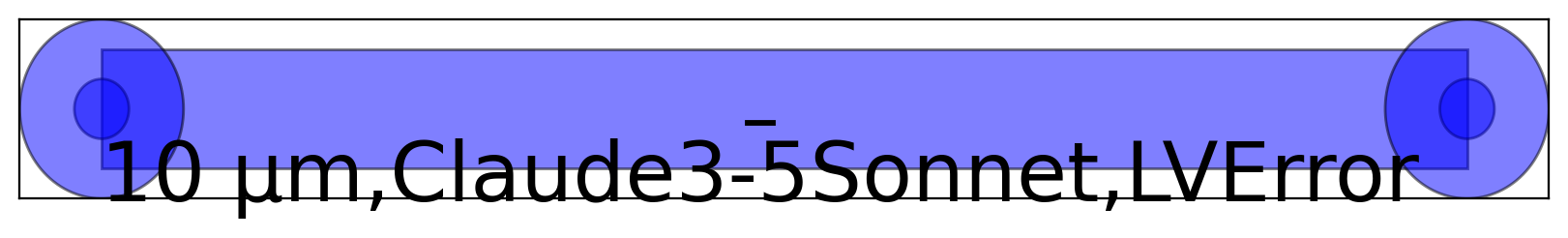} & \includegraphics[width=0.13\textwidth]{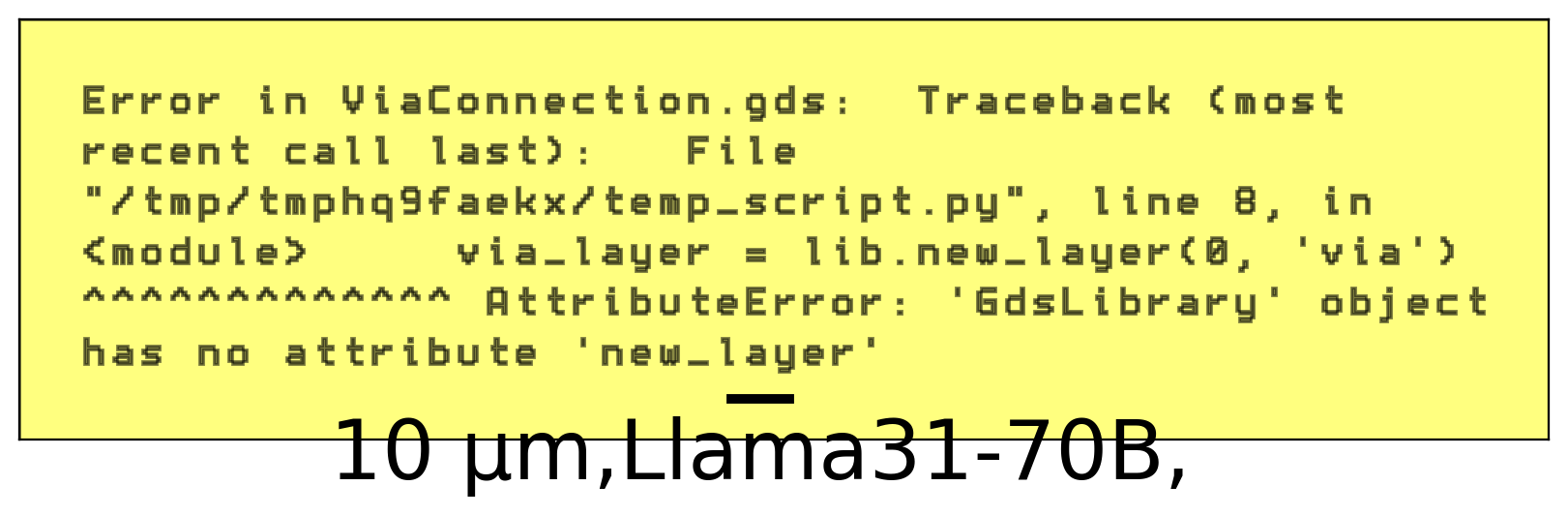} & \includegraphics[width=0.13\textwidth]{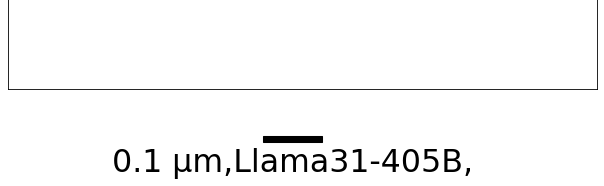} & \includegraphics[width=0.13\textwidth]{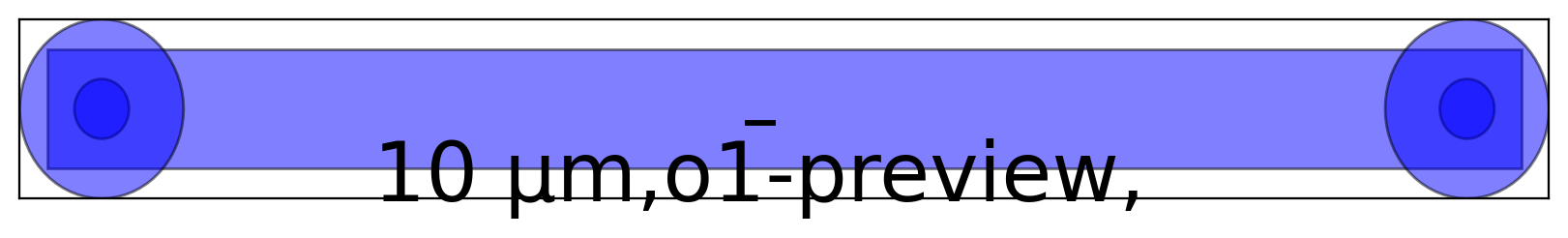} \\
    \begin{tabular}{@{}c@{}}Single LLM \\ Baseline \\ Run 4\end{tabular} & \includegraphics[width=0.13\textwidth]{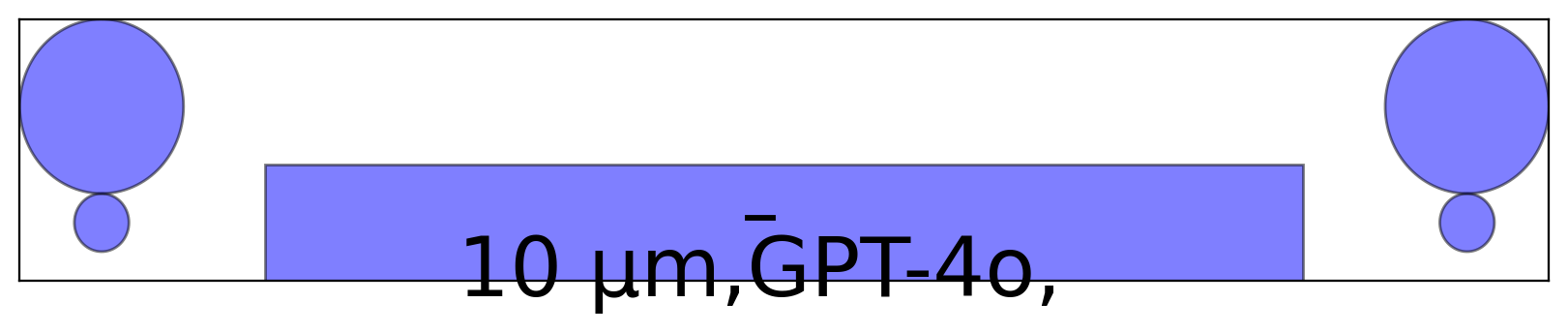} & \includegraphics[width=0.13\textwidth]{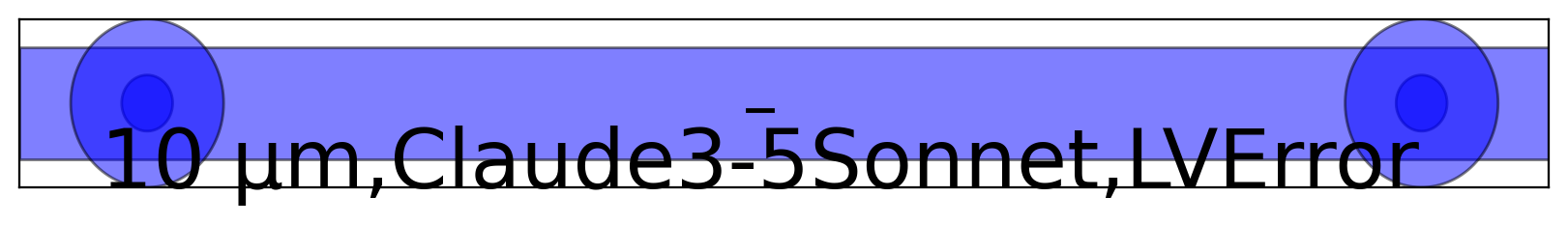} & \includegraphics[width=0.13\textwidth]{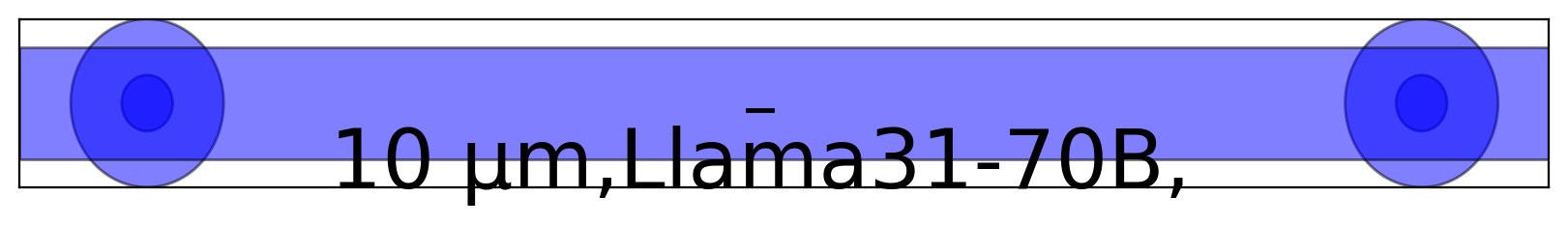} & \includegraphics[width=0.13\textwidth]{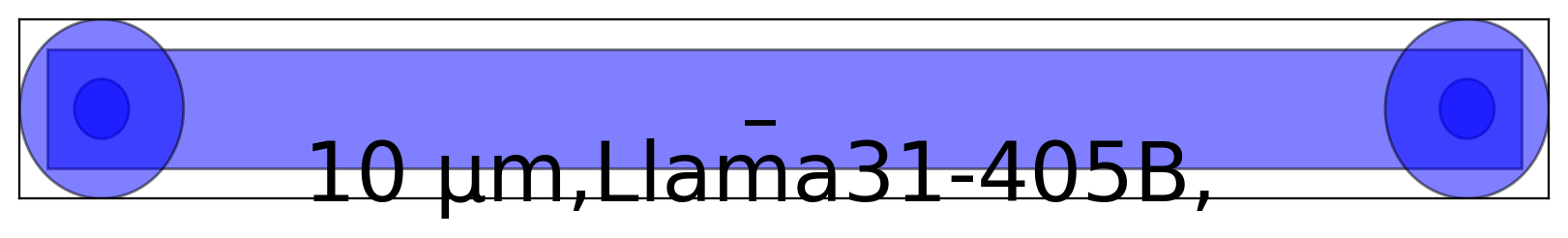} & \includegraphics[width=0.13\textwidth]{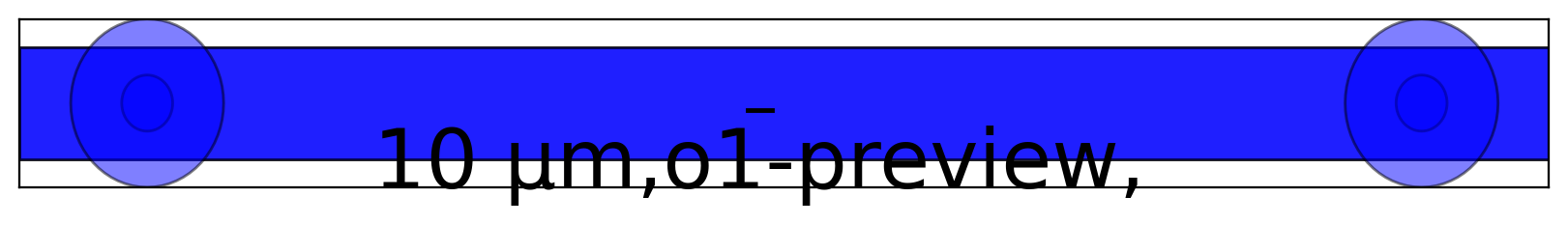} \\
    \begin{tabular}{@{}c@{}}Single LLM \\ Baseline \\ Run 5\end{tabular} & \includegraphics[width=0.13\textwidth]{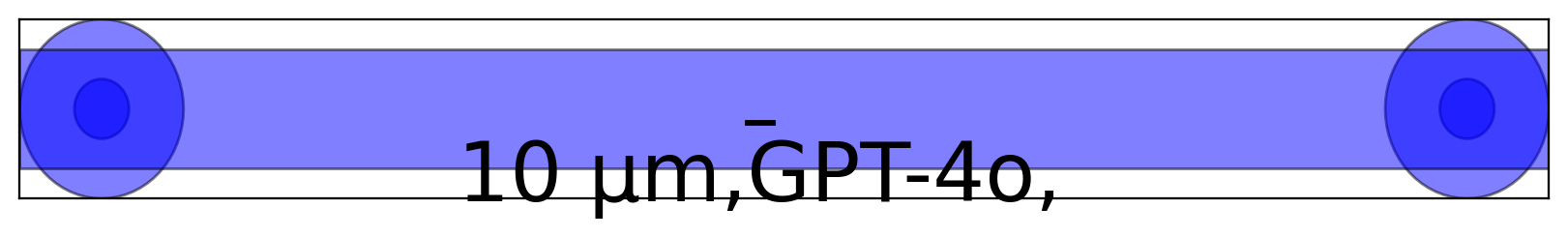} & \includegraphics[width=0.13\textwidth]{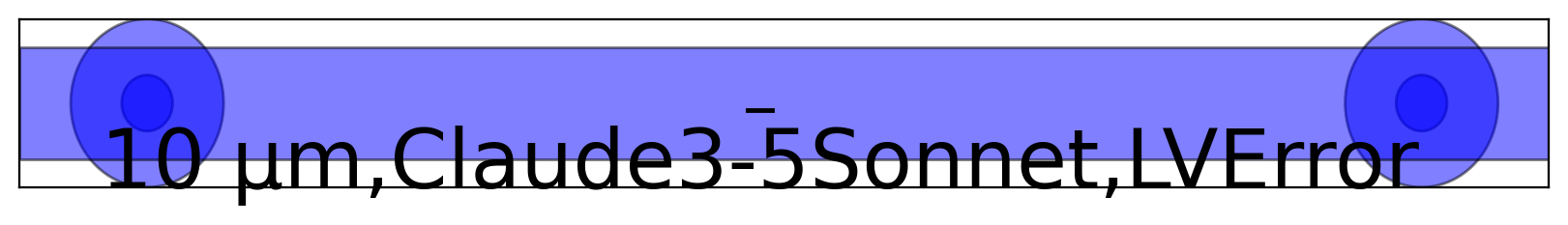} & \includegraphics[width=0.13\textwidth]{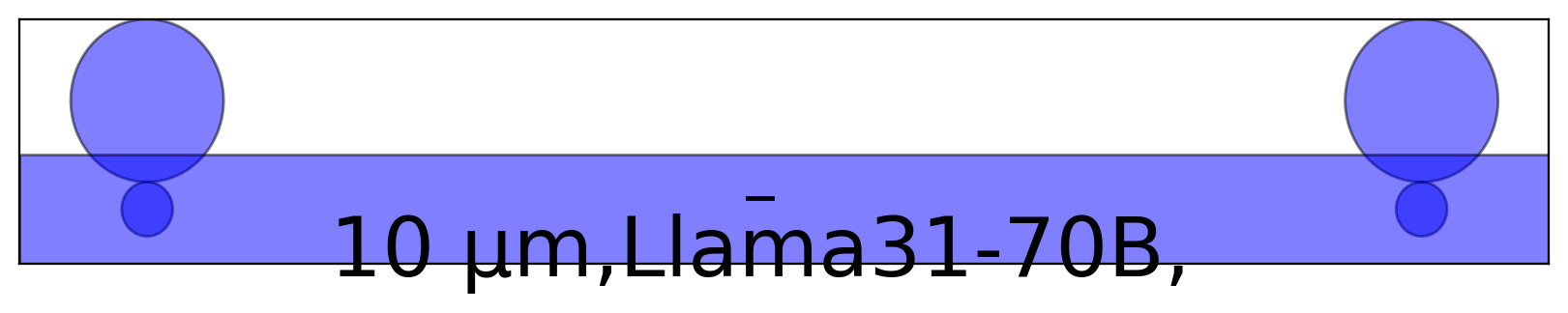} & \includegraphics[width=0.13\textwidth]{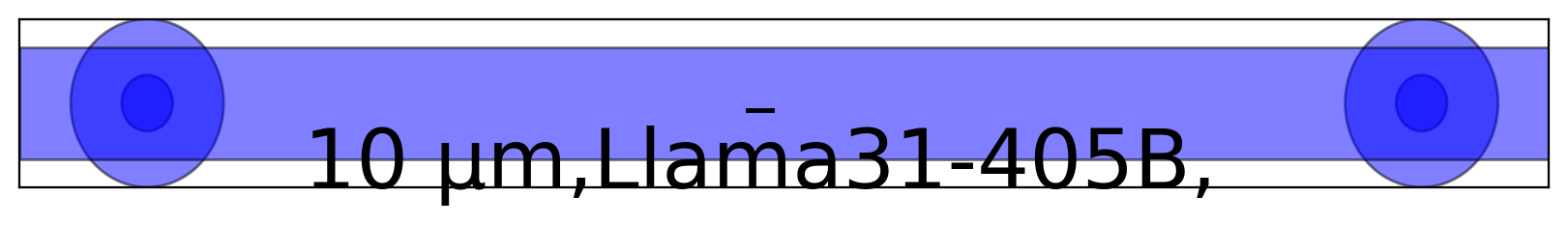} & \includegraphics[width=0.13\textwidth]{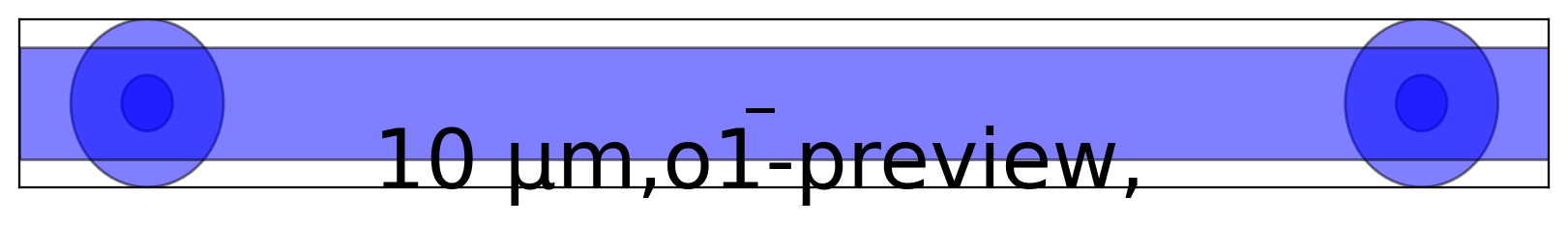} \\
    \bottomrule
  \end{tabularx}
\end{table}

\clearpage
\begin{table}[p]
  \caption{DLDChip Task Question: Draw a deterministic lateral displacement chip - include channel that can hold the array has gap size = 225 nm, circular pillar size = 400 nm, width = 30 pillars, row shift fraction = 0.1, add an inlet and outlet 40 µm diameter before and after the channel, use a 20*50 µm bus to connect the inlet and outlet to the channel.}
  \label{table:dldchip}
  \centering
  \begin{tabularx}{0.9\textwidth}{@{}XXXXXX@{}}
    \toprule
    \begin{tabular}{@{}c@{}}Ground Truth \\ \includegraphics[width=0.13\textwidth]{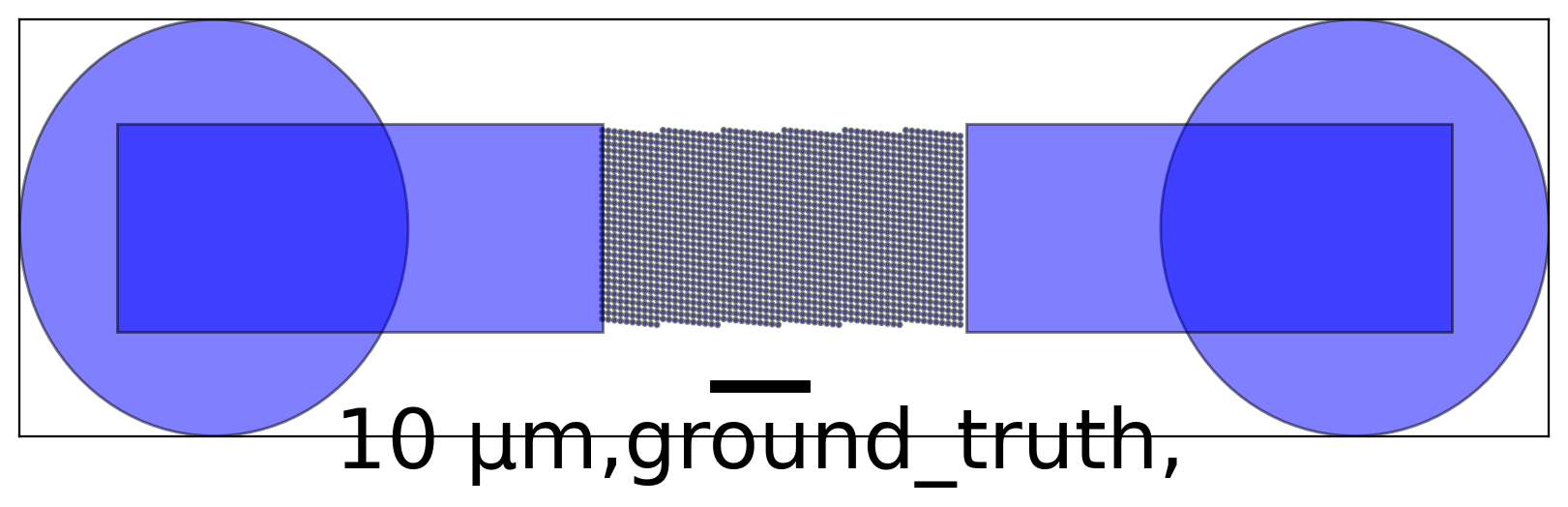}\end{tabular} & GPT-4o & Claude-3.5 & Llama-3-70B & Llama-3-405B & o1-preview \\
    \midrule
    SOLOMON & \includegraphics[width=0.13\textwidth]{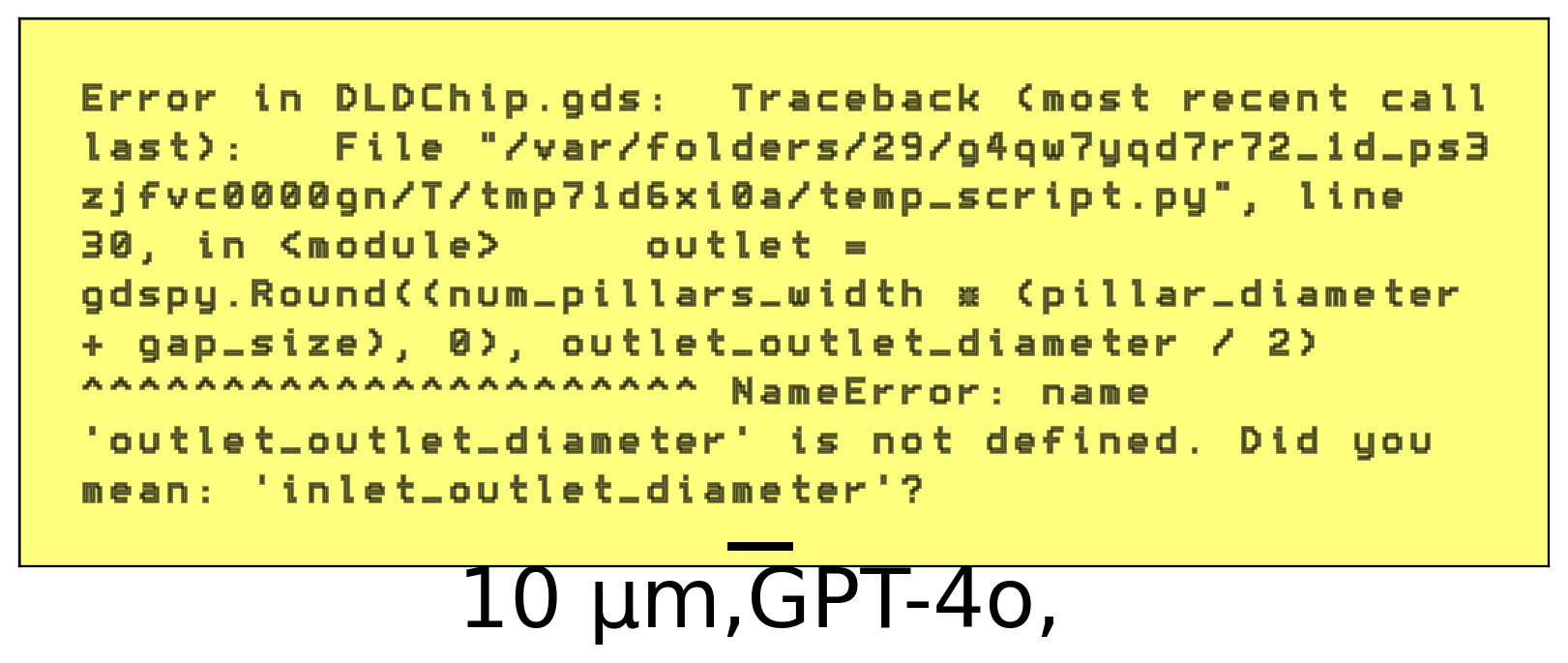} & \includegraphics[width=0.13\textwidth]{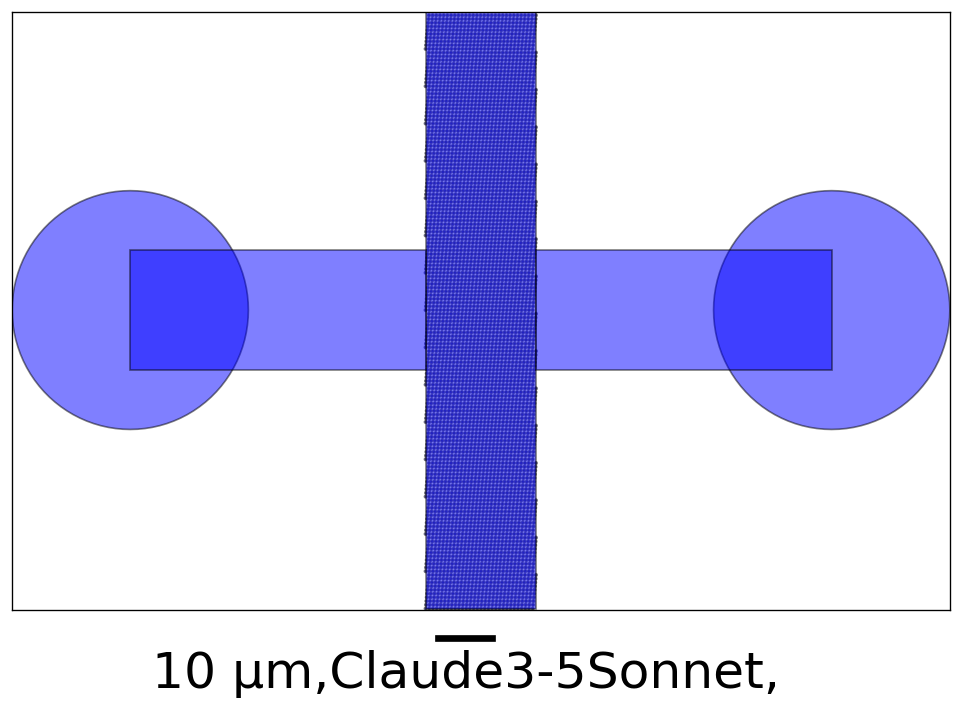} & \includegraphics[width=0.13\textwidth]{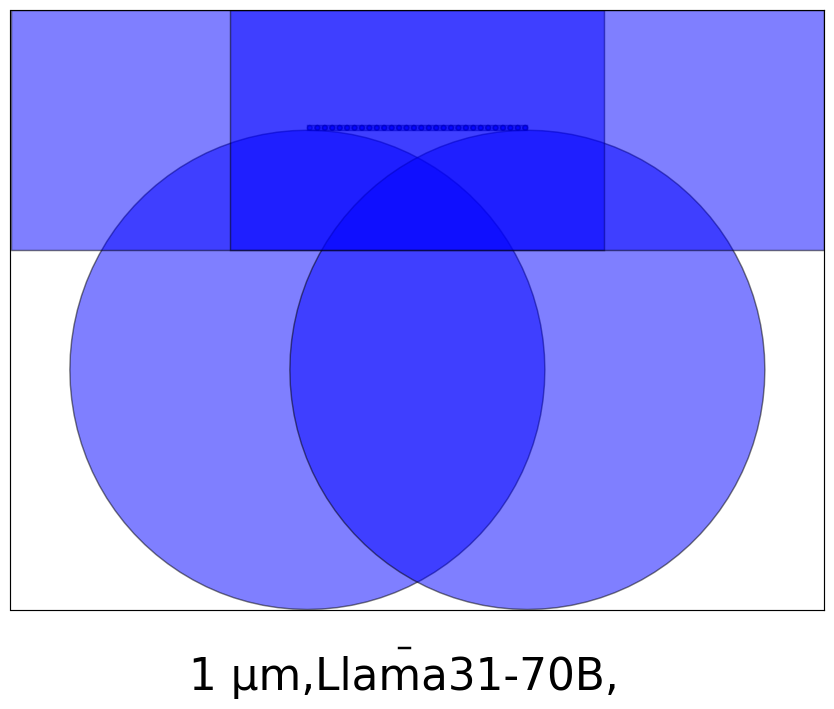} & \includegraphics[width=0.13\textwidth]{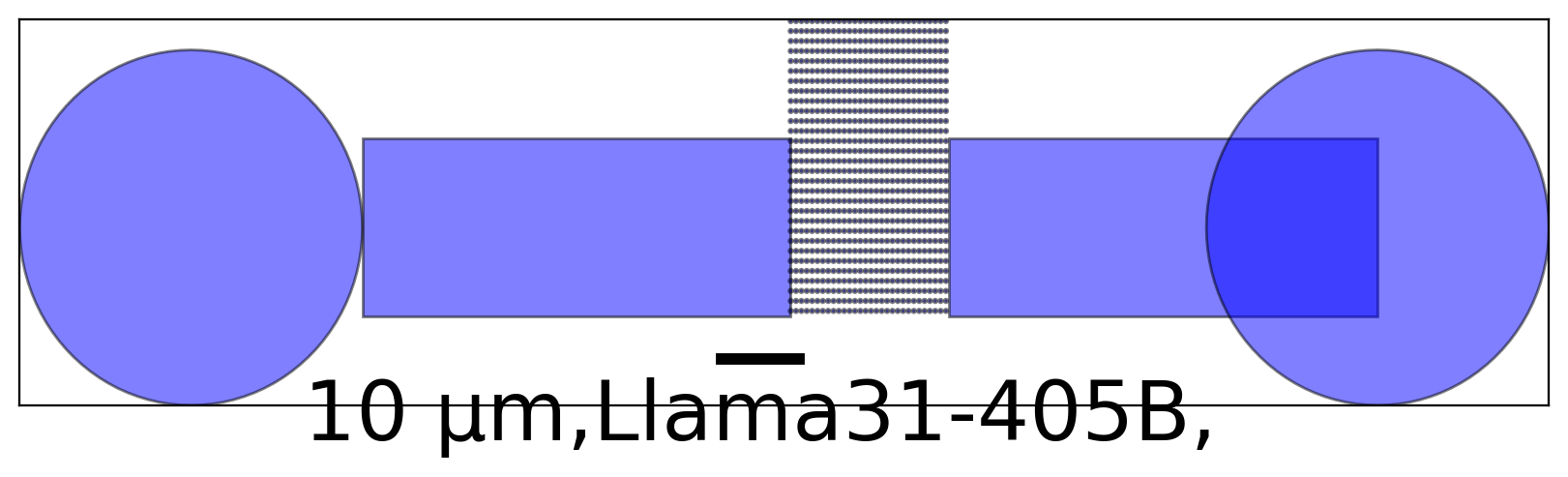} &  \\
    \begin{tabular}{@{}c@{}}Single LLM \\ Baseline \\ Run 1\end{tabular} & \includegraphics[width=0.13\textwidth]{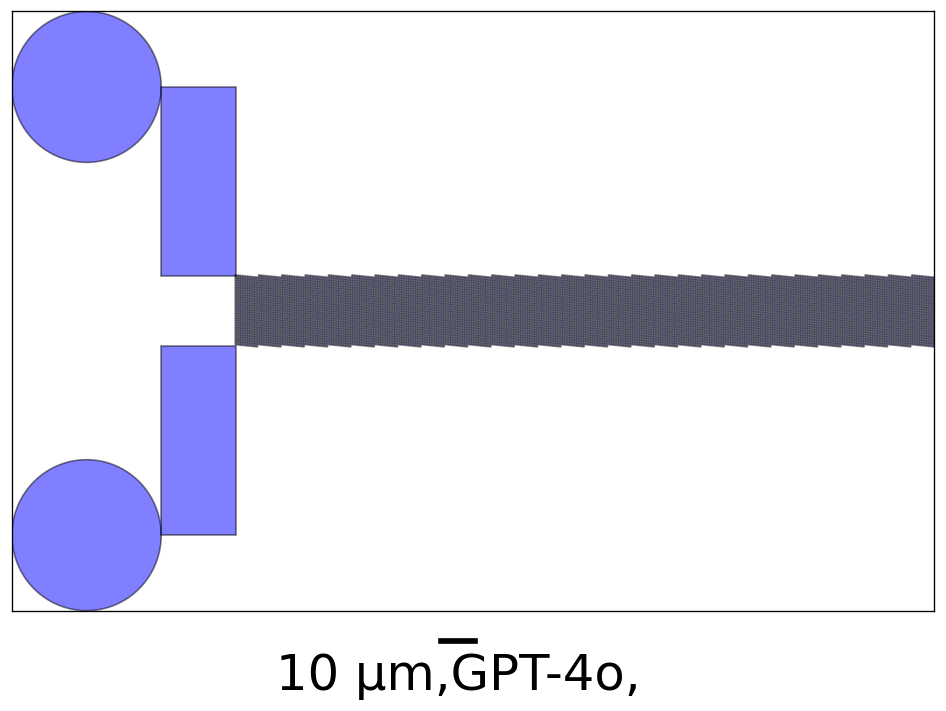} & \includegraphics[width=0.13\textwidth]{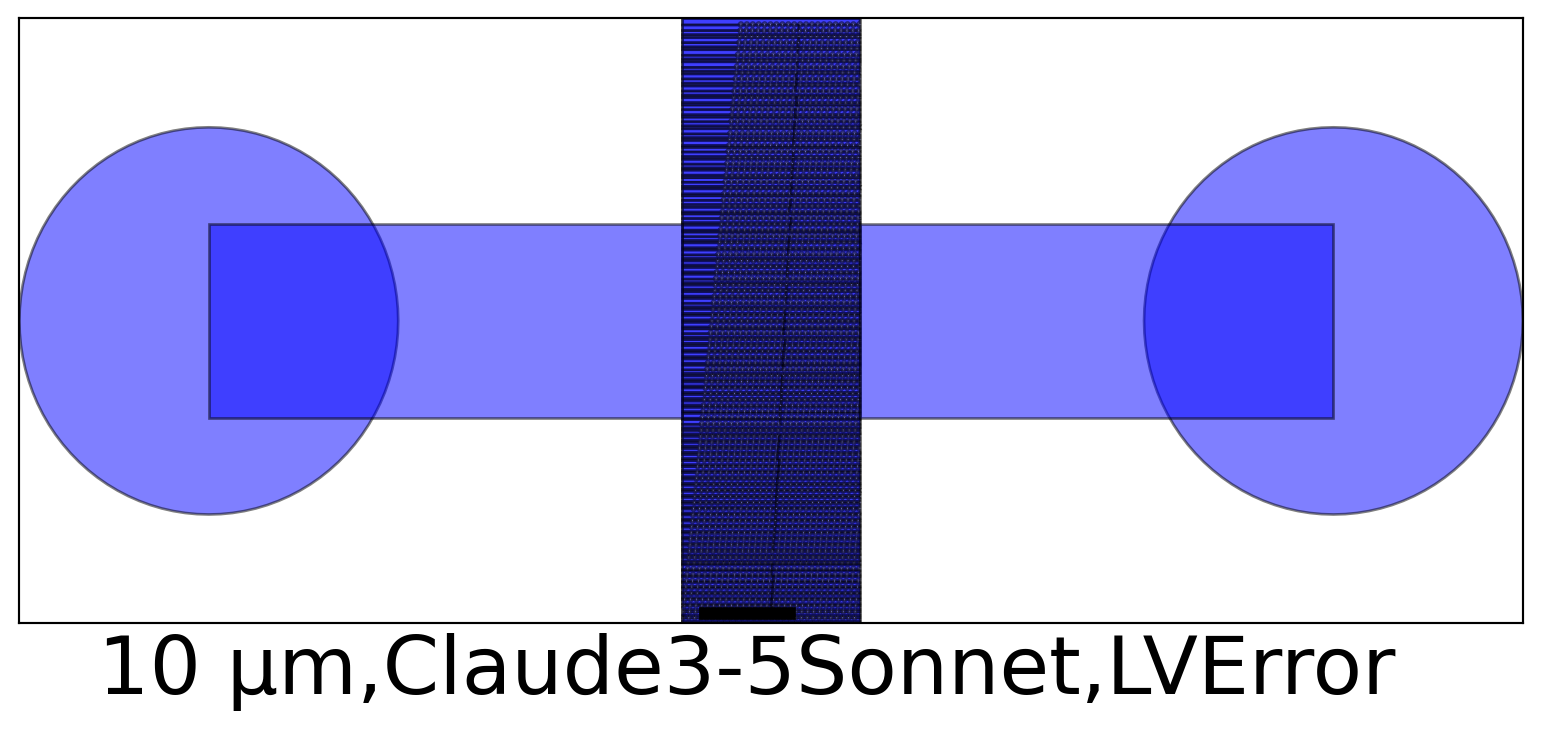} & \includegraphics[width=0.13\textwidth]{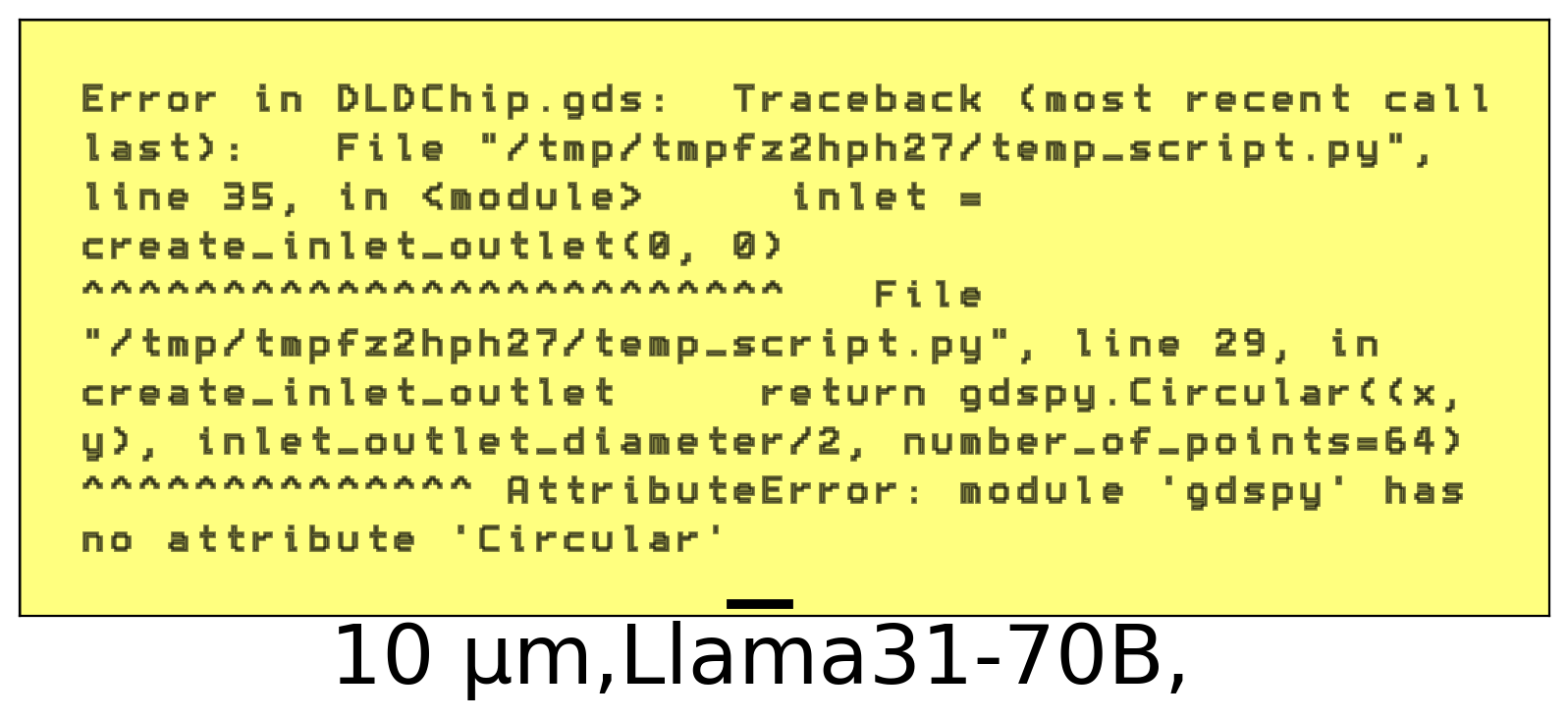} & \includegraphics[width=0.13\textwidth]{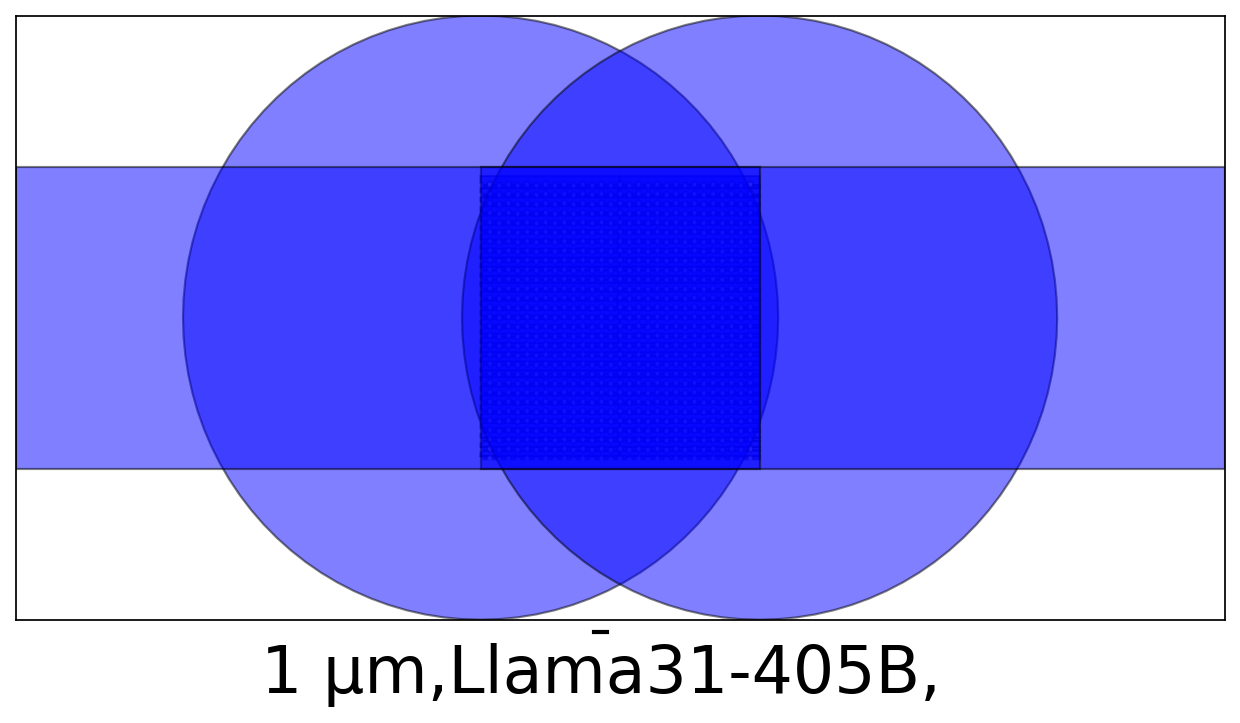} & \includegraphics[width=0.13\textwidth]{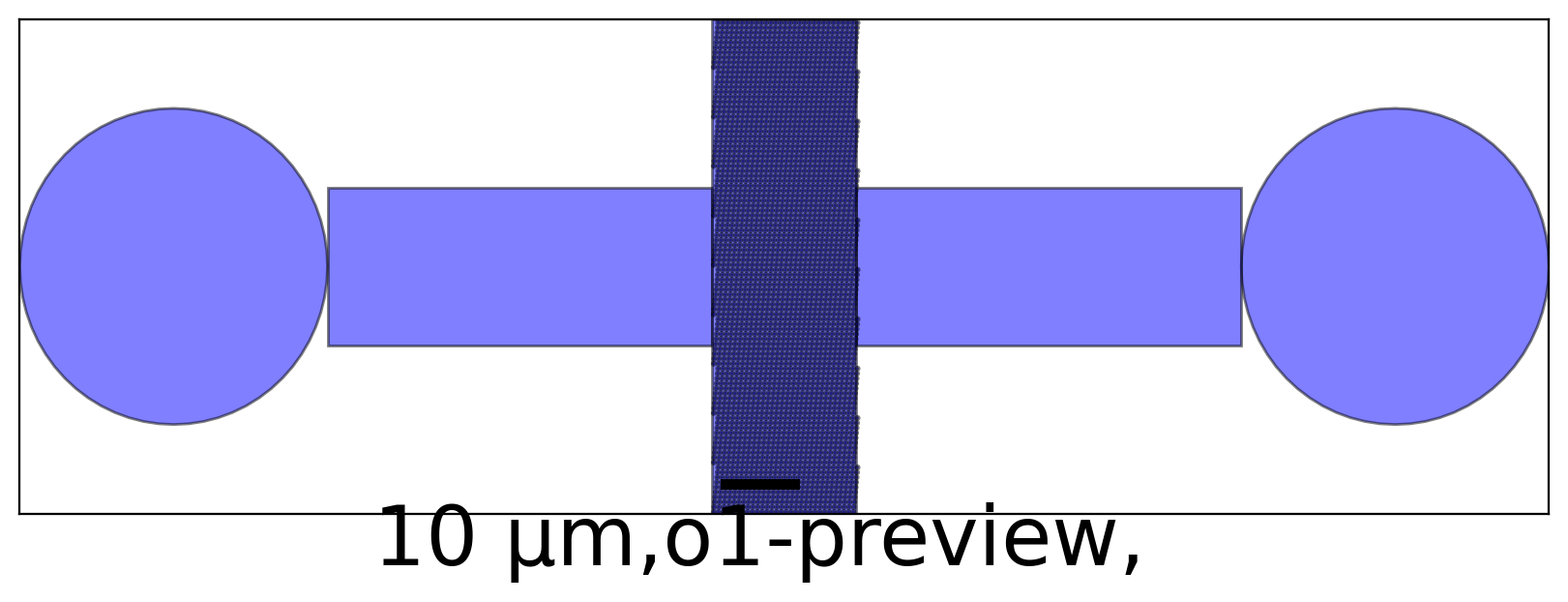} \\
    \begin{tabular}{@{}c@{}}Single LLM \\ Baseline \\ Run 2\end{tabular} & \includegraphics[width=0.13\textwidth]{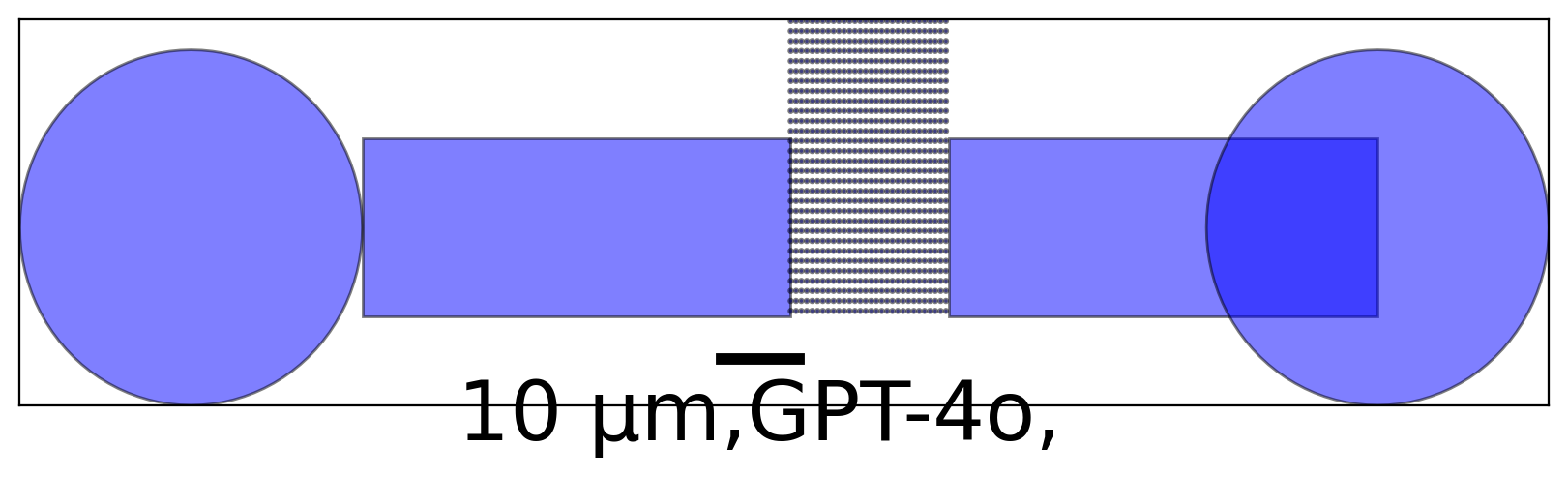} & \includegraphics[width=0.13\textwidth]{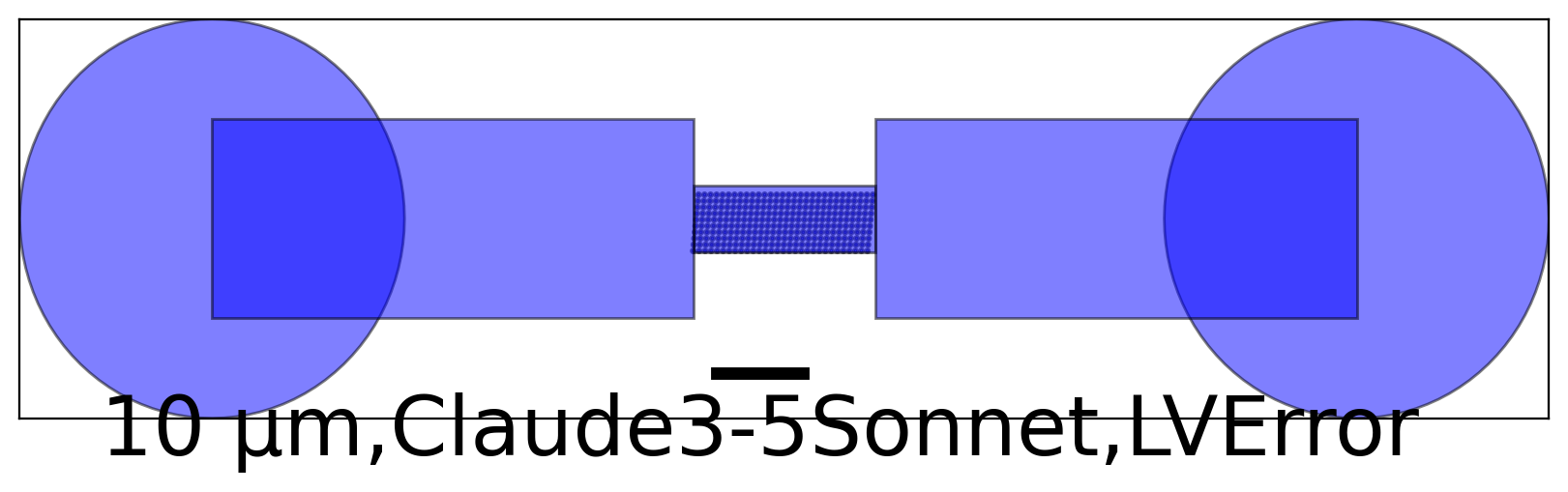} & \includegraphics[width=0.13\textwidth]{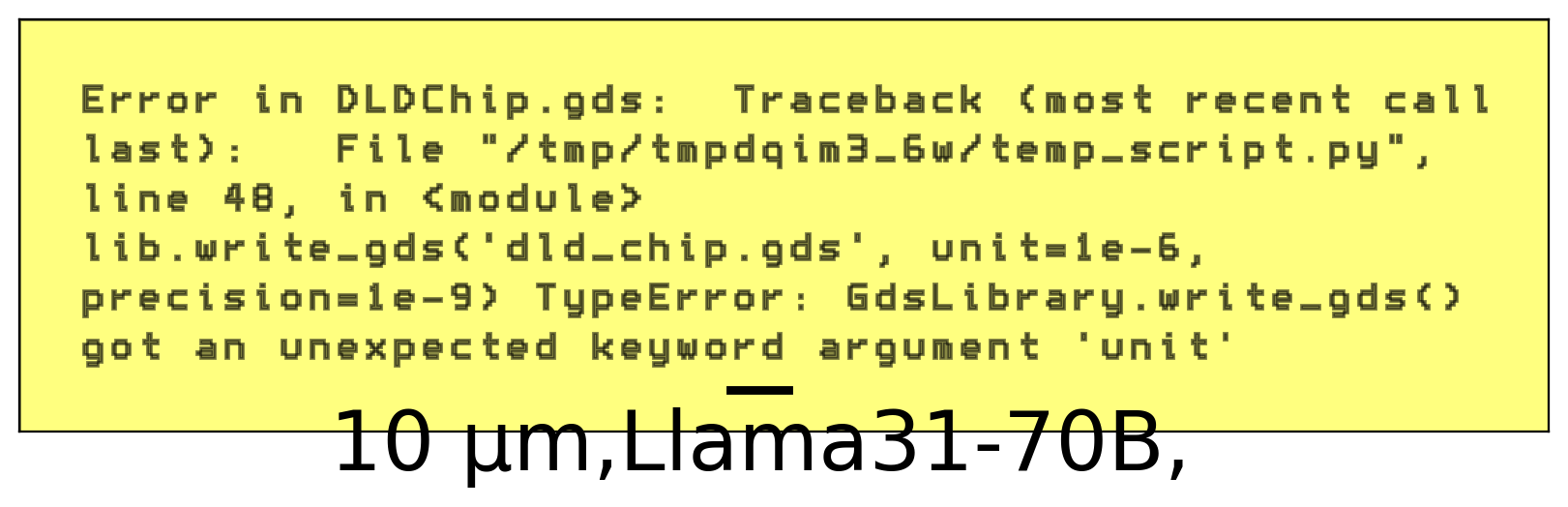} & \includegraphics[width=0.13\textwidth]{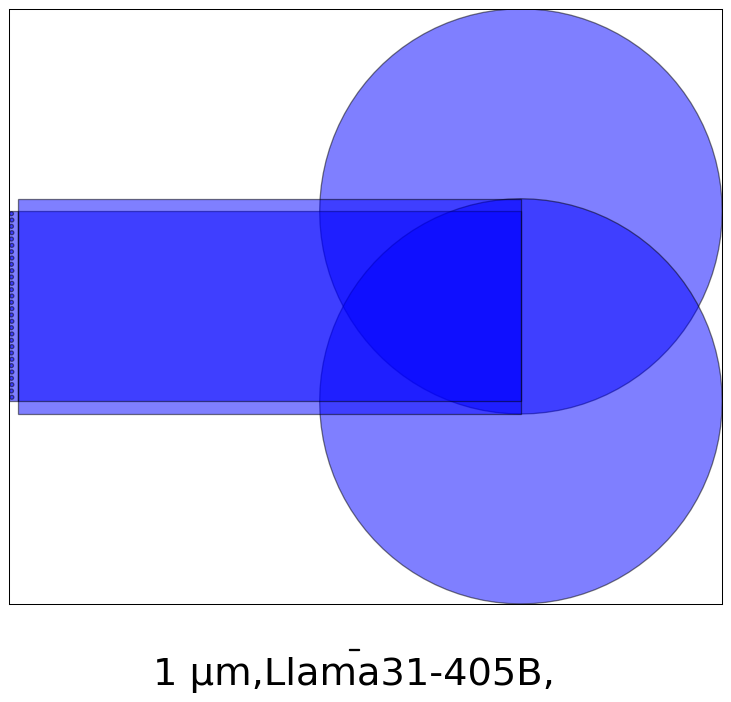} & \includegraphics[width=0.13\textwidth]{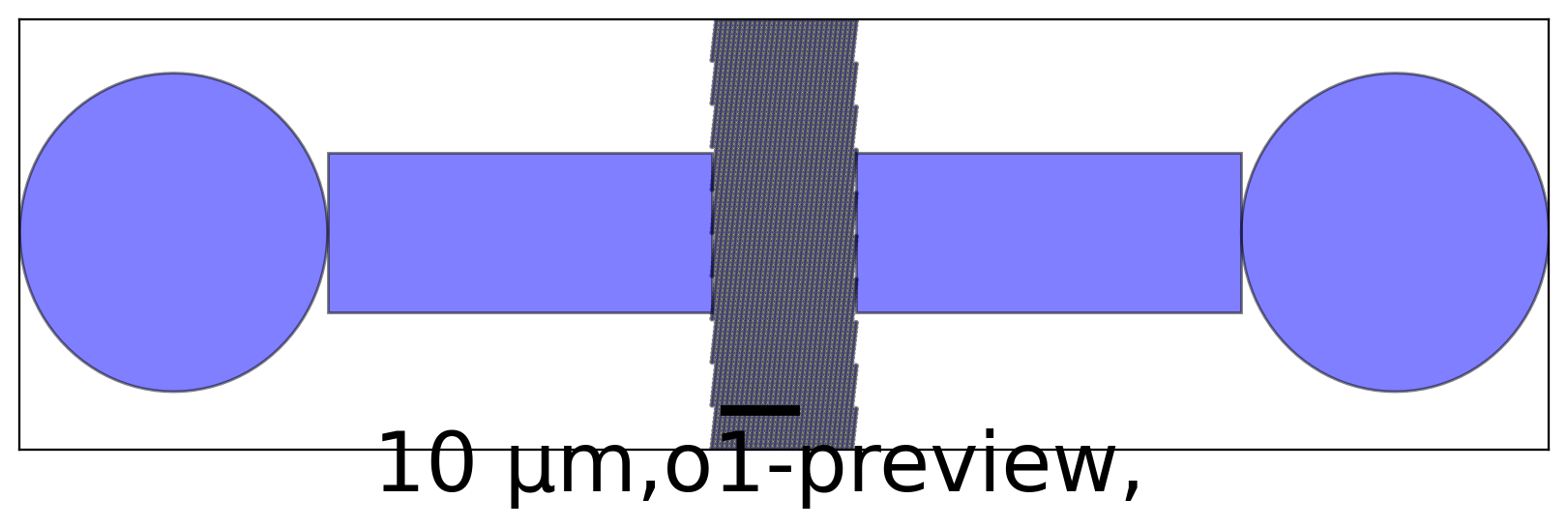} \\
    \begin{tabular}{@{}c@{}}Single LLM \\ Baseline \\ Run 3\end{tabular} & \includegraphics[width=0.13\textwidth]{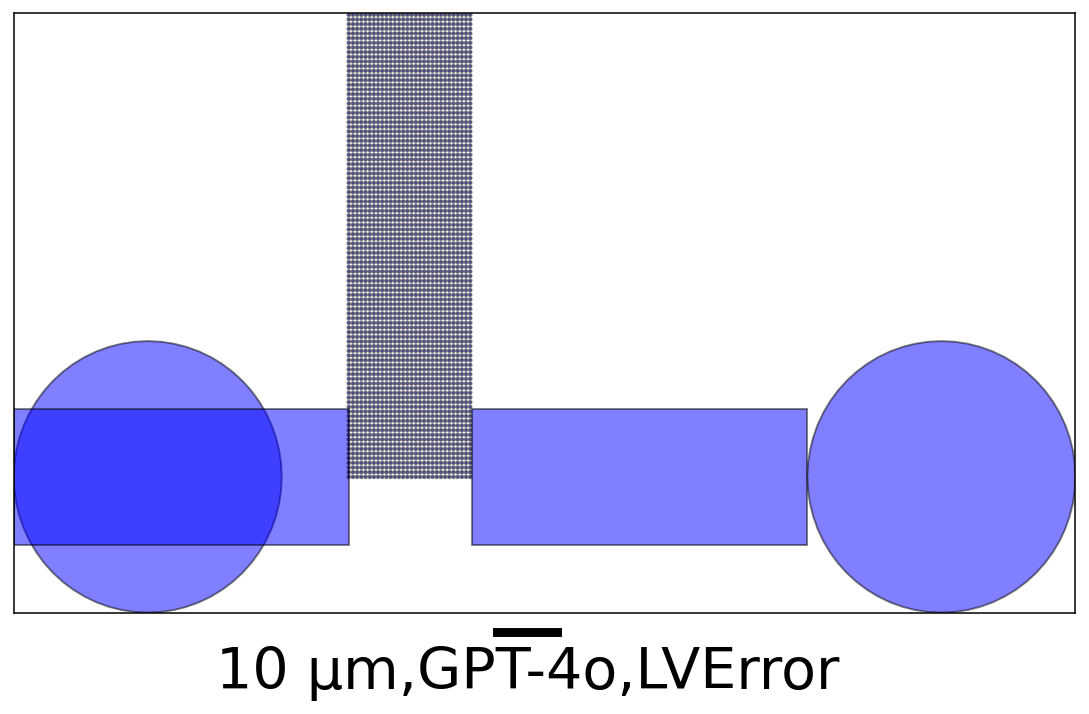} & \includegraphics[width=0.13\textwidth]{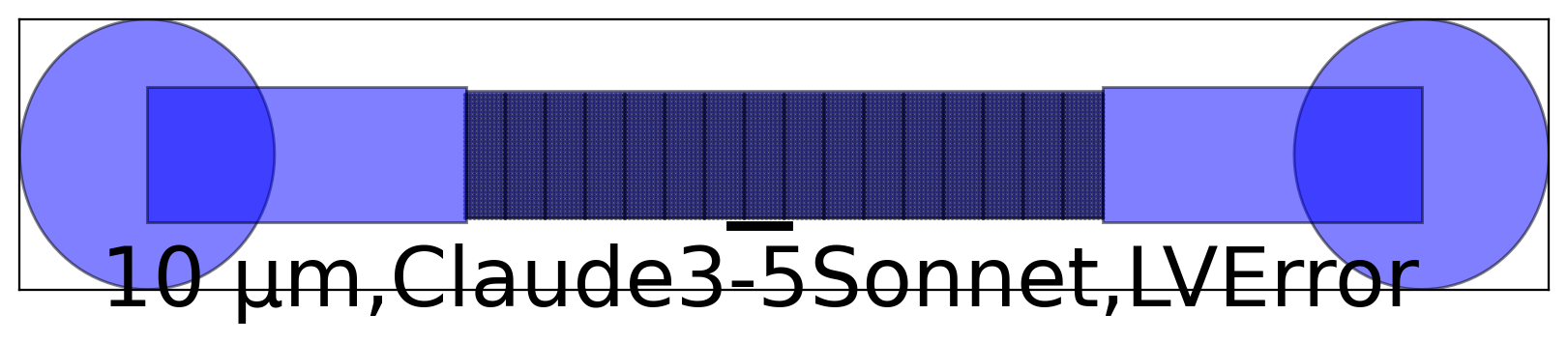} & \includegraphics[width=0.13\textwidth]{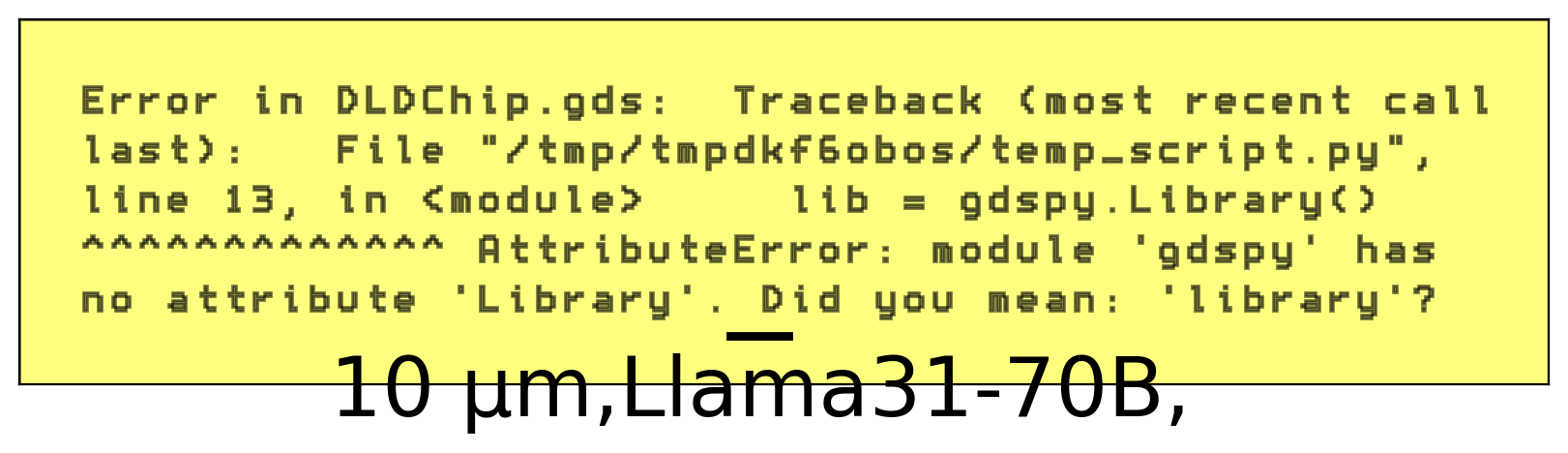} & \includegraphics[width=0.13\textwidth]{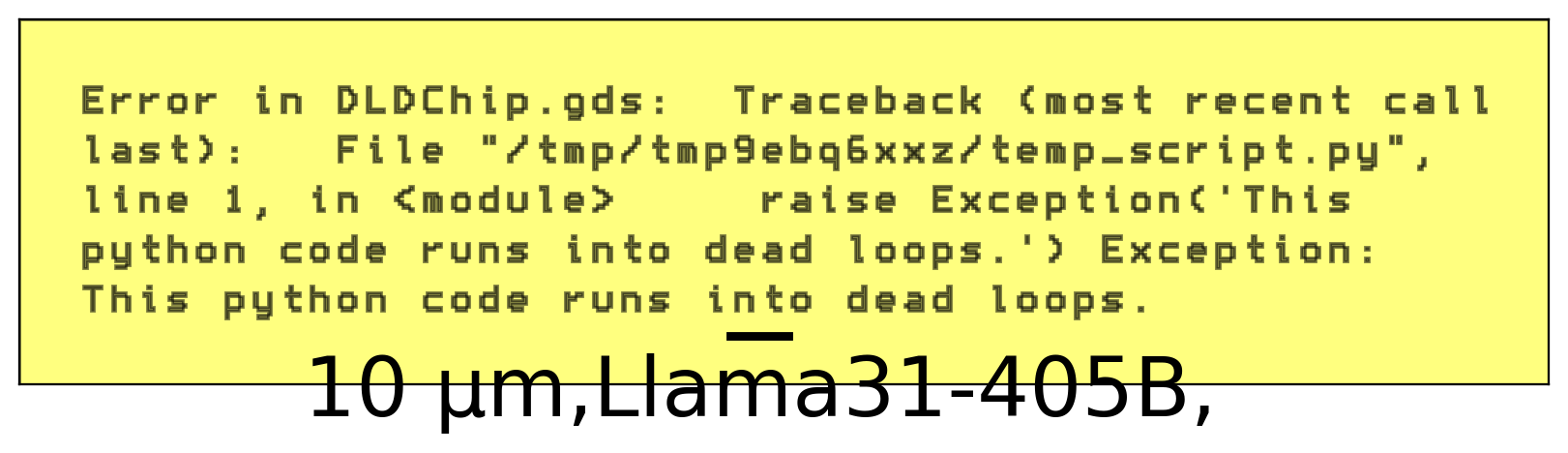} & \includegraphics[width=0.13\textwidth]{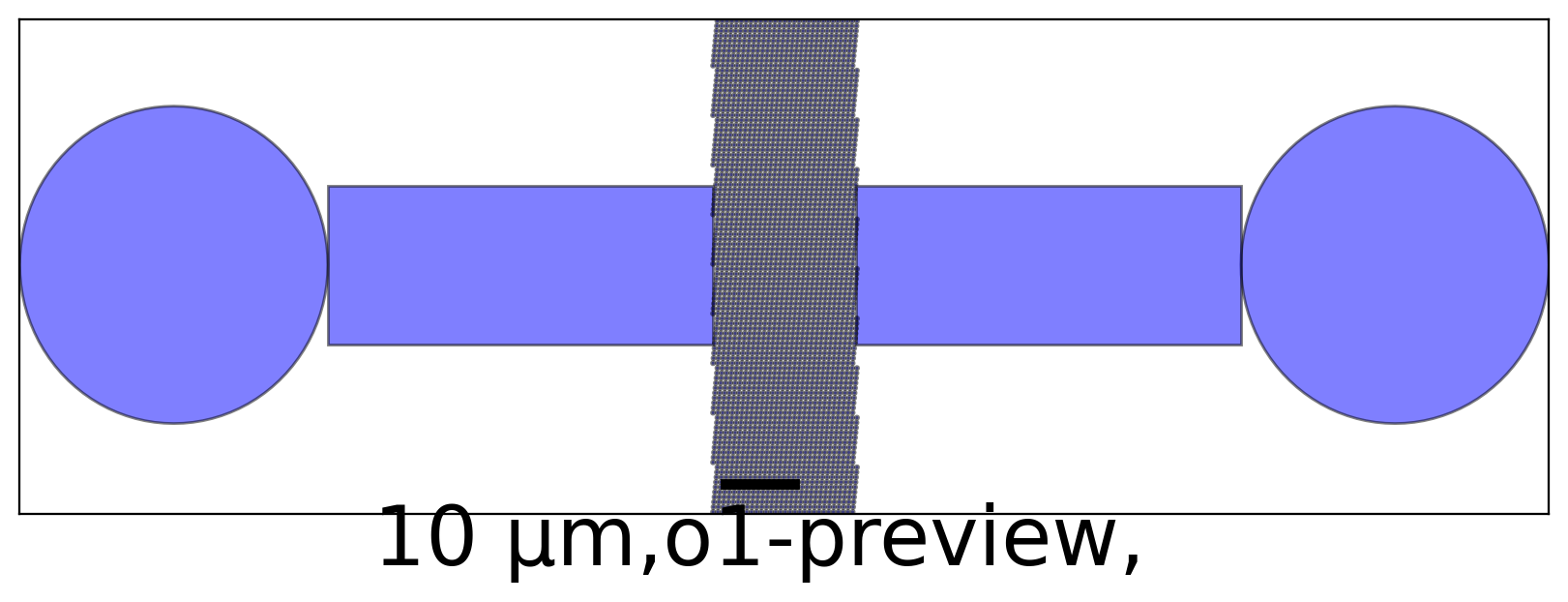} \\
    \begin{tabular}{@{}c@{}}Single LLM \\ Baseline \\ Run 4\end{tabular} & \includegraphics[width=0.13\textwidth]{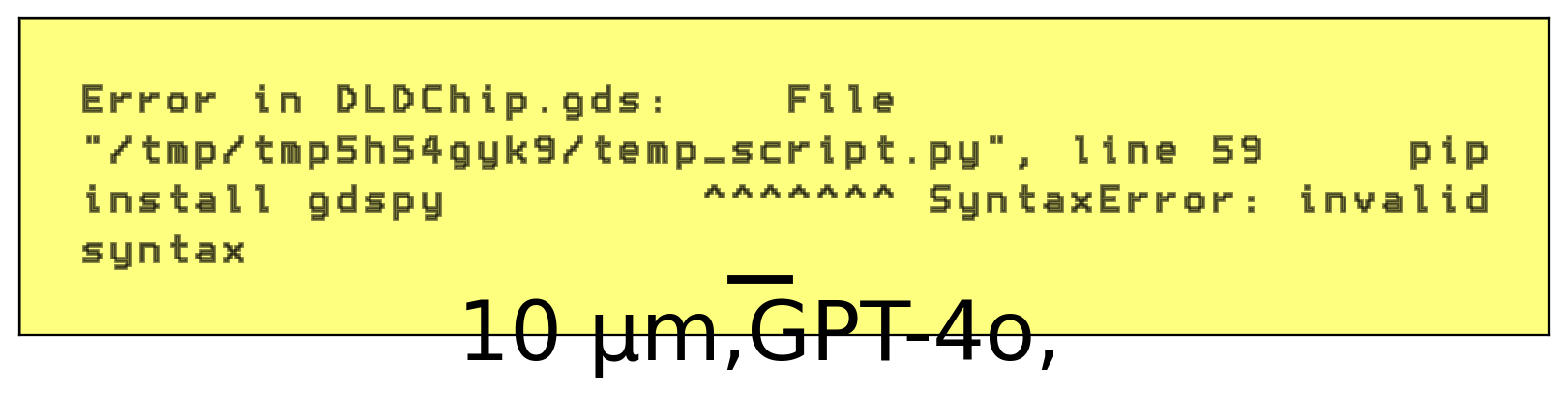} & \includegraphics[width=0.13\textwidth]{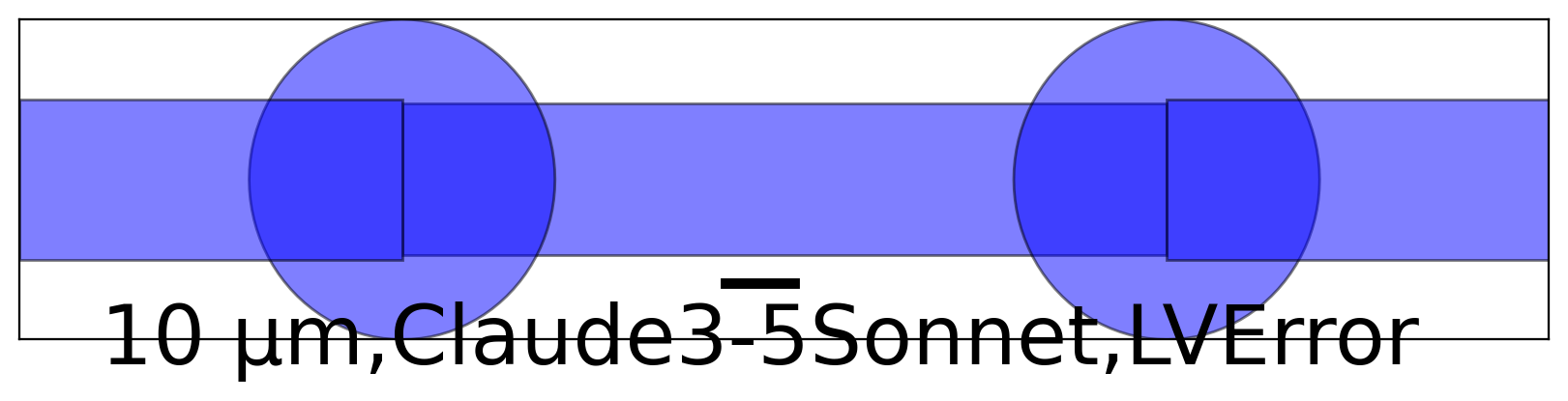} & \includegraphics[width=0.13\textwidth]{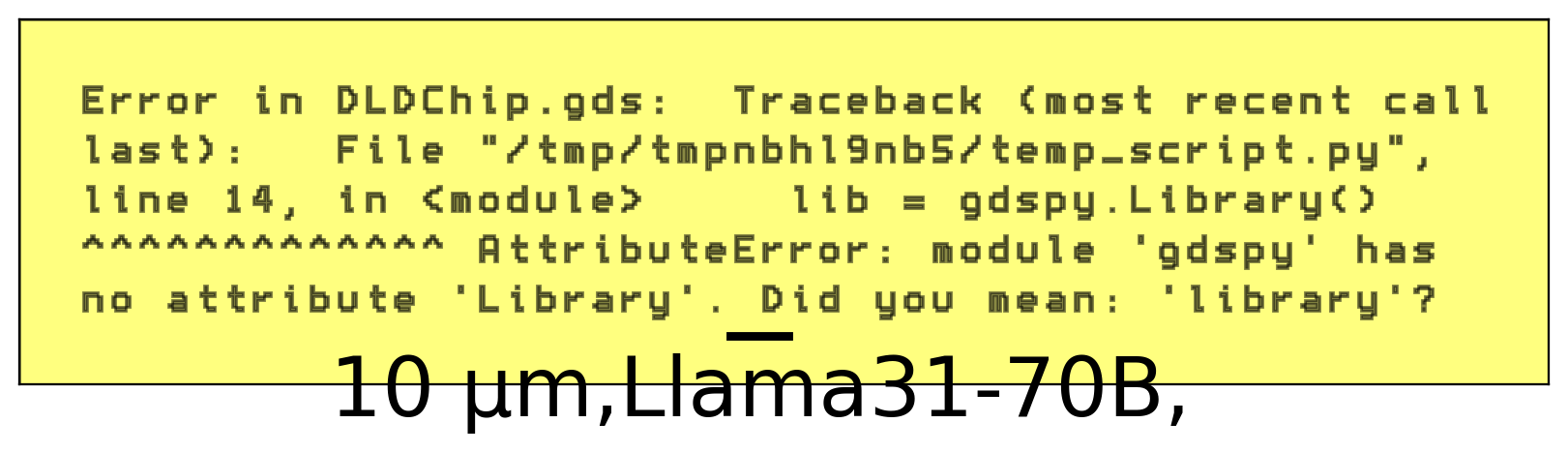} & \includegraphics[width=0.13\textwidth]{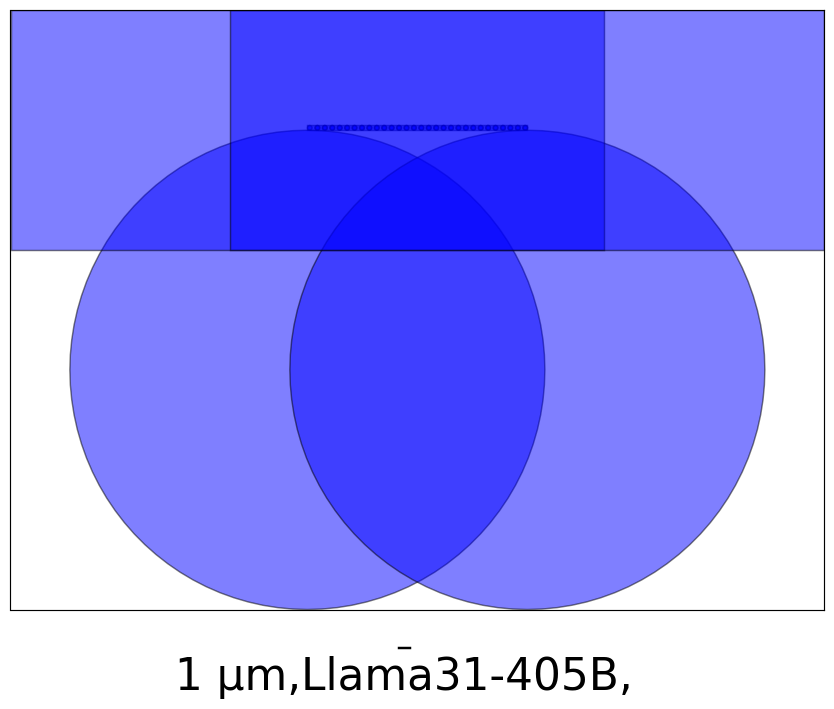} & \includegraphics[width=0.13\textwidth]{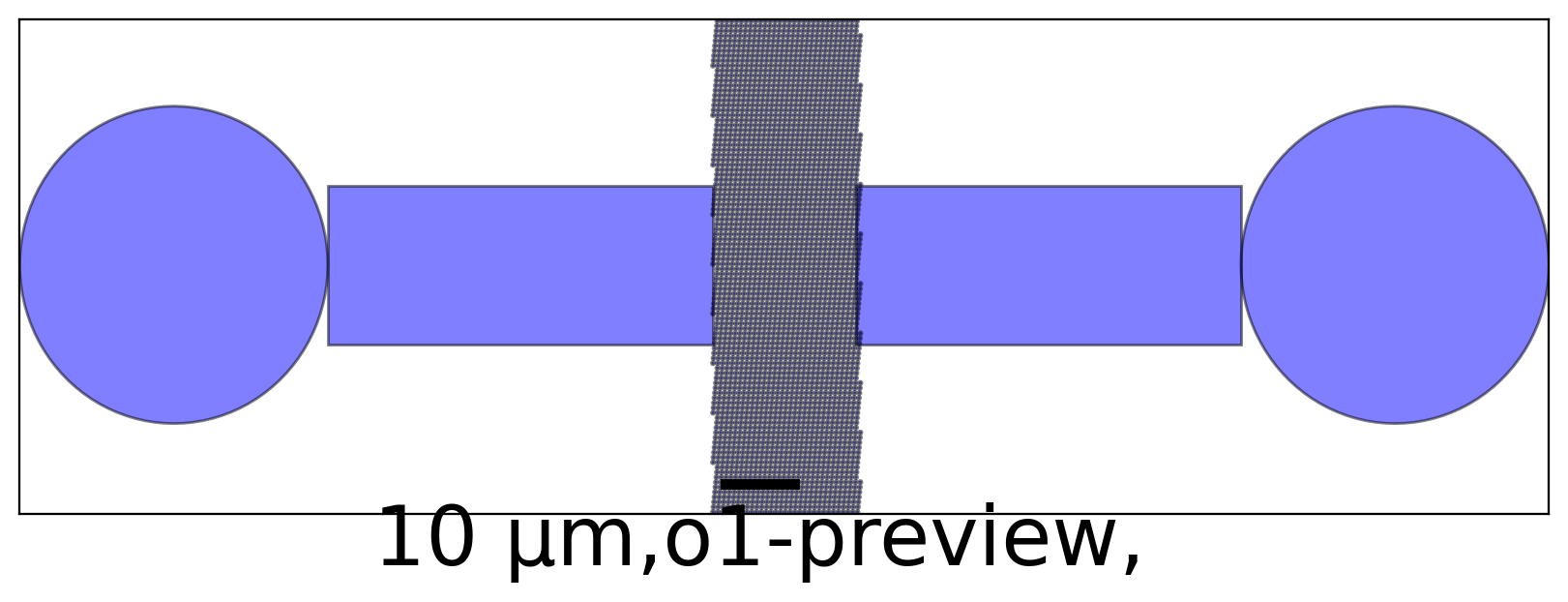} \\
    \begin{tabular}{@{}c@{}}Single LLM \\ Baseline \\ Run 5\end{tabular} & \includegraphics[width=0.13\textwidth]{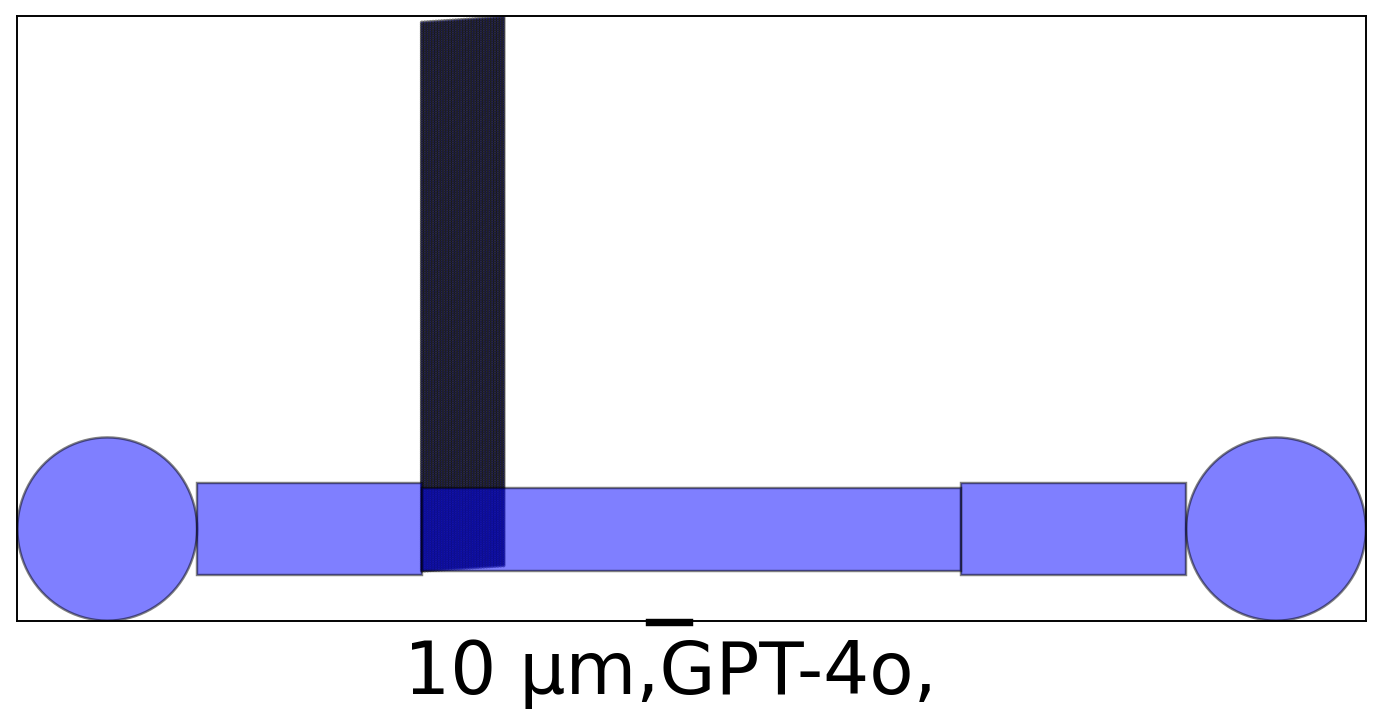} & \includegraphics[width=0.13\textwidth]{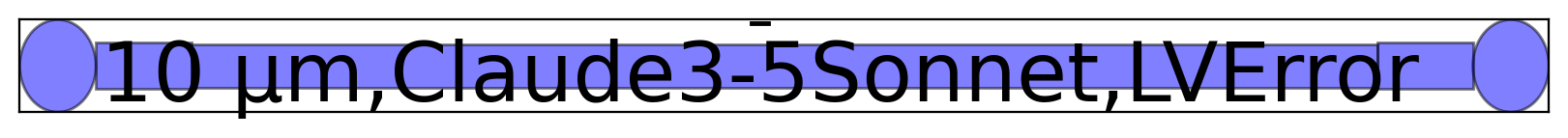} & \includegraphics[width=0.13\textwidth]{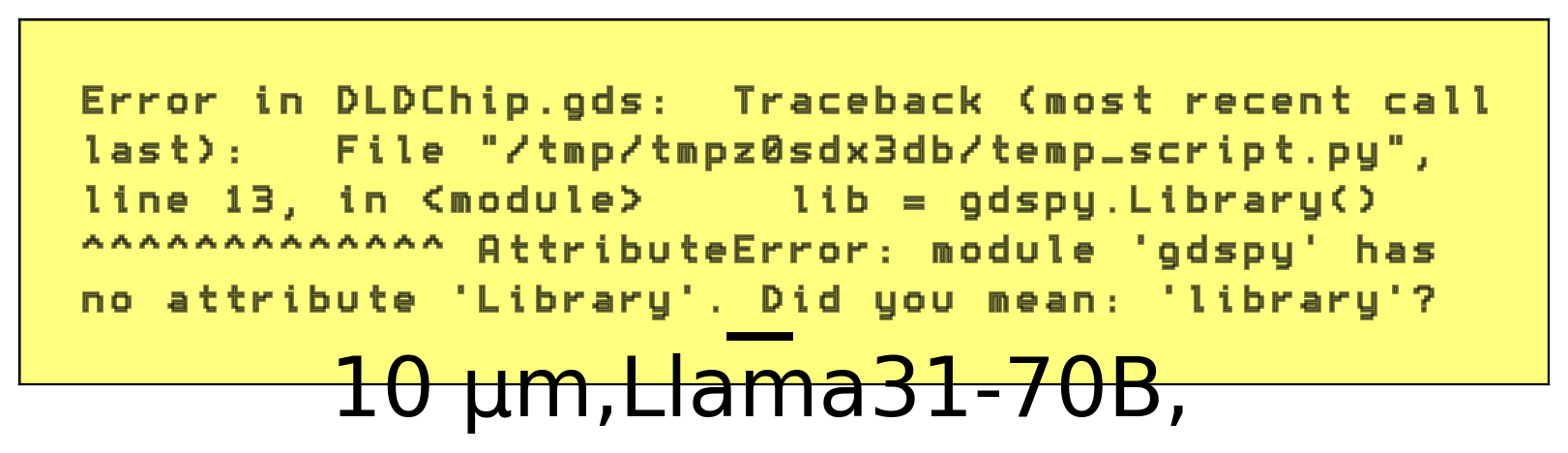} & \includegraphics[width=0.13\textwidth]{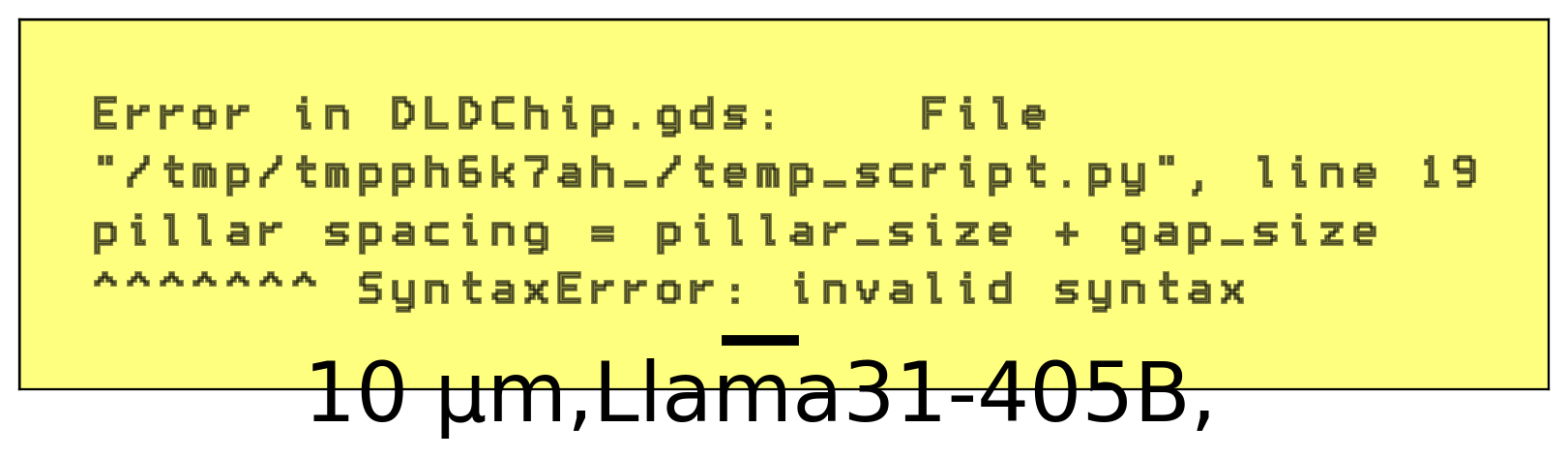} & \includegraphics[width=0.13\textwidth]{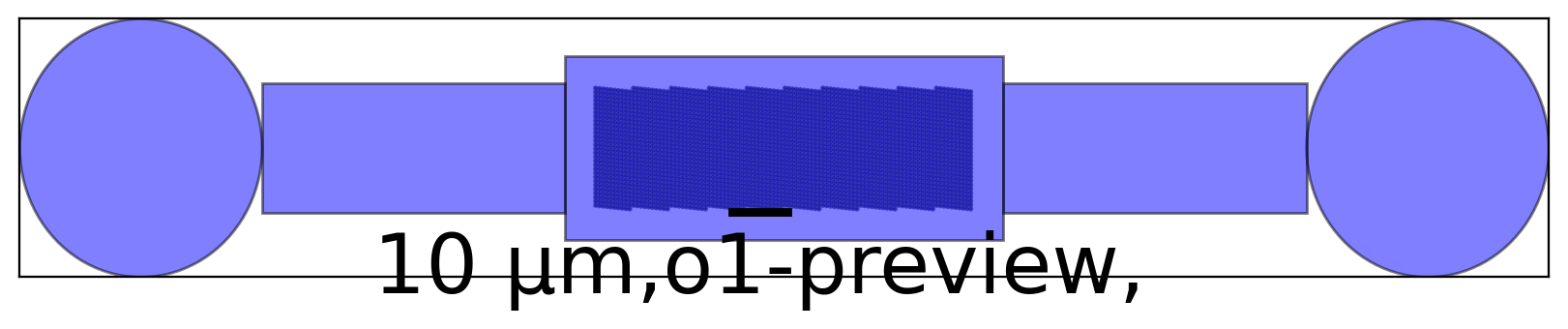} \\
    \bottomrule
  \end{tabularx}
\end{table}



\end{document}